\documentclass[journal,twocolumn]{IEEEtran}
\usepackage{cite}
\usepackage{amsmath,amssymb,amsfonts,bm}

\usepackage{algorithmic}
\usepackage{graphicx}
\usepackage{subfigure}
\usepackage{textcomp}
\usepackage{makecell}
\usepackage{booktabs}
\usepackage{multirow}
\usepackage{color}
\usepackage{bm}

\usepackage{amsthm}
\usepackage{algorithm}
\usepackage{algorithmic}
\usepackage{hyperref}
\usepackage[table]{xcolor}
\usepackage{subcaption}

\ifCLASSINFOpdf
\else
\fi
\begin{document}
%
%
%
%
\title{FL-MedSegBench: A Comprehensive Benchmark for Federated Learning on Medical Image Segmentation}
\author{Meilu Zhu, Zhiwei Wang, Axiu Mao, Yuxing Li, Xiaohan Xing, Yixuan Yuan, Edmund Y. Lam* \thanks{This work described in this paper was fully supported by a grant (ITS/341/23) from the Innovation and Technology Fund of Hong Kong, China.}
\thanks{
M. Zhu, Z. Wang, Y. Li, and Edmund Y. Lam are with Department of Electrical and Computer Engineering, The University of Hong Kong, Hong Kong, China;
A. Mao is with School of Communication Engineering, Hangzhou Dianzi University, Hang Zhou, China;
X. Xing is with Department of Radiation Oncology, Stanford University, California, USA;
Y. Yuan is with Department of Electronic Engineering, Chinese University of Hong Kong, Hong Kong, China.}
\thanks{M. Zhu and Z. Wang contributed equally to this work. Edmund Y. Lam is the corresponding author  (elam@eee.hku.hk).}
}

\markboth{Journal of \LaTeX\ Class Files,~Vol.~14, No.~8, August~2015}%
{Shell \MakeLowercase{\textit{et al.}}: Bare Demo of IEEEtran.cls for IEEE Journals}
%



\maketitle

\begin{abstract}
Federated learning (FL) offers a privacy-preserving paradigm for collaborative medical image analysis without sharing raw data. However, the absence of standardized benchmarks for medical image segmentation hinders fair and comprehensive evaluation of FL methods. To address this gap, we introduce FL-MedSegBench, the first comprehensive benchmark for federated learning on medical image segmentation. Our benchmark encompasses nine segmentation tasks across ten imaging modalities, covering both 2D and 3D formats with realistic clinical heterogeneity. We systematically evaluate eight generic FL (gFL) and five personalized FL (pFL) methods across multiple dimensions: segmentation accuracy, fairness, communication efficiency, convergence behavior, and generalization to unseen domains. Extensive experiments reveal several key insights: (i) pFL methods, particularly those with client-specific batch normalization (\textit{e.g.}, FedBN), consistently outperform generic approaches; (ii) No single method universally dominates, with performance being dataset-dependent; (iii) Communication frequency analysis shows normalization-based personalization methods exhibit remarkable robustness to reduced communication frequency; (iv) Fairness evaluation identifies methods like Ditto and FedRDN that protect underperforming clients; (v) A method's generalization to unseen domains is strongly tied to its ability to perform well across participating clients. We will release an open-source toolkit to foster reproducible research and accelerate clinically applicable FL solutions, providing empirically grounded guidelines for real-world clinical deployment. The source code is available at \href{https://github.com/meiluzhu/FL-MedSegBench}{https://github.com/meiluzhu/FL-MedSegBench}.
\end{abstract}

\section{Introduction}\label{sec1}
Medical image segmentation~\cite{fan2020pranet,zhu2021dsi, zhong2025cross, huang2025multidimensional} plays a vital role in clinical decision-making, enabling precise localization of anatomical structures and pathological regions. Deep learning models have achieved remarkable success in this domain~\cite{butoi2023universeg, azad2024medical}. However, their training typically relies on large and centralized datasets—a requirement that directly conflicts with patient privacy regulations (\textit{e.g.}, HIPAA~\cite{gostin2009beyond}, GDPR~\cite{goddard2017eu}) and the inherently distributed nature of medical data across hospitals and institutions~\cite{fedavg, FedOSS}. Federated learning (FL) has emerged as a privacy-preserving alternative, allowing multiple institutions to collaboratively train models without exchanging local data~\cite{fedavg,fedprox,zhufeddm2023, zhou2023fedftn, jin2023backdoor}. However, data from different sites exhibit inherent heterogeneity due to differences in imaging protocols, device manufacturers, patient populations, and clinical practices~\cite{li2020federated, yang2021federated}. It poses significant challenges to the convergence and generalization of FL models on segmentation tasks~\cite{guan2024federated, kim2024federated, li2025challenges}.

Pioneering FL researchers devoted substantial efforts to develop a global model that performs effectively across all participating clients~\cite{fedavg}. These works, often categorized under generic Federated Learning (\textbf{gFL}), mainly focus on regularizing local training to align client objectives~\cite{li2020federated, li2021model}, designing advanced server-side aggregation rules to improve global model fusion~\cite{shi2025fedawa, wang2020tackling, shi2025fedlws, zhang2025pathfl}, and reducing domain shift with data augmentation~\cite{yan2025simple}. Considering that a single shared global model cannot perform optimally for all clients simultaneously, personalized FL (\textbf{pFL}) has been gaining increasing popularity, which allows each client to learn client-specific models to adapt to local data~\cite{tan2022towards, chen2021personalized, chen2022personalized, liu2025federated}. 
Existing pFL approaches can be broadly categorized into: parameter decoupling (sharing base layers while personalizing higher layers)~\cite{arivazhagan2019federated, chen2021bridging}, model interpolation (mixing global and local models)~\cite{li2021ditto} and batch normalization layer personalization~\cite{andreux2020siloed, li2021fedbn}. The overarching goal of pFL is to balance generalization and specialization, thereby achieving superior performance on heterogeneous local data distributions.

Though fruitful FL methods have been explored, a standardized benchmark for medical image segmentation is still lacking. As a result, current evaluations of FL methods in this domain suffer from the absence of standardized benchmarks and inconsistent implementations, leading to ambiguous conclusions regarding their effectiveness and robustness across diverse practical scenarios. Although two FL benchmarks~\cite{ogier2022flamby, manthe2024federated} have been proposed, they lack comprehensiveness in terms of data scale, task diversity, and modality coverage. For example, FLamby~\cite{ogier2022flamby} only contains three 3D segmentation datasets (covering only two modalities) and does not evaluate pFL methods. Manthe et al.~\cite{manthe2024federated} built a benchmark only for brain tumor segmentation, lacking task and modality diversity.

\begin{figure*}[!t]
	\centering
	\!\!\includegraphics[width=0.99\textwidth]{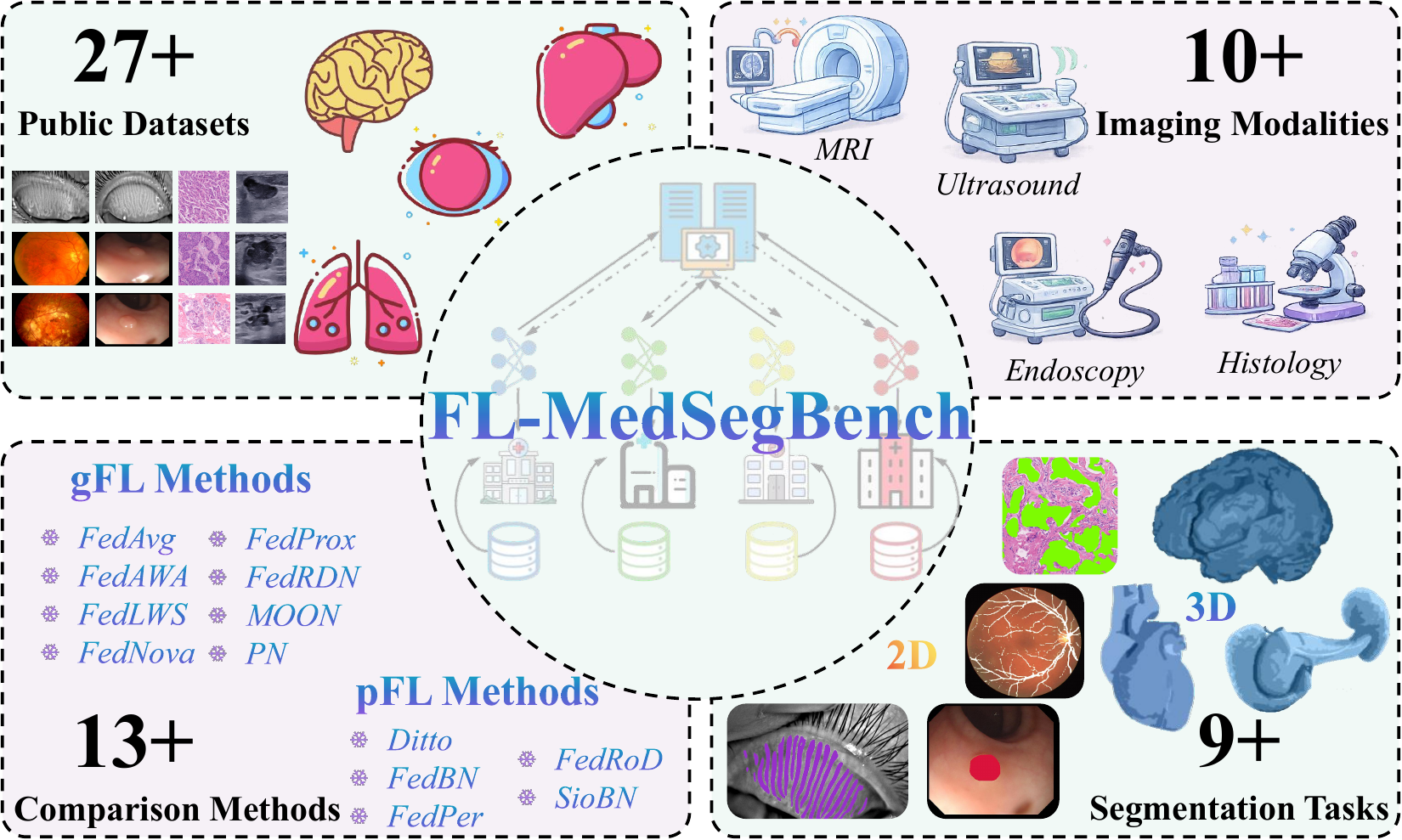}
	\caption{FL-MedSegBench: a comprehensive benchmark for federated learning on medical image segmentation.} 
	\vspace{-5pt}
    \label{flmedsegbench}
\end{figure*}

To address these gaps, we introduce a comprehensive benchmark for federated medical image segmentation named \textbf{FL-MedSegBench}. As shown in Fig.~\ref{flmedsegbench}, its data comes from \textbf{27 datasets} and covers \textbf{nine medical segmentation tasks}
across \textbf{ten imaging modalities} and both \textbf{2D and 3D formats}. To faithfully reflect real-world scenarios, each task incorporates data acquired from different institutions or imaging devices. These tasks include vessel segmentation with fundus images, meibomian gland segmentation with infrared images, adenocarcinoma detection with pathology images, polyp segmentation with endoscopic images, prostate segmentation with MRI slices, breast lesion segmentation with ultrasound images, right ventricle segmentation with CMR volumes, pancreas lesion segmentation with MRI-T1 scans and brain tumor segmentation using multi-modal MRI (four modalities). 
We evaluate \textbf{eight gFL} and \textbf{five pFL methods} under realistic non-IID data. Our benchmark assesses not only segmentation accuracy but also fairness, communication efficiency, convergence behavior, and generalization to unseen clients. Through extensive experiments, we highlight the strengths and limitations of existing approaches and identify promising directions for future research. We hope this work will facilitate the development of more robust, efficient, and clinically applicable FL solutions for medical imaging. Our main contributions are listed as follows: 
\begin{enumerate}
\item \textbf{A comprehensive and curated collection of medical segmentation datasets for FL research}. We collect data from 27 datasets covering nine medical segmentation tasks spanning ten imaging modalities and covering both 2D and 3D formats. To facilitate realistic FL evaluation, we provide standardized data partitions with realistic non-IID distributions that mimic real-world clinical heterogeneity across institutions or devices. 
\item \textbf{The first systematic benchmark for gFL and pFL in medical image segmentation}. We evaluate eight gFL methods (\textit{e.g.}, FedAvg and FedProx) and five pFL methods (\textit{e.g.}, FedBN and Ditto) across all nine datasets. The benchmark supports diverse evaluation dimensions including segmentation accuracy, fairness across clients, communication efficiency, convergence behavior, and generalization to unseen clients, providing a holistic view of method performance.
\item \textbf{Extensive experimental analysis under realistic FL scenarios}. We conduct rigorous experiments across nine medical segmentation tasks. Our analysis reveals several key insights into method performance, communication-computation trade-offs, fairness characteristics, and out-of-distribution generalization vulnerabilities. These findings provide empirically grounded guidance for selecting and deploying FL methods in real-world medical scenes.
\item \textbf{An open-source, modular, and extensible benchmark toolkit}. We release a fully reproducible codebase built on a modular FL framework, enabling easy integration of new datasets, models, and algorithms. The toolkit includes containerized environments, hyperparameter search scripts, and evaluation pipelines to ensure reproducibility and facilitate future research in federated medical image analysis.
\end{enumerate}
\textbf{Roadmap.} The rest of the paper is organized as follows. In Section \ref{sec:background}, we will introduce two learning
paradigms: generic federated learning (gFL) and personalized
federated learning (pFL). Section \ref{sec:MedSegBench} details the proposed FL-MedSegBench. We present the segmentation performance, convergence behavior and generalization performance in Section \ref{sec:experiments}. 
We discuss the choices made as well as the limitations of the benchmark in Section \ref{sec:discussion}.
Finally, the paper is closed with the conclusion in Section \ref{sec:conclusion}.

\section{Background and Problem Formulation}
\label{sec:background}

This section introduces the fundamental concepts of federated learning (FL) in medical image analysis, with a focus on image segmentation tasks. We formally review two learning paradigms: generic federated learning (gFL) and personalized federated learning (pFL).

\subsection{Generic Federated Learning}
\label{subsec:gfl}
Generic Federated Learning (gFL) aims to train a single global model that generalizes effectively on the collective data distribution across all participating clients. In the context of medical image segmentation, this paradigm assumes that data originating from different hospitals or institutions, despite exhibiting variability due to differences in imaging protocols, acquisition devices, or patient demographics, share consistent underlying anatomical and pathological characteristics.

Formally, let $\mathcal{D} = \bigcup_{k=1}^K \mathcal{D}_k$ represent the union of all local datasets from $K$ clients (\textit{e.g.}, clinical sites), where $\mathcal{D}_k = \{(x_i^k, y_i^k)\}_{i=1}^{N_k}$ denotes the local dataset of the client $k$ with $N_k$ samples. Here, $x_i^k \in \mathbb{R}^{H \times W \times D}$ is a training sample (\textit{e.g.}, a 2D ultrasound image or 3D MRI scan), and $y_i^k \in \{0,1\}^{H \times W \times C}$ is the corresponding segmentation mask with $C$ anatomical structures of interest.
The objective of the gFL paradigm is to find a global segmentation model parameterized by $W_g$ that minimizes the aggregate empirical risk:
\vspace{-3mm}
\begin{equation}
\min_{W_g} \mathcal{F}(W_g) = \sum_{k=1}^K \alpha_k \mathcal{L}_k(W_g),
\label{eq:gfl_objective}
\vspace{-3mm}
\end{equation}
where $\alpha_k$ is the aggregation weight of the $k$-th client.  $\mathcal{L}_k(W_g)$ is the local loss function for the client $k$, typically defined for segmentation tasks as:
\vspace{-3mm}
\begin{equation}
\mathcal{L}_k(W_g) = \frac{1}{N_k} \sum_{i=1}^{N_k} \ell_{\text{seg}}\big(f(x_i^k; W_g), y_i^k\big).
\label{eq:seg_loss}
\vspace{-3mm}
\end{equation}
Here, $f(\cdot; W_g)$ denotes the segmentation model (\textit{e.g.}, U-Net), and $\ell_{\text{seg}}$ combines per-pixel losses such as Dice loss and binary cross-entropy (BCE) loss:
\vspace{-3mm}
\begin{equation}
\ell_{\text{seg}}(\hat{y}, y) = \lambda \cdot \ell_{\text{Dice}}(\hat{y}, y) + (1-\lambda) \cdot \ell_{\text{BCE}}(\hat{y}, y),
\label{eq:dice_ce_loss}
\vspace{-2mm}
\end{equation}
where $\hat{y}$ is the predicted segmentation map, and $\lambda \in [0,1]$ balances the two loss terms.

To optimize the objective in Eq.~\eqref{eq:gfl_objective}, the standard gFL method, FedAvg~\cite{fedavg}, performs an iterative training procedure through multiple rounds of communication between the clients and a central server. In each communication round $t \in [T]$, the server first broadcasts the current global model parameters $W_g^{(t)}$ to all participating clients. Subsequently, each client performs $E$ local epochs to minimize its empirical loss function $\mathcal{L}_k$ using its private data $\mathcal{D}_k$. Upon completion of local updates, the refined model $W_k^{(t)}$ is transmitted back to the server, which aggregates them via a weighted averaging scheme:
\vspace{-3mm}
\begin{equation}
W_g^{(t+1)} = \sum_{k=1}^K \alpha_k W_k^{(t)},\quad \mathrm{where}\quad \alpha_k = \frac{N_k}{\sum_{k=1}^KN_k}.
\label{eq:fedavg}
\vspace{-3mm}
\end{equation}
While FedAvg aims to produce a consensus model that performs uniformly across clients, its efficacy can degrade when local data distributions $\mathcal{D}_k$ exhibit heterogeneity (\textit{i.e.}, are non-IID), a prevalent challenge in medical imaging due to inter-institutional variability in acquisition protocols and patient populations. Many works have been proposed to improve FedAvg by introducing a regularization term to local training to align client objectives~\cite{li2020federated, li2021model}, designing adaptive aggregation strategies~\cite{shi2025fedawa, wang2020tackling, shi2025fedlws, zhang2025pathfl}, and reducing domain shift with data augmentation~\cite{yan2025simple}.

\subsection{Personalized Federated Learning}
\label{subsec:pfl}
Personalized Federated Learning (pFL) addresses data heterogeneity by learning client-specific models $\{ W_k\}_{k=1}^K$ that are tailored to local data distributions while benefiting from collaborative training. In medical image segmentation, pFL is crucial when anatomical appearances, pathological patterns, or imaging protocols differ across institutions~\cite{zhu2025fedbm,zhu2024stealing}.

The pFL objective is to balance local model adaptation and global knowledge sharing. Considering the absence of a universally agreed-upon objective function, we review existing methods and categorize their optimization objectives into two principal formulations~\cite{arivazhagan2019federated, chen2021bridging,li2021ditto,andreux2020siloed,li2021fedbn, chen2021personalized, chen2022personalized,t2020personalized}. The first is defined as follows:
\vspace{-3mm}
\begin{equation}
\min_{\{W_1,...,W_K\}} \sum_{k=1}^K \alpha_k\mathcal{L}_k(W_k) + \frac{\mu}{2} \mathcal{R}(W_g, W_1,...,W_K),
\label{eq:pfl_objective}
\vspace{-3mm}
\end{equation}
where $\mathcal{R}(\cdot, \cdot)$ is a regularization term that encourages proximity between the local model $W_k$ and a global reference model $W_g$, and $\mu > 0$ controls the trade-off. A common choice of $\mathcal{R}$ is applying an $L_2$ regularizer between $W_k$ and $W_g$. The corresponding local
training step is generally formulated as
\vspace{-3mm}
\begin{equation}
\mathcal{L}_k(W_k) = \frac{1}{N_k} \sum_{i=1}^{N_k} \ell_{\text{seg}}\big(f(x_i^k; W_k), y_i^k\big) + \frac{\mu}{2}||W_k-W_g||^2_2.
\label{eq:seg_loss}
\vspace{-3mm}
\end{equation}
Here, the global model $W_g$ is updated via Eq.~\eqref{eq:fedavg}. The regularizer term prevents local models from overfitting to small datasets and deviating excessively from the global consensus.

The second type of pFL methods generally decouples the model parameters into base layers $W_\mathrm{\mathbf{B}}$ and personalized layers $W_\mathrm{\mathbf{P}}$. The base layers are jointly optimized by collaborating all clients while the personalized layers are updated locally and are not shared or aggregated with the server. The objective function of this type of pFL methods is formulated as 
\vspace{-3mm}
\begin{equation}
\min_{\{W_\mathrm{\mathbf{B}},W_{(\mathbf{P},1)},...,W_{(\mathbf{P},K)}\}} \sum_{k=1}^K \alpha_k\mathcal{L}_k(W_\mathrm{\mathbf{B}}, W_{(\mathbf{P},\ k)}).
\label{eq:pfl_objective}
\vspace{-3mm}
\end{equation}
The objective of the local training is defined as
\vspace{-3mm}
\begin{equation}
\mathcal{L}_k(W_k) = \frac{1}{N_k} \sum_{i=1}^{N_k} \ell_{\text{seg}}\big(f(x_i^k; W_\mathrm{\mathbf{B}}, W_{(\mathbf{P},\ k)}), y_i^k\big).
\vspace{-3mm}
\end{equation}
Here, the base layers $W_\mathrm{\mathbf{B}}$ are updated via Eq.~\eqref{eq:fedavg}. pFL has shown promise in medical image segmentation by improving accuracy on heterogeneous datasets while maintaining privacy. 

\begin{table*}[!t]
\centering
\small
\caption{Summary of the 2D medical datasets in FL-MedSegBench}
\renewcommand\arraystretch{1}
\setlength{\tabcolsep}{1pt}
\begin{tabular}{|>{\centering\arraybackslash}p{55pt}|p{35pt}|p{150pt}|p{55pt}|p{220pt}|}
\toprule[1pt]
\rowcolor{gray!20}
\makecell[c]{\textbf{Dataset}}  & \makecell[c]{\textbf{Index}}  & \makecell[c]{\textbf{Client Name}}  & \makecell[c]{\textbf{Sample size}}  & \makecell[c]{\textbf{Devices/Institutions}}  \\ 
\midrule[1pt]
\multirow{6}{*}{\makecell[c]{Fed-Vessel \\ (Fundus)}} & \makecell[c]{C1} & \makecell[c]{chaseDB~\cite{fraz2012ensemble}} & \makecell[c]{28}  & \makecell[c]{Nidek NM-200-D}   \\ \cline{2-5} 
                  &  \makecell[c]{C2}  & \makecell[c]{DR-Hagis~\cite{holm2017dr}}  & \makecell[c]{40}  & \makecell[c]{Canon CR DGi/Topcon TRC-NW6s/Topcon TRC-NW8}  \\ \cline{2-5} 
                  & \makecell[c]{C3}  & \makecell[c]{DRIVE~\cite{staal2004ridge}}  & \makecell[c]{40}  & \makecell[c]{Canon CR5 nonmydriatic 3CCD/TopCon TRV-50} \\ \cline{2-5} 
                  & \makecell[c]{C4}  & \makecell[c]{HRF~\cite{odstrcilik2013retinal}}  & \makecell[c]{45}  & \makecell[c]{Canon CF–60 UVi}  \\ \cline{2-5} 
                  & \makecell[c]{C5}  &\makecell[c]{LES-AV~\cite{orlando2018towards}}   & \makecell[c]{22}  & \makecell[c]{N/A}  \\ \cline{2-5} 
                  & \makecell[c]{C6}  & \makecell[c]{ORVS~\cite{sarhan2021transfer}}  & \makecell[c]{49}  & \makecell[c]{Zeiss, Viscuam 200}  \\ 
\midrule[1pt]
\multirow{6}{*}{\makecell[c]{Fed-Prostate \\ (MRI-T2)} } & \makecell[c]{C1}  & \makecell[c]{RUNMC}  &  \makecell[c]{411}  & \makecell[c]{Siemens}  \\ \cline{2-5} 
                  & \makecell[c]{C2}  & \makecell[c]{BMC}  & \makecell[c]{384}  & \makecell[c]{Philips}  \\ \cline{2-5} 
                  & \makecell[c]{C3}  & \makecell[c]{I2CVB}  & \makecell[c]{468}  & \makecell[c]{Siemens}  \\ \cline{2-5} 
                  & \makecell[c]{C4}  & \makecell[c]{UCL}  & \makecell[c]{175}  & \makecell[c]{Siemens}  \\ \cline{2-5} 
                  & \makecell[c]{C5}  & \makecell[c]{BIDMC}  & \makecell[c]{261}  & \makecell[c]{General Electric}  \\ \cline{2-5} 
                  & \makecell[c]{C6}  & \makecell[c]{HK}  & \makecell[c]{157}  & \makecell[c]{Siemens}  \\ 
\midrule[1pt]
\multirow{3}{*}{\makecell[c]{Fed-COSAS \\ (Pathology)}} & \makecell[c]{C1}  & \makecell[c]{teksqray-600p}  & \makecell[c]{60}  & \makecell[c]{TEKSQRAY SQS-600P}  \\ \cline{2-5} 
                  & \makecell[c]{C2}  & \makecell[c]{kfbio-400}  & \makecell[c]{60}  & \makecell[c]{KFBIO KF-PRO-400}  \\ \cline{2-5} 
                  & \makecell[c]{C3}  & \makecell[c]{3d-1000}  & \makecell[c]{60}  & \makecell[c]{3DHISTECH PANNORAMIC 1000}  \\                        
\midrule[1pt]
\multirow{4}{*}{\makecell[c]{Fed-BUS \\ (Ultrasound)}} & \makecell[c]{C1}  & \makecell[c]{BUID~\cite{ardakani2023open}}  & \makecell[c]{232}  & \makecell[c]{Supersonic Imagine Aixplorer Ultimate}  \\ \cline{2-5} 
                  & \makecell[c]{C2}  & \makecell[c]{BUSI~\cite{hesaraki2023bus}}  & \makecell[c]{630}  & \makecell[c]{N/A}  \\ \cline{2-5} 
                  & \makecell[c]{C3}  & \makecell[c]{BUS\_UC~\cite{iqbal2024memory}}  & \makecell[c]{811}  & \makecell[c]{N/A}  \\ \cline{2-5} 
                  & \makecell[c]{C4}  & \makecell[c]{BUS\_UCLM~\cite{vallez2025bus}}  & \makecell[c]{264}  & \makecell[c]{Siemens ACUSON S2000TM Ultrasound System}  \\
                  
\midrule[1pt]
\multirow{2}{*}{\makecell[c]{Fed-MG \\ (Infrared)}} & \makecell[c]{C1}  & \makecell[c]{MGD-1k~\cite{saha2022automated}}  & \makecell[c]{1,000}  & \makecell[c]{LipiView II Ocular Surface Interferometer}  \\ \cline{2-5} 
                  & \makecell[c]{C2}  & \makecell[c]{CAMG~\cite{li2024camg}}  & \makecell[c]{100}  & \makecell[c]{Keratograph 5 M
(Oculus, Germany)}  \\ 
\midrule[1pt]
\multirow{5}{*}{\makecell[c]{Fed-Polyp \\ (Endoscopic)}} & \makecell[c]{C1}  & \makecell[c]{CVC-300~\cite{vazquez2017benchmark}}  & \makecell[c]{60}  & \makecell[c]{N/A}  \\ \cline{2-5} 
                  & \makecell[c]{C2}  & \makecell[c]{CVC-ClinicDB~\cite{bernal2015wm}}  & \makecell[c]{612}  & \makecell[c]{Hospital Clinic of Barcelona, Spain / Colonoscopy}   \\ \cline{2-5} 
                  & \makecell[c]{C3}  &\makecell[c]{CVC-ColonDB~\cite{bernal2012towards}}   & \makecell[c]{380}  & \makecell[c]{Colonoscopy}   \\ \cline{2-5} 
                  & \makecell[c]{C4}  & \makecell[c]{EndoTect~\cite{hicks2021endotect}}  & \makecell[c]{200}  & \makecell[c]{Endoscopy}  \\ \cline{2-5} 
                  & \makecell[c]{C5}  & \makecell[c]{ETISLarib-PolypDB~\cite{silva2014toward}}  & \makecell[c]{196}  & \makecell[c]{Wireless Capsule Endoscopy}  \\

\bottomrule[1pt]
\end{tabular}
\label{tab:2Ddatasets}
\end{table*}

\section{FL-MedSegBench}
\label{sec:MedSegBench}

\subsection{Datasets}
The experimental evaluation was conducted on a total of six 2D tasks (vessel, prostate, adenocarcinom, breast tumor, meibomian gland, and polyp segmentation) and three 3D tasks (pancreas, right ventricular, and brain tumor segmentation). We provide a brief description of each dataset in the FL-MedSegBench benchmark suite, which is summarized in Tables \ref{tab:2Ddatasets} and \ref{tab:3Ddatasets}. 

\begin{table*}[!t]
\centering
\small
\caption{Summary of the 3D medical datasets in FL-MedSegBench}
\renewcommand\arraystretch{1}
\setlength{\tabcolsep}{1pt}
\begin{tabular}{|>{\centering\arraybackslash}p{90pt}|>{\centering\arraybackslash}p{35pt}|>{\centering\arraybackslash}p{70pt}|>{\centering\arraybackslash}p{55pt}|>{\centering\arraybackslash}p{250pt}|}
\toprule[1pt]
\rowcolor{gray!20}
\makecell[c]{\textbf{Dataset}}  & \makecell[c]{\textbf{Index}}  & \makecell[c]{\textbf{Client name}}  & \makecell[c]{\textbf{Sample size}}  & \makecell[c]{\textbf{Devices/Institutions}}  \\ \hline
\multirow{5}{*}{\makecell[c]{Fed-Pancreas \\ (MRI-T1)} } & \makecell[c]{C1}  & \makecell[c]{NYU}  & \makecell[c]{162}  & \makecell[c]{New York University Medical Center}  \\ \cline{2-5} 
                  & \makecell[c]{C2}  & \makecell[c]{MCF}  & \makecell[c]{151}  & \makecell[c]{Mayo Clinic Florida}  \\ \cline{2-5} 
                  & \makecell[c]{C3}  & \makecell[c]{NU}  & \makecell[c]{30}  & \makecell[c]{Northwestern University}  \\ \cline{2-5} 
                  & \makecell[c]{C4}  & \makecell[c]{AHN}  & \makecell[c]{17}  & \makecell[c]{Allegheny Health Network}  \\ \cline{2-5}  
                  & \makecell[c]{C5}  & \makecell[c]{MCA}  & \makecell[c]{25}  & \makecell[c]{Mayo Clinic Arizona}  \\ 
\midrule[1pt]
\multirow{5}{*}{\makecell[c]{Fed-M\&Ms \\ (CMR) }} 
                  & \makecell[c]{C1}  & \makecell[c]{Centre 1}  & \makecell[c]{95}  & \makecell[c]{Vall d'Hebron Hospital, Spain / Siemens}  \\ \cline{2-5} 
                  
                  & \makecell[c]{C2}  & \makecell[c]{Centre 2}  & \makecell[c]{74}  & \makecell[c]{Sagrada Familia Clínica, Spain / Philips}  \\ \cline{2-5} 
                  & \makecell[c]{C3}  & \makecell[c]{Centre 3}  & \makecell[c]{51}  & \makecell[c]{Universitätsklinikum Hamburg-Eppendorf, Germany / Philips}  \\ \cline{2-5} 
                  & \makecell[c]{C4}  & \makecell[c]{Centre 4}  & \makecell[c]{50}  & \makecell[c]{Universitari Dexeus Hospital, Spain / General Electric}  \\ \cline{2-5} 
                  & \makecell[c]{C5}  & \makecell[c]{Centre 5}  & \makecell[c]{50}  & \makecell[c]{Creu Blanca Clínica, Spain / Canon}  \\
\midrule[1pt]
\multirow{5}{*}{\makecell[c]{FeTS2022 \\ (MRI-T1 / T1-Gd / \\ MRI-T2 / T2-FLAIR) }} 
                  & \makecell[c]{C1}  & \makecell[c]{Centre 1}  & \makecell[c]{170}  & \makecell[c]{N/A}  \\ \cline{2-5} 
                  & \makecell[c]{C2}  & \makecell[c]{Centre 2}  & \makecell[c]{170}  & \makecell[c]{N/A}  \\ \cline{2-5}  
                  & \makecell[c]{C3}  & \makecell[c]{Centre 3}  & \makecell[c]{171}  & \makecell[c]{N/A}  \\ \cline{2-5} 
                  & \makecell[c]{C4}  & \makecell[c]{Centre 4}  & \makecell[c]{34}  & \makecell[c]{N/A}  \\ \cline{2-5} 
                  & \makecell[c]{C5}  & \makecell[c]{Centre 5}  & \makecell[c]{30}  & \makecell[c]{N/A}  \\ \cline{2-5}
                  & \makecell[c]{C6}  & \makecell[c]{Centre 6}  & \makecell[c]{127}  & \makecell[c]{N/A}  \\ \cline{2-5}
                  & \makecell[c]{C7}  & \makecell[c]{Centre 7}  & \makecell[c]{127}  & \makecell[c]{N/A}  \\ \cline{2-5}
                  & \makecell[c]{C8}  & \makecell[c]{Centre 8}  & \makecell[c]{128}  & \makecell[c]{N/A} \\ \cline{2-5}
                  & \makecell[c]{C9}  & \makecell[c]{Centre 9}  & \makecell[c]{33}  & \makecell[c]{N/A}  \\            
\bottomrule[1pt]
\end{tabular}
\label{tab:3Ddatasets}
\end{table*}

\noindent\textbf{Fed-Vessel.} 
This dataset is constructed for the task of retinal blood vessel segmentation with 2D fundus images, comprising six client data sources: chaseDB~\cite{fraz2012ensemble}, DR-Hagis~\cite{holm2017dr}, DRIVE~\cite{staal2004ridge}, HRF~\cite{odstrcilik2013retinal}, LES-AV~\cite{orlando2018towards}, and ORVS~\cite{sarhan2021transfer}. chaseDB contains 28 images captured with Nidek NM-200-D fundus cameras; DR-Hagis includes 40 images acquired using three different devices, \textit{i.e.}, the Canon CR DGi 
, the Topcon TRC-NW6s, and the Topcon TRC-NW8 
; DRIVE consists of 40 images obtained with Canon CR5 nonmydriatic 3CCD cameras and TopCon TRV-50 fundus cameras; HRF contains 45 images captured by the Canon CF–60 UVi
; LES-AV includes 22 images; ORVS comprises 49 images collected using the Zeiss Visucam 200. 
All clients provide predefined training and test splits. We further split their training sets into training and validation subsets with a ratio of 4:1. Images are resized to $256\times256$. 

\noindent\textbf{Fed-Prostate.}
This dataset~\cite{liu2021feddg} is constructed for the prostate segmentation task with 2D MRI slices, comprising six client data sources: RUNMC, BMC, I2CVB, UCL, BIDMC, and HK. RUNMC contains 411 images from 30 patients acquired with Siemens scanners; BMC includes 384 images from 30 patients obtained using Philips equipment; I2CVB consists of 468 images from 19 patients captured with Siemens devices; UCL contains 175 images from 13 patients from Siemens scanners; BIDMC includes 261 images from 12 patients collected with General Electric systems; and HK comprises 157 images from 12 patients also acquired using Siemens scanners. We randomly split the patients of each client into training, validation, and test sets with a ratio of 7:1:2 and crop the center region of each image with the size $256\times256$.

\noindent\textbf{Fed-COSAS.}
Fed-COSAS~\cite{cosas2024} is built for adenocarcinoma segmentation with 2D pathology slides. The dataset comprises image patches extracted from whole slide image scans of invasive breast carcinoma tissues, acquired with 3 different scanners (TEKSQRAY SQS-600P, KFBIO KF-PRO-400 and 3DHISTECH PANNORAMIC 1000). Each scanner produces 60 images.
We randomly split the data of per client into training, validation, and test sets with a ratio of 7:1:2. All images are resized to $256\times256$.

\noindent\textbf{Fed-BUS.}
This dataset is constructed for the task of breast tumor segmentation with 2D ultrasound images, comprising four client data sources: BUID~\cite{ardakani2023open}, BUSI~\cite{hesaraki2023bus}, BUS\_UC~\cite{iqbal2024memory}, and BUS\_UCLM~\cite{vallez2025bus}. BUID contains 232 images acquired using AirProrer Ultimate and Supersonic Imagine ultrasound machines; BUSI includes 630 images, for which the acquisition device is not specified; BUS\_UC consists of 811 images, also without device information provided; and BUS\_UCLM comprises 264 images captured with the Siemens ACUSON S2000TM Ultrasound System.
The data of per client are splitted into training, validation, and test sets with a ratio of 7:1:2. All images are resized to $256\times256$.

\noindent\textbf{Fed-MG.}
This dataset is constructed for the task of meibomian gland segmentation with 2D infrared images, comprising two client data sources: MGD-1k~\cite{saha2022automated} and CAMG~\cite{li2024camg}. MGD-1k contains 1,000 images acquired using the LipiView II Ocular Surface Interferometer, while CAMG consists of 100 images, for which specific acquisition device information is not specified.
We randomly split the data of each client into training, validation, and test sets with a ratio of 7:1:2. Images are resized to $192\times384$.

\noindent\textbf{Fed-Polyp.}
This dataset is constructed for polyp segmentation task with 2D endoscopic images, comprising five client data sources: CVC-300~\cite{vazquez2017benchmark} with 60 images, CVC-ClinicDB~\cite{bernal2015wm} with 612 images, CVC-ColonDB~\cite{bernal2012towards} with 380 images, EndoTect~\cite{hicks2021endotect} with 200 images, and ETISLarib-PolypDB~\cite{silva2014toward} with 196 images. Among these, both CVC-ClinicDB and CVC-ColonDB were captured using colonoscopes, while the specific acquisition devices for the remaining datasets are not explicitly provided.
The data of each client are randomly divided into training/validation/test sets with a ratio of 7:1:2. Images are resized to $256\times256$.

\noindent\textbf{Fed-Pancreas.}
This dataset~\cite{zhang2025large} is constructed for the pancreas segmentation task with 3D MRI T1 volumes, comprising five client data sources: New York University (NYU) Medical Center with 162 scans, Mayo Clinic Florida (MCF) with 151 scans, Northwestern University (NU) with 30 scans, Allegheny Health Network (AHN) with 17 scans, and Mayo Clinic Arizona (MCA) with 25 scans. The specific medical imaging devices used for acquisition are not detailed for any of the contributing clients in this federated collection. We divide the data of each client into training/validation/test sets with a ratio of 7:1:2 and randomly sample  patches with the size $32\times256\times256$ as input of model.

\noindent\textbf{Fed-M\&Ms.}
This dataset~\cite{martin2023deep} is built for right ventricular segmentation task with 3D MRI volumes, comprising 5 center data sources including four hospitals from Spain and one hospital from Germany. The Vall d'Hebron Hospital contains 95 volumes acquired using Siemens scanners; the Sagrada Familia Hospital includes 74 images obtained with Philips systems;  Universitätsklinikum Hamburg-Eppendorf provides 51 volumes from Philips scanners; Both Universitari Dexeus Hospital and Creu Blanca Clínica contain 50 volumes captured by General Electric and Canon imaging systems, respectively. We divide data of per client into training/validation/test sets with a ratio of 7:1:2 and randomly sample  patches with the size $16\times256\times256$ as input of model.

\noindent\textbf{FeTS2022.}
This dataset is from the Federated Tumor Segmentation (FeTS) 2022 challenge~\cite{spyridon_bakas_2022_6362409}, which mainly focuses on the construction and evaluation of a consensus model for the brain tumor segmentation task. The dataset contains 1251 patients from 33 institutions. The data of each patient is acquired by multi-parametric magnetic resonance imaging (mpMRI) and has four modalities: T1-Weighted MRI (MRI-T1), post-contrast T1-weighted MRI (MRI-T1Gd), T2-Weighted MRI (MRI-T2), and T2 Fluid Attenuated Inversion Recovery (T2-FLAIR). Besides normal regions, annotations comprise the GD-enhancing tumor (ET), the peritumoral edematous/invaded tissue (ED), and the necrotic tumor core (NCR). We selected the institutions with more than 30 patients, forming 9 clients.  We divide the data of each client into training/validation/test sets with a ratio of 7:1:2 and center-crop per modality and obtain patches with size $4 \times 128\times128\times128$ as input to the model.



\begin{table*}[!t]
\center
\renewcommand\arraystretch{1}
\setlength{\tabcolsep}{5pt}
\caption{The segmentation performance (Dice) of different methods on Fed-Vessel dataset (\%).}
\label{Fundus_Table}
\begin{tabular}{p{50pt}|p{57pt}|p{57pt}|p{57pt}|p{57pt}|p{57pt}|p{57pt}}
\toprule[1pt]
\rowcolor{gray!20}
Methods & \makecell[c]{C1}  & \makecell[c]{C2} & \makecell[c]{C3}& \makecell[c]{C4}  & \makecell[c]{C5} & \makecell[c]{C6} \\
\hline
Local & \makecell[c]{$77.96\pm0.76$} & \makecell[c]{$70.43\pm0.24$} & \makecell[c]{$77.00\pm0.55$} & \makecell[c]{$74.75\pm0.25$} & \makecell[c]{$78.41\pm0.55$} & \makecell[c]{$71.34\pm0.39$} \\
\hline
\multicolumn{7}{l}{\textit{Generic FL methods}}  \\
\hline
FedAvg & \makecell[c]{$74.97\pm0.90$} & \makecell[c]{$68.50\pm0.32$} & \makecell[c]{$70.43\pm0.70$} & \makecell[c]{$74.98\pm0.27$} & \makecell[c]{$78.13\pm0.08$} & \makecell[c]{$71.02\pm0.19$} \\
\hline
FedAWA & \makecell[c]{$77.12\pm0.57$} & \makecell[c]{$68.45\pm0.65$} & \makecell[c]{$72.45\pm0.61$} & \makecell[c]{$75.17\pm0.31$} & \makecell[c]{$79.17\pm0.19$} & \makecell[c]{$70.41\pm0.33$} \\
\hline
FedLWS & \makecell[c]{ $\textbf{79.96}\pm0.37$} & \makecell[c]{ $\textbf{70.24}\pm0.32$} & \makecell[c]{ $\textbf{74.29}\pm0.67$} & \makecell[c]{ $\textbf{76.25}\pm0.16$} & \makecell[c]{ $\textbf{80.35}\pm0.21$} & \makecell[c]{$70.61\pm0.57$} \\
\hline
FedNova & \makecell[c]{$77.23\pm0.93$} & \makecell[c]{$68.86\pm0.25$} & \makecell[c]{$72.65\pm0.36$} & \makecell[c]{$75.38\pm0.38$} & \makecell[c]{$79.34\pm0.37$} & \makecell[c]{$70.46\pm0.11$} \\
\hline
FedProx & \makecell[c]{$75.08\pm1.64$} & \makecell[c]{$68.56\pm0.36$} & \makecell[c]{$70.83\pm0.20$} & \makecell[c]{$75.05\pm0.30$} & \makecell[c]{$78.07\pm0.41$} & \makecell[c]{$70.87\pm0.54$} \\
\hline
FedRDN & \makecell[c]{$75.68\pm0.74$} & \makecell[c]{$67.81\pm0.69$} & \makecell[c]{$69.87\pm0.47$} & \makecell[c]{$75.24\pm0.16$} & \makecell[c]{$78.43\pm0.50$} & \makecell[c]{ $\textbf{71.29}\pm0.18$} \\
\hline
MOON & \makecell[c]{$75.48\pm1.14$} & \makecell[c]{$68.79\pm0.47$} & \makecell[c]{$70.76\pm0.30$} & \makecell[c]{$75.08\pm0.38$} & \makecell[c]{$78.59\pm0.21$} & \makecell[c]{$70.71\pm0.38$} \\
\hline
PN & \makecell[c]{$70.83\pm2.10$} & \makecell[c]{$57.94\pm3.89$} & \makecell[c]{$69.97\pm1.09$} & \makecell[c]{$67.76\pm1.78$} & \makecell[c]{$71.72\pm1.94$} & \makecell[c]{$65.96\pm1.45$} \\
\hline
\multicolumn{7}{l}{\textit{Personalized FL methods}}  \\
\hline
Ditto  & \makecell[c]{$66.22\pm1.96$}  & \makecell[c]{$62.74\pm0.54$}  & \makecell[c]{$71.71\pm0.67$}  & \makecell[c]{$70.69\pm0.38$} & \makecell[c]{$55.79\pm9.73$} & \makecell[c]{$67.38\pm0.26$} \\
\hline
FedBN  & \makecell[c]{ $\textbf{79.67}\pm0.64$}  & \makecell[c]{ $\textbf{69.62}\pm0.30$}  & \makecell[c]{ $\textbf{76.99}\pm0.34$}  & \makecell[c]{ $\textbf{75.44}\pm0.50$} & \makecell[c]{ $\textbf{80.39}\pm0.13$} & \makecell[c]{ $\textbf{72.20}\pm0.08$} \\
\hline
FedPer & \makecell[c]{$77.89\pm1.37$}  & \makecell[c]{$68.65\pm0.31$}  & \makecell[c]{$74.84\pm0.34$}  & \makecell[c]{$75.08\pm0.33$} & \makecell[c]{$79.56\pm0.24$} & \makecell[c]{$71.68\pm0.26$} \\
\hline
FedRoD & \makecell[c]{$40.64\pm12.90$} & \makecell[c]{$36.99\pm25.09$} & \makecell[c]{$56.85\pm11.09$} & \makecell[c]{$65.54\pm6.07$} & \makecell[c]{$53.83\pm9.83$} & \makecell[c]{$53.34\pm9.96$} \\
\hline
SioBN  & \makecell[c]{$78.64\pm0.82$}  & \makecell[c]{$69.08\pm0.38$}  & \makecell[c]{$75.70\pm0.50$}  & \makecell[c]{$74.92\pm0.23$} & \makecell[c]{$80.05\pm0.17$} & \makecell[c]{$72.09\pm0.34$} \\

\bottomrule[1pt]
\end{tabular}
\end{table*}

\subsection{Methods}
In FL-MedSegBench, we conduct extensive comparisons of various FL methods, encompassing eight gFL methods (FedAvg~\cite{fedavg}, FedProx~\cite{li2020federated}, MOON~\cite{li2021model}, FedAWA~\cite{shi2025fedawa}, FedNova~\cite{wang2020tackling}, FedRDN~\cite{yan2025simple}, PN~\cite{wang2025population} and FedLWS~\cite{shi2025fedlws}) and five pFL methods (FedRoD~\cite{chen2021bridging}, SioBN~\cite{andreux2020siloed}, FedBN~\cite{li2021fedbn}, FedPer~\cite{arivazhagan2019federated} and Ditto~\cite{li2021ditto}).

\noindent\textbf{FedAvg.} 
FedAvg~\cite{fedavg} serves as the simplest FL algorithm, operating through iterative rounds of local stochastic gradient descent updates on clients followed by a simple averaging of model parameters on a central server. 
FedAvg's performance is known to degrade under significant data heterogeneity, especially with many local update steps \cite{li2020federated}.

\noindent\textbf{FedProx.}
FedProx~\cite{li2020federated} modifies FedAvg by incorporating a proximal term into the local objective function, with the aim of improving stability under statistical heterogeneity. This term explicitly regularizes the distance between the local and global models, effectively mitigating client drift during local training.

\noindent\textbf{MOON.}
MOON~\cite{li2021model} addresses client data heterogeneity through model-level contrastive learning. It corrects local updates by maximizing the agreement between the representations learned by the current local model and those from the global model, based on the observation that the global model, trained on the entire dataset, generally learns more robust representations than local models trained on skewed data subsets.

\begin{table*}[!t]
\center
\renewcommand\arraystretch{1}
\setlength{\tabcolsep}{5pt}
\caption{The segmentation performance (Dice) of different methods on Fed-Prostate dataset (\%).}
\label{Prostate_Table}
\begin{tabular}{p{50pt}|p{57pt}|p{57pt}|p{57pt}|p{57pt}|p{57pt}|p{57pt}}
\toprule[1pt]
\rowcolor{gray!20}
Methods & \makecell[c]{C1}  & \makecell[c]{C2} & \makecell[c]{C3}& \makecell[c]{C4}  & \makecell[c]{C5} & \makecell[c]{C6} \\
\hline
Local & \makecell[c]{$76.37\pm2.19$} &\makecell[c]{$86.41\pm1.94$} &\makecell[c]{$80.30\pm1.64$} &\makecell[c]{$84.06\pm2.56$} &\makecell[c]{$86.74\pm0.76$} &\makecell[c]{$62.32\pm3.33$} \\
\hline
\multicolumn{7}{l}{\textit{Generic FL methods}}  \\
\hline
FedAvg  & \makecell[c]{  $\textbf{76.35}\pm2.39$} & \makecell[c]{$86.29\pm0.84$} & \makecell[c]{$87.06\pm0.36$} & \makecell[c]{$87.15\pm2.31$} & \makecell[c]{$88.99\pm1.22$} & \makecell[c]{$78.48\pm3.51$} \\
\hline
FedAWA  & \makecell[c]{$74.84\pm7.54$} & \makecell[c]{$86.99\pm0.59$} & \makecell[c]{$87.83\pm1.30$} & \makecell[c]{$86.47\pm0.48$} & \makecell[c]{$89.11\pm0.54$} & \makecell[c]{$78.84\pm3.47$} \\
\hline
FedLWS  & \makecell[c]{$74.49\pm3.96$} & \makecell[c]{ $\textbf{88.33}\pm0.28$} & \makecell[c]{ $\textbf{89.17}\pm0.70$} & \makecell[c]{$87.71\pm1.53$} & \makecell[c]{ $\textbf{89.37}\pm0.71$} & \makecell[c]{ $\textbf{80.45}\pm1.66$} \\
\hline
FedNova & \makecell[c]{$74.94\pm8.94$} & \makecell[c]{$86.45\pm0.32$} & \makecell[c]{$87.99\pm0.82$} & \makecell[c]{ $\textbf{88.22}\pm0.97$} & \makecell[c]{$88.67\pm0.24$} & \makecell[c]{$78.03\pm3.20$} \\
\hline
FedProx & \makecell[c]{$76.00\pm2.99$} & \makecell[c]{$86.56\pm0.80$} & \makecell[c]{$87.86\pm0.58$} & \makecell[c]{$87.20\pm1.70$} & \makecell[c]{$89.33\pm0.50$} & \makecell[c]{$77.00\pm2.60$} \\
\hline
FedRDN  & \makecell[c]{$71.92\pm8.24$} & \makecell[c]{$86.84\pm0.34$} & \makecell[c]{$86.15\pm1.29$} & \makecell[c]{$88.12\pm2.18$} & \makecell[c]{$88.94\pm0.24$} & \makecell[c]{$78.58\pm2.75$} \\
\hline
MOON    & \makecell[c]{$75.26\pm3.72$} & \makecell[c]{$86.66\pm0.26$} & \makecell[c]{$86.87\pm0.40$} & \makecell[c]{$87.58\pm1.76$} & \makecell[c]{$88.63\pm0.11$} & \makecell[c]{$77.67\pm3.52$} \\
\hline
PN      & \makecell[c]{$75.24\pm6.26$} & \makecell[c]{$86.39\pm1.25$} & \makecell[c]{$87.00\pm0.53$} & \makecell[c]{$86.55\pm1.66$} & \makecell[c]{$88.96\pm0.84$} & \makecell[c]{$77.25\pm4.72$} \\
\hline
\multicolumn{7}{l}{\textit{Personalized FL methods}}  \\
\hline
Ditto  & \makecell[c]{ $\textbf{78.42}\pm0.48$} & \makecell[c]{$84.11\pm0.62$} & \makecell[c]{$83.43\pm1.48$} & \makecell[c]{$84.61\pm2.84$} & \makecell[c]{$84.88\pm0.98$} & \makecell[c]{ $\textbf{77.83}\pm2.66$} \\
\hline
FedBN  & \makecell[c]{$73.86\pm6.40$} & \makecell[c]{ $\textbf{88.31}\pm0.49$} & \makecell[c]{ $\textbf{87.68}\pm0.60$} & \makecell[c]{$87.95\pm1.95$} & \makecell[c]{$88.07\pm0.25$} & \makecell[c]{$75.44\pm3.65$} \\
\hline
FedPer & \makecell[c]{$76.64\pm1.21$} & \makecell[c]{$86.48\pm0.81$} & \makecell[c]{$86.01\pm1.28$} & \makecell[c]{ $\textbf{88.13}\pm0.47$} & \makecell[c]{ $\textbf{89.12}\pm0.37$} & \makecell[c]{$75.96\pm3.31$} \\
\hline
FedRoD & \makecell[c]{$71.02\pm3.44$} & \makecell[c]{$83.98\pm0.96$} & \makecell[c]{$85.97\pm0.61$} & \makecell[c]{$87.98\pm1.92$} & \makecell[c]{$86.62\pm2.00$} & \makecell[c]{$62.38\pm26.17$} \\
\hline
SioBN  & \makecell[c]{$78.07\pm4.36$} & \makecell[c]{$88.19\pm0.62$} & \makecell[c]{$83.24\pm5.24$} & \makecell[c]{$87.11\pm3.34$} & \makecell[c]{$88.32\pm0.48$} & \makecell[c]{$76.03\pm6.49$} \\

\bottomrule[1pt]
\end{tabular}
\end{table*}

\noindent\textbf{FedAWA.}
FedAWA~\cite{shi2025fedawa} uses an adaptive weighting mechanism during server aggregation by dynamically adjusting weights based on client vectors. These vectors capture the direction of local model updates, reflecting variations in client data distributions. This approach enhances robustness against non-IID data without requiring additional datasets or compromising privacy.

\noindent\textbf{FedNova.}
FedNova~\cite{wang2020tackling} tackles the objective inconsistency caused by heterogeneous client participation and varying numbers of local updates. It proposes a normalized averaging scheme on server, which accounts for differences in local update progress to ensure convergence to a consistent stationary point while maintaining a fast convergence rate.

\noindent\textbf{FedRDN.}
FedRDN~\cite{yan2025simple} addresses the feature shift caused by non-IID data in FL through an input-level data augmentation method. It enhances the generalization of local feature representations by randomly injecting global statistical information into the client's local data, thereby enhancing the generalization of local feature representations. 

\noindent\textbf{PN.}
PN~\cite{wang2025population} addresses the limitations of batch normalization in FL by learning the normalization statistics as trainable parameters, rather than computing them from client mini-batches. This creates homogeneous normalization layers across clients, stabilizing training under data heterogeneity and enabling effective learning even with very small batch sizes.

\noindent\textbf{FedLWS.}
FedLWS~\cite{shi2025fedlws} addresses the global weight shrinking effect in FL by adaptively designing layer-wise shrinking factors during model aggregation. It calculates these factors based on distinctions among different layers of the global model.

\noindent\textbf{FedRoD.}
FedRoD~\cite{chen2021bridging} bridges the gap between generalization and personalization by decoupling learning into two components: a global predictor trained with a consistent objective across clients, and a lightweight client-specific module optimized on top of it to minimize each client's empirical risk. This dual-predictor framework simultaneously achieves strong global performance and effective local adaptation in heterogeneous federated settings.

\noindent\textbf{SioBN.}
SioBN~\cite{andreux2020siloed} addresses multi-centric data heterogeneity in federated medical imaging by keeping batch normalization (BN) statistics local to each client (or silo), while only averaging the weight parameters globally. This approach maintains data privacy, adapts to site-specific characteristics, and reduces potential information leakage.

\noindent\textbf{FedBN.}
FedBN~\cite{li2021fedbn} alleviates feature shift in non-IID settings by localizing BN layers. While convolutional layers are aggregated centrally, each client maintains its own batch-specific statistics. This allows the model to adapt to local data distributions without sharing normalization parameters.

\noindent\textbf{FedPer.}
FedPer~\cite{arivazhagan2019federated} introduces personalized FL by splitting the model into base and personalization layers. While the base layers are trained collaboratively using federated averaging, the personalization layers are updated locally on each device, effectively capturing user-specific characteristics while maintaining a shared feature representation.

\noindent\textbf{Ditto.}
Ditto~\cite{li2021ditto} jointly learns a global model and personalized local models through a multi-task learning objective. It introduces a regularization term that penalizes the deviation of each personalized model from the global model, allowing a tunable balance between local adaptation and global consensus.

\subsection{Implementation details} We implement FL-MedSegBench based on the PyTorch platform. We use the Adam optimizer~\cite{kingma2014adam} for all datasets. The initial learning rate is selected from $\{0.01, 0.001, 0.0001\}$ via grid search. Two learning rate scheduling strategies are considered: keeping the learning rate constant or halving it every $E/4$ rounds, where $E$ denotes the total number of training rounds. In our experiments, for Fed-Vessel and Fed-MG, we set the initial learning rate to $0.001$ and apply the halving schedule every $E/4$ rounds. For the remaining datasets, a fixed learning rate of $0.0001$ is used. The batch size is set to $8$ for Fed-Vessel, Fed-Prostate, Fed-Polyp, and Fed-BUS; $4$ for Fed-MG, Fed-COSAS, and FeTS2022; and $2$ for Fed-Pancreas and Fed-M\&Ms. We set $\mu$ to 0.001 in FedProx and Ditto, 0.01 in MOON as suggested in their official implementations. We adopt the 2D U-Net~\cite{ronneberger2015u} as the backbone network of all 2D datasets and 3D U-Net for Fed-M\&Ms and FeTS2022. SA-Net~\cite{chen2025decentralized} is used for Fed-Pancreas. During training, we apply data augmentation including Gaussian noise, elastic deformation, brightness and contrast adjustments, random horizontal and vertical flips, random 90-degree rotations, and random cropping. Performance is measured by Dice coefficient, and all results are averaged over three independent trials with mean and standard deviation reported. To show degrees of fairness of different FL methods, we report the equally-weighted average of Dice scores (aDice), the worst-client Dice scores (wDice), and the standard deviation (Std) of Dice scores of clients.

\begin{table*}[!t]  
\caption{The segmentation performance (Dice) of FL methods on Fed-COSAS and Fed-BUS datasets (\%).}
\vspace{-1mm}
\centering
\renewcommand\arraystretch{1}
\setlength{\tabcolsep}{1pt}
\begin{minipage}[t]{0.37\textwidth}
\captionof{subtable}{Fed-COSAS}  
\vspace{-2mm}
\label{COSAS_Table}
\centering
\begin{tabular}{p{40pt}|p{57pt}|p{57pt}|p{57pt}}
\toprule[1pt]
\rowcolor{gray!20}
Methods & \makecell[c]{C1}  & \makecell[c]{C2} & \makecell[c]{C3} \\
\hline
Local & \makecell[c]{$71.99\pm7.65$} & \makecell[c]{$68.21\pm3.04$} & \makecell[c]{$74.15\pm9.04$} \\ 
\hline
\multicolumn{4}{l}{\textit{Generic FL methods}}  \\
\hline
FedAvg   & \makecell[c]{$68.53\pm9.44$}  & \makecell[c]{$60.14\pm5.83$} & \makecell[c]{$71.61\pm2.63$} \\
\hline
FedAWA   & \makecell[c]{$69.68\pm9.19$}  & \makecell[c]{$61.82\pm3.23$} & \makecell[c]{$70.68\pm2.93$} \\
\hline
FedLWS   & \makecell[c]{$68.54\pm8.77$}  & \makecell[c]{$57.11\pm8.56$} & \makecell[c]{$72.07\pm3.71$} \\
\hline
FedNova  & \makecell[c]{$68.53\pm9.44$}  & \makecell[c]{$60.14\pm5.83$} & \makecell[c]{$71.61\pm2.63$} \\
\hline
FedProx  & \makecell[c]{$69.10\pm8.45$}  & \makecell[c]{$59.25\pm6.37$} & \makecell[c]{$70.52\pm2.12$} \\
\hline
FedRDN   & \makecell[c]{ $\textbf{77.67}\pm4.50$}  & \makecell[c]{$67.60\pm3.19$} & \makecell[c]{$79.83\pm0.15$} \\
\hline
MOON     & \makecell[c]{$67.30\pm10.06$} & \makecell[c]{$61.29\pm7.33$} & \makecell[c]{$72.35\pm4.38$} \\
\hline
PN       & \makecell[c]{$73.96\pm3.70$}  & \makecell[c]{ $\textbf{67.79}\pm4.21$} & \makecell[c]{ $\textbf{80.40}\pm1.91$} \\
\hline
\multicolumn{4}{l}{\textit{Personalized FL methods}}  \\
\hline
Ditto  & \makecell[c]{$72.02\pm5.56$}  & \makecell[c]{$64.64\pm4.23$} & \makecell[c]{$71.01\pm1.96$} \\
\hline
FedBN  & \makecell[c]{ $\textbf{76.32}\pm4.69$}  & \makecell[c]{ $\textbf{67.64}\pm3.06$} & \makecell[c]{ $\textbf{79.21}\pm1.21$} \\
\hline
FedPer & \makecell[c]{$71.59\pm7.70$}  & \makecell[c]{$58.24\pm4.35$} & \makecell[c]{$71.43\pm2.88$} \\
\hline
FedRoD & \makecell[c]{$54.64\pm22.90$} & \makecell[c]{$50.61\pm7.26$} & \makecell[c]{$69.20\pm4.36$} \\
\hline
SioBN  & \makecell[c]{$75.92\pm5.17$}  & \makecell[c]{$67.34\pm2.58$} & \makecell[c]{$78.41\pm2.88$}  \\
\bottomrule[1pt]
\end{tabular}
\label{tab:table1}
\end{minipage}
\hfill
\begin{minipage}[t]{0.57\textwidth}
\captionof{subtable}{Fed-BUS}  
\vspace{-2mm}
\label{Ultrasound_Table}
\centering
\begin{tabular}{p{40pt}|p{57pt}|p{57pt}|p{57pt}|p{57pt}}
\toprule[1pt]
\rowcolor{gray!20}
Methods & \makecell[c]{C1}  & \makecell[c]{C2} & \makecell[c]{C3}& \makecell[c]{C4} \\
\hline
Local &\makecell[c]{$82.72\pm0.19$} & \makecell[c]{$74.58\pm1.96$} & \makecell[c]{$90.30\pm0.08$} & \makecell[c]{$74.86\pm0.91$} \\
\hline
\multicolumn{5}{l}{\textit{Generic FL methods}}  \\
\hline
FedAvg & \makecell[c]{$86.32\pm1.07$} & \makecell[c]{$73.72\pm1.06$} & \makecell[c]{$90.56\pm0.55$} & \makecell[c]{$71.90\pm2.73$} \\
\hline
FedAWA & \makecell[c]{$85.69\pm0.40$} & \makecell[c]{ $\textbf{74.89}\pm1.49$} & \makecell[c]{$89.56\pm0.46$} & \makecell[c]{$77.11\pm3.05$} \\
\hline
FedLWS & \makecell[c]{$84.63\pm0.35$} & \makecell[c]{$70.11\pm0.63$} & \makecell[c]{$87.90\pm0.23$} & \makecell[c]{$68.59\pm4.94$} \\
\hline
FedNova & \makecell[c]{ $\textbf{86.35}\pm1.14$} & \makecell[c]{$73.44\pm1.67$} & \makecell[c]{$89.55\pm0.38$} & \makecell[c]{ $\textbf{77.30}\pm4.06$} \\
\hline
FedProx & \makecell[c]{$85.68\pm1.99$} & \makecell[c]{$74.39\pm1.83$} & \makecell[c]{$90.04\pm0.58$} & \makecell[c]{$72.52\pm4.16$} \\
\hline
FedRDN & \makecell[c]{$86.30\pm0.74$} & \makecell[c]{$73.33\pm1.00$} & \makecell[c]{ $\textbf{90.78}\pm0.57$} & \makecell[c]{$71.34\pm5.33$} \\
\hline
MOON & \makecell[c]{$85.77\pm1.35$} & \makecell[c]{$73.88\pm1.89$} & \makecell[c]{$90.09\pm0.16$} & \makecell[c]{$74.57\pm1.35$} \\
\hline
PN & \makecell[c]{$85.99\pm1.45$} & \makecell[c]{$73.04\pm2.04$} & \makecell[c]{$90.56\pm0.53$} & \makecell[c]{$73.63\pm1.53$} \\
\hline
\multicolumn{5}{l}{\textit{Personalized FL methods}}  \\
\hline
Ditto  & \makecell[c]{$82.30\pm0.91$} & \makecell[c]{$74.46\pm0.33$} & \makecell[c]{$88.53\pm0.78$} & \makecell[c]{$71.64\pm0.93$} \\
\hline
FedBN  & \makecell[c]{$85.45\pm2.66$} & \makecell[c]{ $\textbf{74.52}\pm0.95$} & \makecell[c]{ $\textbf{91.04}\pm0.19$} & \makecell[c]{$70.62\pm4.60$} \\
\hline
FedPer & \makecell[c]{ $\textbf{86.44}\pm0.62$} & \makecell[c]{$73.77\pm1.62$} & \makecell[c]{$90.68\pm0.93$} & \makecell[c]{ $\textbf{73.36}\pm3.14$} \\
\hline
FedRoD & \makecell[c]{$81.86\pm2.32$} & \makecell[c]{$73.03\pm2.21$} & \makecell[c]{$89.89\pm1.26$} & \makecell[c]{$73.28\pm1.38$} \\
\hline
SioBN  & \makecell[c]{$85.00\pm1.06$} & \makecell[c]{$73.49\pm0.55$} & \makecell[c]{$90.62\pm0.42$} & \makecell[c]{$71.71\pm1.47$} \\
\bottomrule[1pt]
\end{tabular}
\label{tab:table2}
\end{minipage}
\label{tab:both_tables}
\end{table*}

\section{Experimental Results and Analysis}
\label{sec:experiments}
\subsection{Performance Analysis}
In this section, we compare the segmentation performance of various gFL and pFL methods across different medical tasks. The number of communication rounds $T$ is set to 100 for all datasets.

\subsubsection{Performance Analysis of gFL Methods}


\noindent\textbf{Fed-Vessel.} Among gFL methods, FedLWS demonstrates superior performance, achieving the highest scores in five out of six clients (C1 to C5) within its category, as shown in Table~\ref{Fundus_Table}. Its performance on C1 (79.96\%) and C5 (80.35\%) surpasses the local baseline training, suggesting its adaptive weighting mechanism is particularly effective at leveraging cross-client information without succumbing to client drift. In contrast, standard methods like FedAvg and FedProx show a noticeable performance degradation compared with local training on several clients (\textit{e.g.}, FedAvg on C3 drops to 70.43\% from 77.00\%), a classic sign of model divergence under data heterogeneity. Methods like FedAWA, FedNova, and MOON offer marginal improvements over vanilla FedAvg on some clients but fail to consistently outperform the local training across all clients. PN exhibits the weakest performance overall, with high variance and significantly lower scores (\textit{e.g.}, 57.94\% on C2), indicating its instability in this federated context.

\noindent\textbf{Fed-Prostate.} According to Table~\ref{Prostate_Table}, FedLWS emerges as the most robust and high-performing method. It achieves the best overall scores on four out of six clients, with particularly notable results on C3 (89.17\%) and C5 (89.37\%), surpassing all competitors by a clear margin. This reinforces the effectiveness of its adaptively layer-wise aggregation strategy. FedNova also demonstrates strong performance, securing the top position on C4 (88.22\%). Standard methods like FedAvg and FedProx perform competitively on several clients, often matching or slightly exceeding the local training, but they do not achieve the peak performance of FedLWS. For instance, on C3, FedAvg reaches 87.06\%, which is strong but still 2.11 percentage points behind FedLWS. Methods like FedAWA, FedRDN, MOON, and PN show comparable performance on most clients but exhibit higher variance on C1 (\textit{e.g.}, FedRDN: 71.92\%), suggesting that these methods may be less stable when learning from the specific data characteristics of this client, potentially due to its unique acquisition protocol or patient population.

\noindent\textbf{Fed-COSAS.} From Table~\ref{COSAS_Table}, FedRDN and PN emerge as the top performers. FedRDN achieves the highest score on C1 (77.67\%) and a near-best result on C3 (79.83\%). 
PN, on the other hand, secures the best overall performance on C2 (67.79\%) and C3 (80.40\%), outperforming all competitors on these clients. In contrast, standard methods like FedAvg and FedProx consistently underperform relative to the local training on C1 and C2 (\textit{e.g.}, FedAvg: 68.53\% on C1 vs. Local 71.99\%), indicating that simple averaging suffers from client drift under severe domain shifts. FedLWS, despite its success on previous datasets, shows degraded performance here (\textit{e.g.}, 57.11\% on C2), suggesting that its loss-based weighting may be sensitive to the specific heterogeneity patterns of Fed-COSAS.

\noindent\textbf{Fed-BUS.} As shown in Table~\ref{Ultrasound_Table}, FedNova and FedRDN demonstrate the consistent performance. FedNova achieves the highest score on C1 (86.35\%) and ties for the best on C4 (77.30\%), showcasing the benefit of its normalized gradient aggregation. FedRDN secures the top position on C3 (90.78\%) and performs competitively on other clients. FedAWA also stands out, achieving the best result on C2 (74.89\%) and strong performance on C4 (77.11\%), indicating its adaptive weighting mechanism effectively handles client heterogeneity. Notably, FedLWS, which excelled on previous datasets, shows degraded performance here, particularly on C2 (70.11\%) and C4 (68.59\%), suggesting that its loss-based weighting strategy may be less suited to the specific domain characteristics of ultrasound imaging. Standard methods like FedAvg and FedProx perform reasonably well on C1 and C3 but struggle on the more challenging C4, where they fall short of the local training.

\noindent\textbf{Fed-MG.} As demonstrated in Table~\ref{Gland_Table}, FedAvg achieves the highest performance on C1 (79.07\%), though this remains below the local training. FedNova secures the top generic result on C2 (69.86\%), demonstrating the benefit of its normalized aggregation scheme. Methods such as FedAWA, FedProx, and MOON perform comparably, hovering around 78$\sim$79\% on C1 and 69\% on C2, but none successfully bridge the gap to local training. FedLWS and PN exhibit noticeable performance degradation, particularly on C2 where PN drops to 65.39\%, suggesting that their respective weighting and proximal mechanisms may over-regularize in this two-client setting. This underperformance relative to the local training underscores the difficulty of extracting beneficial shared representations when only two heterogeneous clients participate.

\begin{table*}[!t]  
\caption{The segmentation performance (Dice) of FL methods on Fed-MG and Fed-Polyp datasets (\%).}  
\centering
\renewcommand\arraystretch{1}
\setlength{\tabcolsep}{1pt}
\begin{minipage}[t]{0.22\textwidth}
\centering
\captionof{subtable}{Fed-MG}  
\vspace{-2mm}
\label{Gland_Table}
\begin{tabular}{p{40pt}|p{57pt}|p{57pt}}
\toprule[1pt]
\rowcolor{gray!20}
Methods & \makecell[c]{C1}  & \makecell[c]{C2} \\
\hline
Local & \makecell[c]{$80.94\pm1.55$} & \makecell[c]{$71.28\pm2.54$} \\
\hline
\multicolumn{3}{l}{\textit{Generic FL methods}}  \\
\hline
FedAvg  & \makecell[c]{$\textbf{79.07}\pm1.71$} & \makecell[c]{$69.60\pm3.51$} \\
FedAWA  & \makecell[c]{$78.55\pm1.56$} & \makecell[c]{$69.79\pm3.58$} \\
FedLWS  & \makecell[c]{$77.11\pm0.91$} & \makecell[c]{$66.52\pm2.60$} \\
FedNova & \makecell[c]{$77.37\pm1.37$} & \makecell[c]{$\textbf{69.86}\pm3.54$} \\
FedProx & \makecell[c]{$78.95\pm2.23$} & \makecell[c]{$69.21\pm3.42$} \\
FedRDN  & \makecell[c]{$78.65\pm1.19$} & \makecell[c]{$69.48\pm3.67$} \\
MOON    & \makecell[c]{$78.90\pm1.53$} & \makecell[c]{$69.58\pm3.98$} \\
PN      & \makecell[c]{$77.35\pm1.09$} & \makecell[c]{$65.39\pm4.39$} \\
\hline
\multicolumn{3}{l}{\textit{Personalized FL methods}}  \\
\hline
Ditto  & \makecell[c]{$78.78\pm0.50$} & \makecell[c]{$68.24\pm2.42$} \\
\hline
FedBN  & \makecell[c]{ $\textbf{79.46}\pm1.91$} & \makecell[c]{$\textbf{70.54}\pm4.19$} \\
\hline
FedPer & \makecell[c]{$78.09\pm0.88$} & \makecell[c]{$68.74\pm2.66$} \\
\hline
FedRoD & \makecell[c]{$65.48\pm20.53$} & \makecell[c]{$69.48\pm3.48$} \\
\hline
SioBN  & \makecell[c]{$79.18\pm1.08$} & \makecell[c]{$64.12\pm1.19$} \\
\bottomrule[1pt]
\end{tabular}
\label{tab:table1}
\end{minipage}
\hfill
\begin{minipage}[t]{0.67\textwidth}
\captionof{subtable}{Fed-Polyp}  
\label{Polyp_Table}
\centering
\begin{tabular}{p{40pt}|p{57pt}|p{57pt}|p{57pt}|p{57pt}|p{57pt}}
\toprule[1pt]
\rowcolor{gray!20}
Methods & \makecell[c]{C1}  & \makecell[c]{C2} & \makecell[c]{C3}& \makecell[c]{C4}& \makecell[c]{C5} \\
\hline
Local & \makecell[c]{$72.84\pm6.79$} & \makecell[c]{$88.33\pm0.82$} & \makecell[c]{$80.03\pm2.23$} & \makecell[c]{$65.41\pm4.76$} & \makecell[c]{$71.59\pm0.58$} \\
\hline
\multicolumn{6}{l}{\textit{Generic FL methods}}  \\
\hline
FedAvg  & \makecell[c]{$89.99\pm1.55$} & \makecell[c]{$85.14\pm2.80$} & \makecell[c]{$78.06\pm2.34$} & \makecell[c]{$68.94\pm2.71$} & \makecell[c]{$49.68\pm4.95$} \\
FedAWA  & \makecell[c]{$90.83\pm1.77$} & \makecell[c]{$80.70\pm3.70$} & \makecell[c]{$76.42\pm2.35$} & \makecell[c]{$70.98\pm4.01$} & \makecell[c]{$53.55\pm1.86$} \\
FedLWS  & \makecell[c]{$87.38\pm6.74$} & \makecell[c]{$81.14\pm4.16$} & \makecell[c]{$69.98\pm4.12$} & \makecell[c]{ $\textbf{80.06}\pm3.84$} & \makecell[c]{$49.38\pm9.92$} \\
FedNova & \makecell[c]{ $\textbf{91.00}\pm2.09$} & \makecell[c]{$80.57\pm3.15$} & \makecell[c]{$77.46\pm3.07$} & \makecell[c]{$71.79\pm7.47$} & \makecell[c]{$58.79\pm5.15$} \\
FedProx & \makecell[c]{$90.46\pm1.58$} & \makecell[c]{$83.92\pm2.38$} & \makecell[c]{$76.40\pm2.50$} & \makecell[c]{$70.52\pm2.42$} & \makecell[c]{$47.57\pm7.24$} \\
FedRDN  & \makecell[c]{$90.29\pm2.10$} & \makecell[c]{ $\textbf{86.78}\pm2.32$} & \makecell[c]{ $\textbf{78.69}\pm3.72$} & \makecell[c]{$72.64\pm5.53$} & \makecell[c]{ $\textbf{65.85}\pm0.60$} \\
MOON    & \makecell[c]{$90.37\pm0.81$} & \makecell[c]{$84.63\pm2.78$} & \makecell[c]{$78.24\pm1.91$} & \makecell[c]{$68.99\pm4.03$} & \makecell[c]{$50.71\pm7.39$} \\
PN      & \makecell[c]{$88.45\pm2.01$} & \makecell[c]{$85.84\pm2.57$} & \makecell[c]{$77.01\pm1.74$} & \makecell[c]{$77.53\pm3.09$} & \makecell[c]{$52.99\pm8.01$} \\
\hline
\multicolumn{6}{l}{\textit{Personalized FL methods}}  \\
\hline
Ditto  & \makecell[c]{$77.32\pm4.85$} & \makecell[c]{$84.78\pm1.70$} & \makecell[c]{$75.12\pm2.42$} & \makecell[c]{$71.85\pm4.62$} & \makecell[c]{ $\textbf{68.07}\pm5.10$} \\
\hline
FedBN  & \makecell[c]{$85.87\pm4.56$} & \makecell[c]{ $\textbf{87.72}\pm1.06$} & \makecell[c]{ $\textbf{79.48}\pm1.53$} & \makecell[c]{ $\textbf{78.27}\pm3.96$} & \makecell[c]{$60.19\pm4.49$} \\
\hline
FedPer & \makecell[c]{ $\textbf{89.46}\pm2.55$} & \makecell[c]{$84.35\pm1.82$} & \makecell[c]{$77.46\pm2.77$} & \makecell[c]{$68.81\pm4.04$} & \makecell[c]{$49.52\pm6.92$} \\
\hline
FedRoD & \makecell[c]{$88.92\pm1.18$} & \makecell[c]{$83.46\pm1.16$} & \makecell[c]{$76.31\pm4.19$} & \makecell[c]{$69.44\pm4.63$} & \makecell[c]{$33.98\pm7.05$} \\
\hline
SioBN  & \makecell[c]{$88.64\pm2.07$} & \makecell[c]{$87.05\pm1.54$} & \makecell[c]{$77.43\pm1.38$} & \makecell[c]{$74.67\pm5.77$} & \makecell[c]{$62.76\pm1.73$} \\

\bottomrule[1pt]
\end{tabular}
\label{tab:table2}
\end{minipage}
\label{tab:both_tables}
\end{table*}

\noindent\textbf{Fed-Polyp.} As shown in Table~\ref{Polyp_Table}, FedNova achieves the highest score on C1 (91.00\%) and remains competitive on other clients. FedRDN delivers the most consistent performance across clients: it secures the top generic result on C2 (86.78\%) and C3 (78.69\%), and achieves the highest Dice on C5 (65.85\%) among generic methods—a critical advantage on this challenging client. FedLWS stands out on C4, reaching 80.06\%, the highest among all methods. FedAvg, FedAWA, FedProx, and MOON perform well on C1–C3 but suffer catastrophic degradation on C5 (\textit{e.g.}, FedAvg: 49.68\%), highlighting the vulnerability of standard aggregation to outlier clients.

\noindent\textbf{Fed-Pancreas.} In Table~\ref{Pancreas_Table}, the local training exhibits severe performance degradation on clients with limited data (\textit{e.g.}, 22.58\% on C1 and 16.23\% on C3), underscoring the necessity of collaborative FL to improve generalization. Among generic FL methods, FedRDN achieves the highest performance, securing the best results on three clients (C2: 74.37\%, C4: 73.88\%, C5: 74.99\%) and competitive scores on the remaining ones. FedProx excels on C1 (61.23\%), while FedNova obtains the top score on C3 (41.83\%). These results suggest that different aggregation or regularization strategies cater to different types of client heterogeneity. Notably, FedRDN's consistent superiority may be attributed to its robustness in handling domain shifts.

\noindent\textbf{Fed-M\&Ms.}
As shown in Table~\ref{MMS_Table}, on the M\&Ms dataset, all generic FL methods except PN maintain relatively high performance across the five clients, without clear performance separation. FedRDN achieves the highest result on C1 with 82.33\%, FedAWA reaches 85.66\% on C2, FedAvg attains the best performance on C3 with 87.85\%, and FedNova performs best on C4 with 81.78\%. On C5, FedAWA achieves the highest score of 79.76\%. Overall, different methods show client-specific advantages, and no single approach consistently dominates across all clients. In contrast, PN yields substantially lower performance on all clients, indicating a noticeable performance gap.

\noindent\textbf{FeTS2022.} Table~\ref{FeTS_Table} presents the experimental results on the FeTS2022 dataset. FedProx, FedRDN, and FedNova demonstrate robust performance across clients. FedProx achieves the highest generic scores on C1 (78.63\%), C2 (88.83\%), and C7 (86.01\%), showcasing the effectiveness of its proximal term in stabilizing training under heterogeneity. FedRDN secures the top generic results on C3 (88.77\%), C6 (80.54\%), and C8 (90.64\%).
FedNova leads on C4 (83.17\%) and C9 (88.79\%), benefiting from its normalized gradient aggregation. FedAvg and FedAWA also perform competitively, often ranking among the top two or three on multiple clients. In contrast, MOON and PN exhibit consistent underperformance: MOON lags behind on most clients (\textit{e.g.}, C4: 76.18\% vs. best 83.17\%), while PN collapses catastrophically, with scores dropping to 66$\sim$78\%, far below even the local training. This reinforces the unsuitability of contrastive or proximal regularization for dense tasks.

\begin{table}[!t]
\center
\renewcommand\arraystretch{1}
\setlength{\tabcolsep}{1pt}
\caption{The performance (Dice) of different methods on Fed-Pancreas dataset.}
\label{Pancreas_Table}
\resizebox{0.495\textwidth}{!}{
\begin{tabular}{p{36pt}|p{44pt}|p{44pt}|p{44pt}|p{44pt}|p{44pt}}
\toprule[1pt]
\rowcolor{gray!20}
Methods & \makecell[c]{C1}  & \makecell[c]{C2} & \makecell[c]{C3}& \makecell[c]{C4}  & \makecell[c]{C5}\\
\hline
Local &\makecell[c]{$22.58_{\pm11.84}$} & \makecell[c]{$68.14_{\pm1.78}$} & \makecell[c]{$16.23_{\pm8.95}$} & \makecell[c]{$58.57_{\pm5.38}$} & \makecell[c]{$72.00_{\pm2.24}$} \\
\hline
\multicolumn{6}{l}{\textit{Generic FL methods}}  \\
\hline
FedAvg  &\makecell[c]{$54.71_{\pm19.44}$} &\makecell[c]{$72.11_{\pm1.52}$} &\makecell[c]{$33.97_{\pm6.16}$} &\makecell[c]{$71.35_{\pm0.30}$} &\makecell[c]{$72.81_{\pm5.03}$} \\ \hline
FedAWA  &\makecell[c]{$55.73_{\pm17.01}$} &\makecell[c]{$71.77_{\pm1.67}$} &\makecell[c]{$36.68_{\pm7.00}$} &\makecell[c]{$71.99_{\pm1.39}$} &\makecell[c]{$70.97_{\pm4.43}$} \\ \hline
FedLWS  &\makecell[c]{$51.26_{\pm12.99}$} &\makecell[c]{$68.76_{\pm1.21}$} &\makecell[c]{$26.44_{\pm6.15}$} &\makecell[c]{$70.33_{\pm4.08}$} &\makecell[c]{$67.24_{\pm4.48}$} \\ \hline
FedNova &\makecell[c]{$53.40_{\pm15.76}$} &\makecell[c]{$72.93_{\pm1.67}$} &\makecell[c]{ $\textbf{41.83}_{\pm15.96}$} &\makecell[c]{$71.99_{\pm2.44}$} &\makecell[c]{$72.61_{\pm4.94}$} \\ \hline
FedProx &\makecell[c]{ $\textbf{61.23}_{\pm15.43}$} &\makecell[c]{$74.26_{\pm0.58}$} &\makecell[c]{$34.43_{\pm7.99}$} &\makecell[c]{$73.04_{\pm3.27}$} &\makecell[c]{$74.02_{\pm3.47}$} \\ \hline
FedRDN  &\makecell[c]{$59.94_{\pm18.49}$} &\makecell[c]{ $\textbf{74.37}_{\pm1.01}$} &\makecell[c]{$32.99_{\pm9.82}$} &\makecell[c]{ $\textbf{73.88}_{\pm5.06}$} &\makecell[c]{ $\textbf{74.99}_{\pm3.39}$} \\ \hline
MOON    &\makecell[c]{$55.67_{\pm13.96}$} &\makecell[c]{$72.33_{\pm3.79}$} &\makecell[c]{$33.37_{\pm6.54}$} &\makecell[c]{$72.00_{\pm5.75}$} &\makecell[c]{$72.95_{\pm4.06}$} \\ \hline
PN      &\makecell[c]{$19.56_{\pm6.60}$} &\makecell[c]{$48.20_{\pm4.49}$} &\makecell[c]{$9.52_{\pm2.25}$}  &\makecell[c]{$38.44_{\pm10.33}$} &\makecell[c]{$41.90_{\pm1.63}$} \\ 
\hline
\multicolumn{6}{l}{\textit{Personalized FL methods}}  \\
\hline
Ditto  &\makecell[c]{$17.84_{\pm13.32}$} &\makecell[c]{$67.50_{\pm2.10}$} &\makecell[c]{$14.60_{\pm7.67}$} &\makecell[c]{$57.18_{\pm3.10}$} &\makecell[c]{$72.50_{\pm4.29}$} \\ \hline
FedBN  &\makecell[c]{$51.00_{\pm8.72}$}  &\makecell[c]{$74.21_{\pm2.34}$} &\makecell[c]{ $\textbf{34.36}_{\pm8.88}$} &\makecell[c]{$70.95_{\pm1.64}$} &\makecell[c]{ $\textbf{74.42}_{\pm5.00}$} \\ \hline
FedPer &\makecell[c]{$56.96_{\pm9.63}$} &\makecell[c]{$73.13_{\pm1.56}$} &\makecell[c]{$33.60_{\pm5.77}$} &\makecell[c]{$73.54_{\pm1.99}$} &\makecell[c]{$74.33_{\pm2.92}$} \\ \hline
FedRoD &\makecell[c]{ $\textbf{57.54}_{\pm13.93}$} &\makecell[c]{$73.20_{\pm0.29}$} &\makecell[c]{$31.48_{\pm6.81}$} &\makecell[c]{ $\textbf{73.58}_{\pm5.01}$} &\makecell[c]{$73.90_{\pm4.32}$} \\ \hline
SioBN  &\makecell[c]{$54.81_{\pm15.76}$} &\makecell[c]{ $\textbf{74.24}_{\pm0.37}$} &\makecell[c]{$33.89_{\pm6.54}$} &\makecell[c]{$73.05_{\pm5.03}$} &\makecell[c]{$73.20_{\pm3.43}$} \\ 
\bottomrule[1pt]
\end{tabular}
}
\end{table}

\subsubsection{Performance Analysis of pFL Methods}
\noindent\textbf{Fed-Vessel.} In Table~\ref{Fundus_Table}, FedBN emerges as the top-performing method across all clients (C1 to C6). Its performance on C3 (76.99\%), C5 (80.39\%), and C6 (72.20\%) not only surpasses all generic FL methods but also consistently exceeds the local training. The results validate the core premise of FedBN: learning private batch normalization statistics for each client effectively captures domain-specific characteristics. SioBN and FedPer also demonstrate strong personalization capabilities, ranking as the second and third best. SioBN, for instance, achieves competitive results on C1 (78.64\%) and C6 (72.09\%), further emphasizing the importance of personalized feature normalization layers. Conversely, Ditto and FedRoD exhibit unstable and often poor performance.

\begin{table}[!t]
\center
\renewcommand\arraystretch{1}
\setlength{\tabcolsep}{1pt}
\caption{The performance (Dice) of different methods on Fed-M\&Ms dataset.}
\label{MMS_Table}
\resizebox{0.495\textwidth}{!}{
\begin{tabular}{p{36pt}|p{44pt}|p{44pt}|p{44pt}|p{44pt}|p{44pt}}
\toprule[1pt]
\rowcolor{gray!20}
Methods & \makecell[c]{C1}  & \makecell[c]{C2} & \makecell[c]{C3}& \makecell[c]{C4}  & \makecell[c]{C5}\\
\hline
Local & \makecell[c]{$81.18_{\pm1.90}$} & \makecell[c]{$82.32_{\pm2.27}$} & \makecell[c]{$82.85_{\pm1.25}$} & \makecell[c]{$72.00_{\pm1.33}$} & \makecell[c]{$76.70_{\pm3.06}$} \\
\hline
\multicolumn{6}{l}{\textit{Generic FL methods}}  \\
\hline
FedAvg  &\makecell[c]{$80.67_{\pm0.91}$} &\makecell[c]{$85.35_{\pm0.86}$} &\makecell[c]{ $\textbf{87.85}_{\pm0.76}$} &\makecell[c]{$79.28_{\pm1.60}$} &\makecell[c]{$79.00_{\pm0.63}$} \\
\hline
FedAWA  &\makecell[c]{$78.75_{\pm1.83}$} &\makecell[c]{ $\textbf{85.66}_{\pm1.08}$} &\makecell[c]{$87.52_{\pm0.75}$} &\makecell[c]{$79.57_{\pm3.97}$} &\makecell[c]{ $\textbf{79.76}_{\pm1.41}$} \\
\hline
FedLWS  &\makecell[c]{$79.52_{\pm0.83}$} &\makecell[c]{$85.32_{\pm0.51}$} &\makecell[c]{$87.52_{\pm0.66}$} &\makecell[c]{$79.71_{\pm0.81}$} &\makecell[c]{$79.55_{\pm2.79}$} \\
\hline
FedNova &\makecell[c]{$79.10_{\pm1.30}$} &\makecell[c]{$85.39_{\pm0.65}$} &\makecell[c]{$87.44_{\pm0.96}$} &\makecell[c]{ $\textbf{81.78}_{\pm1.16}$} &\makecell[c]{$79.01_{\pm1.40}$} \\
\hline
FedProx &\makecell[c]{$79.30_{\pm2.43}$} &\makecell[c]{$85.60_{\pm1.07}$} &\makecell[c]{$87.75_{\pm0.35}$} &\makecell[c]{$80.32_{\pm1.08}$} &\makecell[c]{$79.53_{\pm2.07}$} \\
\hline
FedRDN  &\makecell[c]{ $\textbf{82.33}_{\pm0.33}$} &\makecell[c]{$85.56_{\pm0.84}$} &\makecell[c]{$87.05_{\pm1.04}$} &\makecell[c]{$80.85_{\pm1.16}$} &\makecell[c]{$76.92_{\pm1.50}$} \\
\hline
MOON    &\makecell[c]{$79.21_{\pm1.63}$} &\makecell[c]{$85.54_{\pm1.01}$} &\makecell[c]{$87.29_{\pm0.73}$} &\makecell[c]{$80.80_{\pm0.80}$} &\makecell[c]{$78.40_{\pm2.40}$} \\
\hline
PN      &\makecell[c]{$56.77_{\pm3.73}$} &\makecell[c]{$63.72_{\pm2.77}$} &\makecell[c]{$68.26_{\pm1.85}$} &\makecell[c]{$57.21_{\pm3.69}$} &\makecell[c]{$54.37_{\pm5.64}$} \\
\hline

\multicolumn{6}{l}{\textit{Personalized FL methods}}  \\
\hline
Ditto  &\makecell[c]{$81.44_{\pm0.80}$} &\makecell[c]{$82.39_{\pm1.36}$} &\makecell[c]{$79.42_{\pm2.87}$} &\makecell[c]{$75.85_{\pm1.71}$} &\makecell[c]{$74.75_{\pm4.61}$} \\ \hline
FedBN  &\makecell[c]{ $\textbf{82.84}_{\pm0.51}$} &\makecell[c]{$85.57_{\pm1.07}$} &\makecell[c]{ $\textbf{87.39}_{\pm0.67}$} &\makecell[c]{$80.93_{\pm0.31}$} &\makecell[c]{$78.81_{\pm0.79}$} \\ \hline
FedPer &\makecell[c]{$80.15_{\pm1.41}$} &\makecell[c]{$85.54_{\pm0.82}$} &\makecell[c]{$87.31_{\pm0.84}$} &\makecell[c]{$80.83_{\pm0.98}$} &\makecell[c]{ $\textbf{80.40}_{\pm1.13}$} \\ \hline
FedRoD &\makecell[c]{$78.09_{\pm1.59}$} &\makecell[c]{$84.34_{\pm1.18}$} &\makecell[c]{$87.05_{\pm0.38}$} &\makecell[c]{$80.35_{\pm0.69}$} &\makecell[c]{$78.00_{\pm1.74}$} \\ \hline
SioBN  &\makecell[c]{$82.52_{\pm0.61}$} &\makecell[c]{ $\textbf{85.67}_{\pm0.90}$} &\makecell[c]{$87.17_{\pm0.23}$} &\makecell[c]{ $\textbf{81.29}_{\pm0.74}$} &\makecell[c]{$75.77_{\pm2.62}$} \\ 

\bottomrule[1pt]
\end{tabular}
}
\end{table}

\begin{table*}[!t]
\center
\renewcommand\arraystretch{1}
\setlength{\tabcolsep}{1pt}
\caption{The segmentation performance (Dice) of different methods on FeTS2022 dataset (\%).}
\label{FeTS_Table}
\begin{tabular}{p{40pt}|p{48pt}|p{48pt}|p{48pt}|p{48pt}|p{48pt}|p{48pt}|p{48pt}|p{48pt}|p{48pt}}
\toprule[1pt]
\rowcolor{gray!20}
Methods & \makecell[c]{C1}  & \makecell[c]{C2} & \makecell[c]{C3}& \makecell[c]{C4}  & \makecell[c]{C5} & \makecell[c]{C6} & \makecell[c]{C7} & \makecell[c]{C8} & \makecell[c]{C9}\\
\hline
Local & \makecell[c]{$78.01_{\pm1.80}$} & \makecell[c]{$87.60_{\pm1.57}$} & \makecell[c]{$88.06_{\pm1.37}$} & \makecell[c]{$77.01_{\pm2.55}$} & \makecell[c]{$82.45_{\pm0.45}$} & \makecell[c]{$76.55_{\pm2.88}$} & \makecell[c]{$85.80_{\pm1.31}$} & \makecell[c]{$90.14_{\pm1.22}$} & \makecell[c]{$79.63_{\pm5.53}$} \\
\hline
\multicolumn{10}{l}{\textit{Generic FL methods}}  \\
\hline
FedAvg  &\makecell[c]{$78.40_{\pm2.86}$} &\makecell[c]{$88.19_{\pm2.13}$} &\makecell[c]{$87.90_{\pm1.35}$} &\makecell[c]{$82.43_{\pm5.70}$} &\makecell[c]{ $\textbf{88.97}_{\pm3.17}$} &\makecell[c]{$79.77_{\pm1.57}$} &\makecell[c]{$85.86_{\pm1.55}$} &\makecell[c]{$90.25_{\pm1.26}$} &\makecell[c]{$88.58_{\pm2.37}$} \\
\hline
FedAWA  &\makecell[c]{$77.60_{\pm2.26}$} &\makecell[c]{$88.21_{\pm1.06}$} &\makecell[c]{$87.53_{\pm0.90}$} &\makecell[c]{$83.04_{\pm5.08}$} &\makecell[c]{$88.32_{\pm2.70}$} &\makecell[c]{$79.71_{\pm1.44}$} &\makecell[c]{$85.02_{\pm1.55}$} &\makecell[c]{$90.43_{\pm0.60}$} &\makecell[c]{$88.62_{\pm2.62}$} \\
\hline
FedLWS  &\makecell[c]{$77.68_{\pm3.07}$} &\makecell[c]{$88.38_{\pm1.22}$} &\makecell[c]{$87.29_{\pm1.10}$} &\makecell[c]{$82.48_{\pm5.30}$} &\makecell[c]{$88.92_{\pm2.86}$} &\makecell[c]{$80.15_{\pm0.98}$} &\makecell[c]{$85.44_{\pm1.65}$} &\makecell[c]{$89.76_{\pm1.33}$} &\makecell[c]{$88.55_{\pm2.28}$} \\
\hline
FedNova &\makecell[c]{$78.11_{\pm1.85}$} &\makecell[c]{$88.33_{\pm1.03}$} &\makecell[c]{$88.06_{\pm0.87}$} &\makecell[c]{ $\textbf{83.17}_{\pm4.75}$} &\makecell[c]{$88.90_{\pm2.91}$} &\makecell[c]{$79.83_{\pm1.15}$} &\makecell[c]{$85.39_{\pm1.95}$} &\makecell[c]{$90.29_{\pm0.48}$} &\makecell[c]{ $\textbf{88.79}_{\pm2.22}$} \\
\hline
FedProx &\makecell[c]{ $\textbf{78.63}_{\pm2.62}$} &\makecell[c]{ $\textbf{88.83}_{\pm1.29}$} &\makecell[c]{$87.84_{\pm1.09}$} &\makecell[c]{$81.80_{\pm5.21}$} &\makecell[c]{$88.27_{\pm3.58}$} &\makecell[c]{$79.47_{\pm1.94}$} &\makecell[c]{ $\textbf{86.01}_{\pm1.74}$} &\makecell[c]{$90.58_{\pm0.63}$} &\makecell[c]{$88.66_{\pm2.56}$} \\
\hline
FedRDN  &\makecell[c]{$77.71_{\pm3.74}$} &\makecell[c]{$88.23_{\pm1.16}$} &\makecell[c]{ $\textbf{88.77}_{\pm1.42}$} &\makecell[c]{$81.85_{\pm6.00}$} &\makecell[c]{$88.93_{\pm3.43}$} &\makecell[c]{ $\textbf{80.54}_{\pm0.41}$} &\makecell[c]{$85.63_{\pm1.82}$} &\makecell[c]{ $\textbf{90.64}_{\pm0.90}$} &\makecell[c]{$88.62_{\pm2.68}$} \\
\hline
MOON    &\makecell[c]{$73.07_{\pm4.94}$} &\makecell[c]{$83.69_{\pm5.43}$} &\makecell[c]{$83.29_{\pm3.85}$} &\makecell[c]{$76.18_{\pm10.29}$} &\makecell[c]{$84.85_{\pm4.17}$} &\makecell[c]{$75.90_{\pm2.76}$} &\makecell[c]{$80.73_{\pm6.09}$} &\makecell[c]{$85.53_{\pm3.53}$} &\makecell[c]{$84.32_{\pm2.40}$} \\
\hline
PN      &\makecell[c]{$66.00_{\pm1.40}$} &\makecell[c]{$77.14_{\pm1.30}$} &\makecell[c]{$77.30_{\pm1.70}$} &\makecell[c]{$68.46_{\pm4.45}$} &\makecell[c]{$76.92_{\pm3.45}$} &\makecell[c]{$67.02_{\pm1.00}$} &\makecell[c]{$73.67_{\pm2.87}$} &\makecell[c]{$78.73_{\pm1.34}$} &\makecell[c]{$77.84_{\pm2.20}$} \\
\hline
\multicolumn{6}{l}{\textit{Personalized FL methods}}  \\
\hline
Ditto  &\makecell[c]{$76.92_{\pm2.16}$} &\makecell[c]{$87.79_{\pm1.04}$} &\makecell[c]{ $\textbf{88.19}_{\pm1.48}$} &\makecell[c]{$75.95_{\pm1.52}$} &\makecell[c]{$78.95_{\pm1.44}$} &\makecell[c]{$76.44_{\pm1.58}$} &\makecell[c]{$83.38_{\pm0.93}$} &\makecell[c]{$89.55_{\pm1.07}$} &\makecell[c]{$82.34_{\pm0.83}$} \\
\hline
FedBN  &\makecell[c]{ $\textbf{77.71}_{\pm2.32}$} &\makecell[c]{ $\textbf{88.78}_{\pm1.00}$} &\makecell[c]{$88.14_{\pm0.53}$} &\makecell[c]{ $\textbf{82.98}_{\pm4.76}$} &\makecell[c]{ $\textbf{89.22}_{\pm2.60}$} &\makecell[c]{ $\textbf{80.74}_{\pm1.47}$} &\makecell[c]{$86.12_{\pm1.49}$} &\makecell[c]{$89.82_{\pm1.39}$} &\makecell[c]{$87.16_{\pm3.40}$} \\
\hline
FedPer &\makecell[c]{$77.18_{\pm2.27}$} &\makecell[c]{$88.52_{\pm0.74}$} &\makecell[c]{$87.90_{\pm1.01}$} &\makecell[c]{$80.82_{\pm4.50}$} &\makecell[c]{$87.16_{\pm3.05}$} &\makecell[c]{$79.65_{\pm0.95}$} &\makecell[c]{$86.02_{\pm1.32}$} &\makecell[c]{ $\textbf{90.98}_{\pm0.97}$} &\makecell[c]{$86.92_{\pm1.65}$} \\
\hline
FedRoD &\makecell[c]{$76.36_{\pm1.59}$} &\makecell[c]{$87.79_{\pm0.40}$} &\makecell[c]{$83.86_{\pm5.03}$} &\makecell[c]{$80.28_{\pm3.08}$} &\makecell[c]{$87.70_{\pm1.77}$} &\makecell[c]{$79.70_{\pm1.03}$} &\makecell[c]{$84.64_{\pm2.77}$} &\makecell[c]{$89.86_{\pm1.14}$} &\makecell[c]{$81.20_{\pm9.04}$} \\
\hline
SioBN  &\makecell[c]{$75.79_{\pm1.64}$} &\makecell[c]{$87.86_{\pm1.00}$} &\makecell[c]{$87.10_{\pm1.39}$} &\makecell[c]{$81.46_{\pm4.58}$} &\makecell[c]{$89.13_{\pm2.79}$} &\makecell[c]{$78.95_{\pm1.24}$} &\makecell[c]{ $\textbf{86.19}_{\pm1.27}$} &\makecell[c]{$87.85_{\pm1.24}$} &\makecell[c]{ $\textbf{88.56}_{\pm2.23}$} \\
\bottomrule[1pt]
\end{tabular}
\end{table*}

\noindent\textbf{Fed-Prostate.} In Table~\ref{Prostate_Table}, Ditto stands out on C1, achieving the highest score overall (78.42\%), significantly outperforming both generic and other personalized methods on that client. This suggests that learning a personalized model while preserving a global objective is particularly beneficial for C1's data characteristics. FedBN and FedPer also deliver strong results, with FedBN achieving strong results on C2 (88.31\%) and C3 (87.68\%), and FedPer achieving top performance on C4 (88.13\%) and C5 (89.12\%). This further validates the importance of personalized feature normalization and personalized layers in handling domain shifts. However, SioBN shows high variance on some clients (\textit{e.g.}, C3: $83.24\pm5.24$, C6: $76.03\pm6.49$), indicating potential instability. FedRoD exhibits catastrophic failure on C6 (62.38\%), with extreme variance that renders its results unreliable, again underscoring its unsuitability for dense prediction tasks.

\begin{table*}[!t]
\center
\renewcommand\arraystretch{1}
\setlength{\tabcolsep}{7pt}
\caption{The generalization performance (Dice) of different methods (\%).}
\begin{tabular}{p{50pt}|p{57pt}|p{57pt}|p{57pt}|p{57pt}|p{57pt}}
\toprule[1pt]
\rowcolor{gray!20}
Methods & \makecell[c]{ARIA}  & \makecell[c]{IOSTAR} & \makecell[c]{MSD}& \makecell[c]{Kvasir-SEG}  & \makecell[c]{BUS}  \\
\hline
FedAvg  & \makecell[c]{$63.37\pm0.85$} & \makecell[c]{$63.96\pm0.31$} & \makecell[c]{$82.20\pm1.36$} & \makecell[c]{$72.11\pm0.75$} & \makecell[c]{$74.43\pm0.74$} \\
\hline
FedAWA  & \makecell[c]{$64.79\pm0.60$} & \makecell[c]{$65.27\pm0.79$} & \makecell[c]{$82.13\pm1.01$} & \makecell[c]{$71.45\pm0.71$} & \makecell[c]{ $\textbf{74.54}\pm1.32$} \\
\hline
FedLWS  & \makecell[c]{ $\textbf{66.41}\pm0.58$} & \makecell[c]{ $\textbf{72.40}\pm2.21$} & \makecell[c]{ $\textbf{84.27}\pm1.36$} & \makecell[c]{$20.69\pm9.68$} & \makecell[c]{$74.20\pm0.37$} \\
\hline
FedNova & \makecell[c]{$65.13\pm0.61$} & \makecell[c]{$65.79\pm0.45$} & \makecell[c]{$82.55\pm0.72$} & \makecell[c]{$73.43\pm0.85$} & \makecell[c]{$73.97\pm2.41$} \\
\hline
FedProx & \makecell[c]{$63.63\pm0.62$} & \makecell[c]{$63.74\pm0.24$} & \makecell[c]{$82.25\pm1.42$} & \makecell[c]{$72.99\pm0.34$} & \makecell[c]{$73.86\pm1.38$} \\
\hline
MOON    & \makecell[c]{$62.94\pm0.71$} & \makecell[c]{$63.87\pm0.86$} & \makecell[c]{$82.18\pm0.89$} & \makecell[c]{$72.40\pm0.56$} & \makecell[c]{$73.52\pm0.35$} \\
\hline
PN      & \makecell[c]{$60.71\pm1.86$} & \makecell[c]{$59.51\pm1.93$} & \makecell[c]{$83.72\pm1.13$} & \makecell[c]{ $\textbf{74.17}\pm1.26$} & \makecell[c]{$74.15\pm3.50$} \\
\bottomrule[1pt]
\end{tabular}
\label{generalization}
\end{table*}

\noindent\textbf{Fed-COSAS.} From Table~\ref{COSAS_Table}, FedBN achieves strong results across all clients: 76.32\% on C1, 67.64\% on C2, and 79.21\% on C3, ranking second on C1 and C3 and first on C2. SioBN follows closely with 75.92\%, 67.34\%, and 78.41\% on C1$\sim$C3, respectively, reaffirming the benefit of client-specific batch normalization statistics. These pFL methods not only surpass the local training but also rival the best generic methods---FedBN's performance on C1 and C3 is comparable to FedRDN's, while it ties with PN on C2. Ditto and FedPer show moderate improvements but do not consistently outperform the generic top performers. FedRoD also exhibits catastrophic failure on this dataset, with extremely low means and high variances.

\noindent\textbf{Fed-BUS.} In Table~\ref{Ultrasound_Table}, FedPer emerges as the top overall performer, achieving the highest score on C1 (86.44\%) and C4 (73.36\%). This underscores the value of maintaining client-specific feature extractors when domain shifts are pronounced. FedBN delivers the best performance on C3 (91.04\%) and remains competitive on C2 (74.52\%), reaffirming the importance of personalized batch normalization statistics. SioBN also performs strongly, particularly on C3 (90.62\%) and C1 (85.00\%). In contrast, Ditto and FedRoD consistently underperform both generic top performers and other personalized methods, with FedRoD showing particular weakness on C1 (81.86\%).

\begin{figure*}[!t]
    \centering
    \begin{minipage}[t]{0.24\textwidth}
        \centering
        \includegraphics[width=\linewidth]{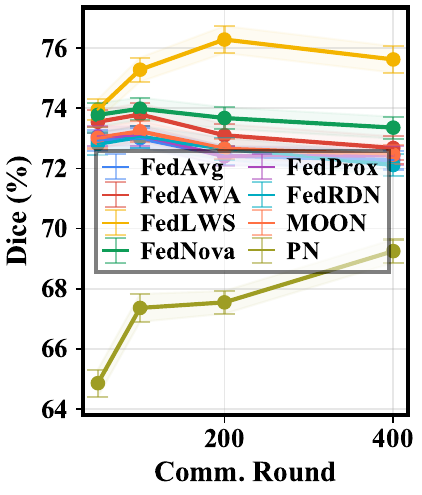}
        \vspace{-6mm}
        \captionof{subfigure}{Fed-Vessel}
    \end{minipage}
    \hfill
    \begin{minipage}[t]{0.24\textwidth}
        \centering
        \includegraphics[width=\linewidth]{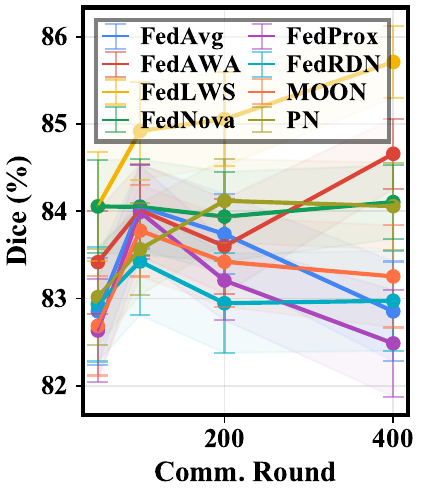}
        \vspace{-6mm}
        \captionof{subfigure}{Fed-Prostate}
    \end{minipage}
    \hfill
    \begin{minipage}[t]{0.24\textwidth}
        \centering
        \includegraphics[width=\linewidth]{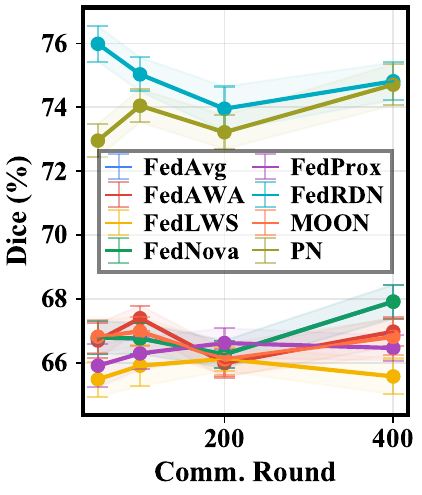}
        \vspace{-6mm}
        \captionof{subfigure}{Fed-COSAS}
    \end{minipage}
    \hfill
    \begin{minipage}[t]{0.24\textwidth}
        \centering
        \includegraphics[width=\linewidth]{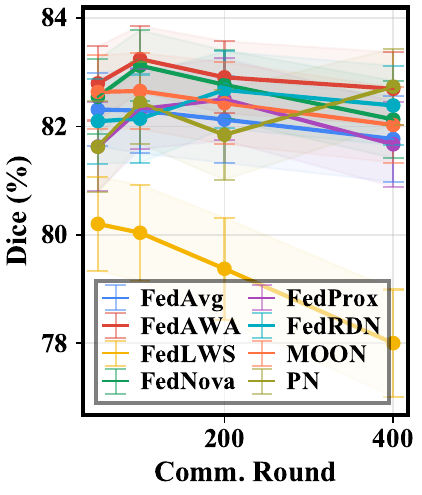}
        \vspace{-6mm}
        \captionof{subfigure}{Fed-BUS}
    \end{minipage}
    
    
    \begin{minipage}[t]{0.24\textwidth}
        \centering
        \includegraphics[width=\linewidth]{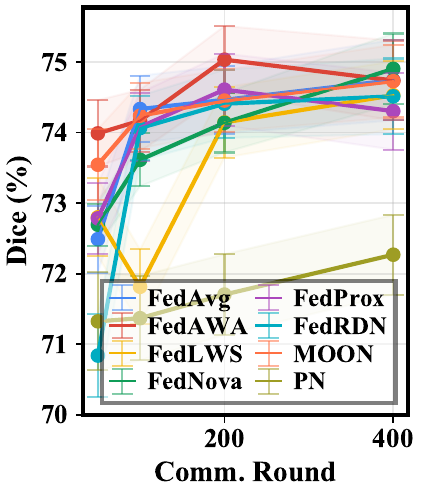}
        \vspace{-6mm}
        \captionof{subfigure}{Fed-MG}
    \end{minipage}
    \hfill
      \begin{minipage}[t]{0.24\textwidth}
        \centering
        \includegraphics[width=\linewidth]{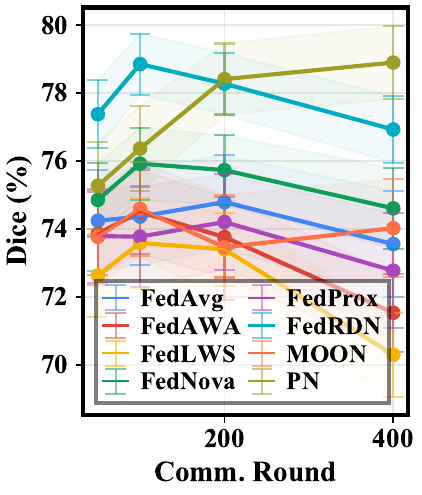}
        \vspace{-6mm}
        \captionof{subfigure}{Fed-Polyp}
    \end{minipage}  
    \hfill
    \begin{minipage}[t]{0.24\textwidth}
        \centering
        \includegraphics[width=\linewidth]{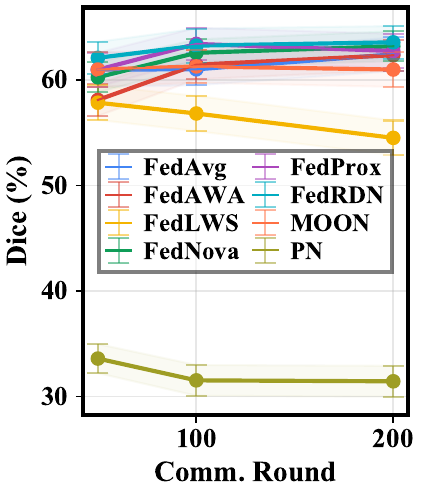}
        \vspace{-6mm}
        \captionof{subfigure}{Fed-Pancreas}
    \end{minipage}
    \hfill
    \begin{minipage}[t]{0.24\textwidth}
        \centering
        \includegraphics[width=\linewidth]{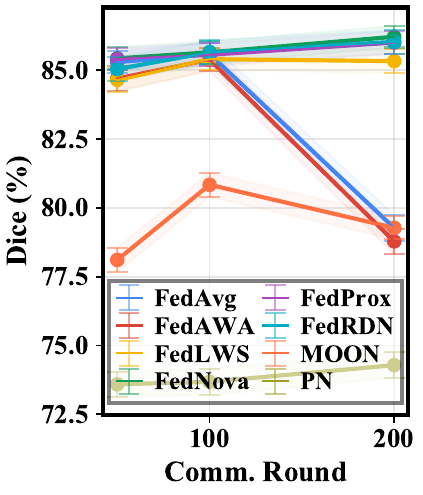}
        \vspace{-6mm}
        \captionof{subfigure}{FeTS2022}
    \end{minipage}   
    \caption{The segmentation performance of different gFL methods with different communication rounds on 2D and 3D datasets.} 
    \label{fig:comm_effect_gfl}
\end{figure*}

\noindent\textbf{Fed-MG.} Table~\ref{Gland_Table} reports the performance of pFL methods on Fed-MG. We observe that FedBN emerges as the overall best performer, achieving the highest scores on both C1 (79.46\%) and C2 (70.54\%). Critically, FedBN's result on C2 surpasses all generic methods. SioBN also performs strongly on C1 (79.18\%), though it shows unexpected weakness on C2 (64.12\%), indicating potential instability in its normalization strategy for certain domains. Ditto and FedPer deliver moderate improvements but do not match FedBN's performance. FedRoD exhibits catastrophic failure on C1 (65.48\%), further confirming its unsuitability for segmentation tasks.

\noindent\textbf{Fed-Polyp.} As shown in Table~\ref{Polyp_Table}, pFL methods exhibit noticeable performance variation across clients. FedBN achieves 87.72\%, 79.48\%, and 78.27\% on C2, C3, and C4, respectively, demonstrating strong competitiveness. Ditto achieves the highest score on C5 (68.07\%), which is the only method to approach the local training, underscoring its effectiveness in learning personalized models while maintaining global structure. FedPer and SioBN also perform competitively on several clients, with SioBN achieving 87.05\% on C2 and 88.64\% on C1. FedRoD exhibits instability again, particularly on C5 (33.98\%).

\noindent\textbf{Fed-Pancreas.} In Table~\ref{Pancreas_Table}, pFL methods still exhibit noticeable performance variation across clients on Fed-Pancreas. FedRoD achieves the highest result on C1 with 57.54\%, while FedBN reaches 34.36\% on C3 and 74.42\% on C5, demonstrating certain competitiveness. SioBN attains 74.24\% on C2, which is close to the best-performing generic method. In contrast, Ditto yields lower results on C1 and C3 with substantial fluctuations. FedPer maintains relatively stable performance across multiple clients but does not demonstrate a clear leading advantage.

\noindent\textbf{Fed-M\&Ms.} As shown in Table~\ref{MMS_Table}, FedBN achieves the highest overall score on C1 (82.84\%) and ties for best on C2 (85.57\%) and C3 (87.39\%), reaffirming the value of client-specific batch normalization statistics. SioBN matches FedBN on C2 (85.67\%) and secures the second-best result on C4 (81.29\%), demonstrating its effectiveness. FedPer delivers the top score on C5 (80.40\%), underscoring the benefit of personalized feature extractors. However, Ditto and FedRoD show weaker performance, particularly on C4 and C5, where they fall behind both generic leaders and other personalized methods. This suggests that their personalization mechanisms may not fully exploit the available cross-client knowledge.

\noindent\textbf{FeTS2022. } In Table~\ref{FeTS_Table}, FedBN stands out as the most consistent pFL method, achieving top performance on C5 (89.22\%), C6 (80.74\%), and tying for best on C2 (88.78\%) and C4 (82.98\%). This reaffirms the value of client-specific batch normalization statistics in capturing domain-specific intensity distributions in MRI. SioBN delivers the highest score on C7 (86.19\%) and ties on C9 (88.56\%), demonstrating its effectiveness. FedPer secures the best result on C8 (90.98\%), highlighting the benefit of personalized feature extractors. Ditto and FedRoD performed poorly across multiple clients. In particular, Ditto falls behind on C5 (78.95\%) and C9 (82.34\%), suggesting that its dual-model personalization may limit beneficial knowledge transfer in larger-scale federations.

\begin{figure*}[!t]
    \centering
    

    \begin{minipage}[t]{0.24\textwidth}
        \centering
        \includegraphics[width=\linewidth]{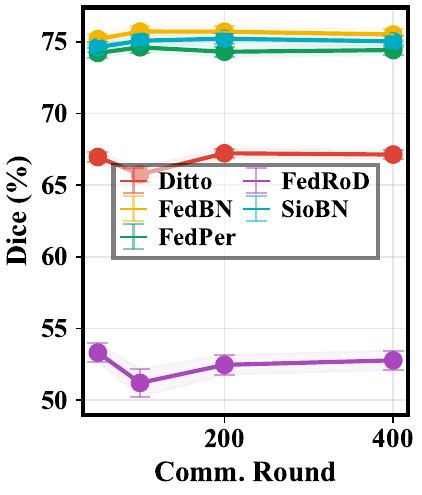}
        \vspace{-6mm}
        \captionof{subfigure}{Fed-Vessel}
    \end{minipage}
    \hfill
    \begin{minipage}[t]{0.24\textwidth}
        \centering
        \includegraphics[width=\linewidth]{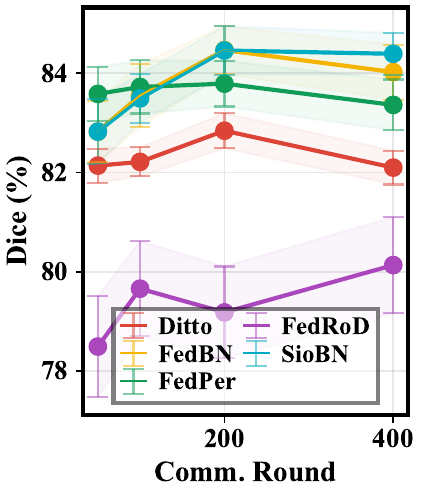}
         \vspace{-6mm}
        \captionof{subfigure}{Fed-Prostate}
    \end{minipage}
    \hfill
     \begin{minipage}[t]{0.24\textwidth}
        \centering
        \includegraphics[width=\linewidth]{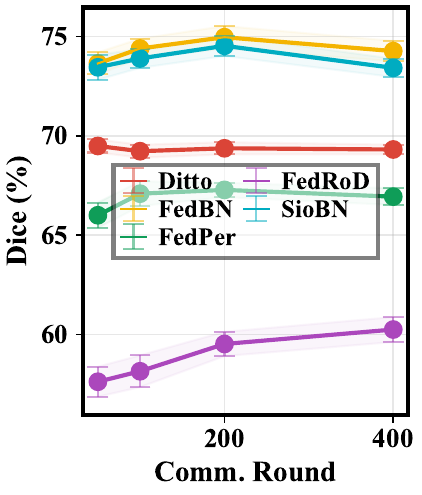}
         \vspace{-6mm}
        \captionof{subfigure}{Fed-COSAS}
    \end{minipage}   
    \hfill
    \begin{minipage}[t]{0.24\textwidth}
        \centering
        \includegraphics[width=\linewidth]{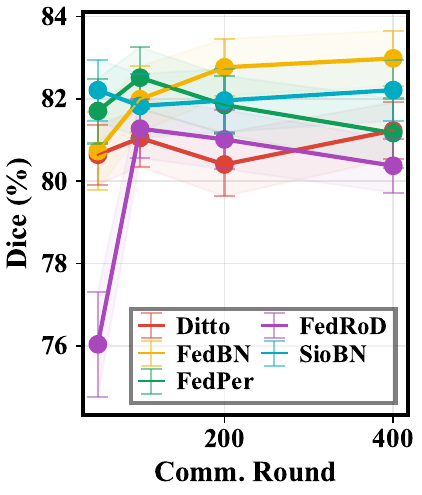}
        \vspace{-6mm}
        \captionof{subfigure}{Fed-BUS}
    \end{minipage}    
    
    
    \begin{minipage}[t]{0.24\textwidth}
        \centering
        \includegraphics[width=\linewidth]{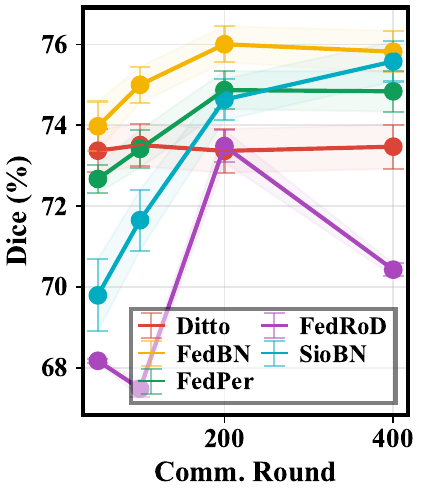}
         \vspace{-6mm}
        \captionof{subfigure}{Fed-MG}
    \end{minipage}
    \hfill
      \begin{minipage}[t]{0.24\textwidth}
        \centering
        \includegraphics[width=\linewidth]{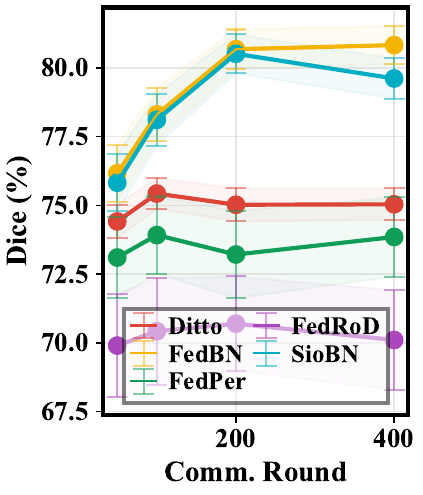}
         \vspace{-6mm}
        \captionof{subfigure}{Fed-Polyp}
    \end{minipage}  
    \hfill
    \begin{minipage}[t]{0.24\textwidth}
        \centering
        \includegraphics[width=\linewidth]{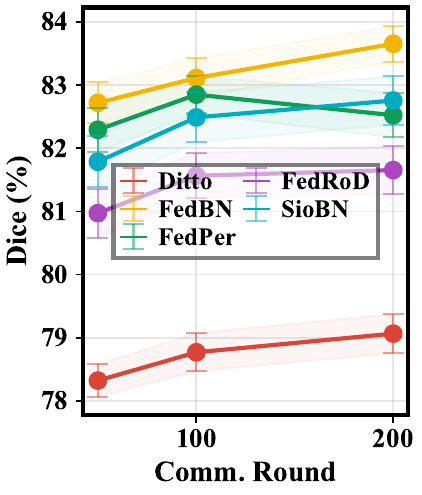}
         \vspace{-6mm}
        \captionof{subfigure}{Fed-M\&Ms}
    \end{minipage}   
    \hfill
    \begin{minipage}[t]{0.24\textwidth}
        \centering
        \includegraphics[width=\linewidth]{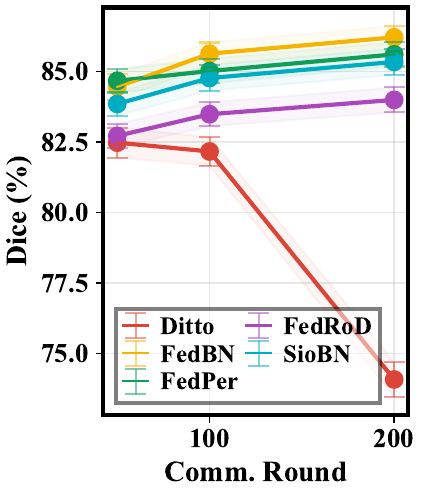}
         \vspace{-6mm}
        \captionof{subfigure}{FeTS2022}
    \end{minipage}
    \caption{The segmentation performance of different pFL method with different communication rounds on 2D and 3D datasets.}
    \label{fig:comm_effect_pfl}
\end{figure*}

\subsection{Generalization}
To assess the practical applicability of gFL methods in real-world clinical scenarios, we evaluate their generalization capability on five unseen datasets spanning diverse imaging modalities: ARIA~\cite{farnell2008enhancement} and IOSTAR~\cite{7530915} (vessel segmentation with fundus images), MSD-prostate~\cite{antonelli2022medical} (prostate segmentation with MRI images), Kvasir-SEG~\cite{jha2020kvasir} (polyp segmentation with endoscopic images), and BUS~\cite{yap2017automated} (breast tumor segmentation with ultrasound images). These datasets are entirely held out during training, serving as a rigorous test of each method's ability to segment images from previously unseen domains without any fine-tuning.

As shown in Table~\ref{generalization}, FedLWS demonstrates remarkable generalization on three of the five datasets, achieving the highest Dice scores on ARIA (66.41\%), IOSTAR (72.40\%), and MSD (84.27\%). Its adaptive weighting mechanism appears to learn representations that transfer well to unseen fundus and organ segmentation tasks. However, FedLWS catastrophically fails on Kvasir-SEG, plummeting to only 20.69\%, with a high variance. The marked performance collapse indicates that the method over-optimizes for the specific feature distributions during federation, which in turn narrows its generalizability and leads to failure on unseen endoscopic data. This result serves as an important caution against deploying adaptive weighting strategies without comprehensive out-of-distribution (OOD) validation. PN exhibits the strongest performance on Kvasir-SEG (74.17\%) and competitive results on BUS (74.15\%) and MSD (83.72\%). Its learnable global normalization statistics seem to capture domain-invariant features beneficial for endoscopic and ultrasound images, though it lags behind on retinal datasets (ARIA: 60.71\%, IOSTAR: 59.51\%). The method's stability across most OOD tasks is noteworthy, but its moderate performance on fundus data indicates limitations in handling domain shifts characterized by fine vessel structures. FedAWA and FedNova deliver consistently mid-to-high performance across all five datasets, with FedAWA slightly edging out others on BUS (74.54\%). Their balanced generalization suggests that simple aggregation with adaptive weighting or normalized averaging provides a reasonable trade-off between fitting training domains and retaining OOD robustness. 
FedAvg, FedProx, and MOON demonstrated similar performance profiles, with Dice scores clustering within a narrow range of 63$\sim$74\% across most datasets. This consistency suggests that conventional FL methods provide a moderate, yet reliable, level of generalization, though they do not exhibit a clear advantage when adapting to any specific unseen domain.

\begin{figure*}[!t] 
    \centering
    
    \begin{minipage}[t]{0.30\textwidth}
        \centering
        \includegraphics[width=\linewidth]{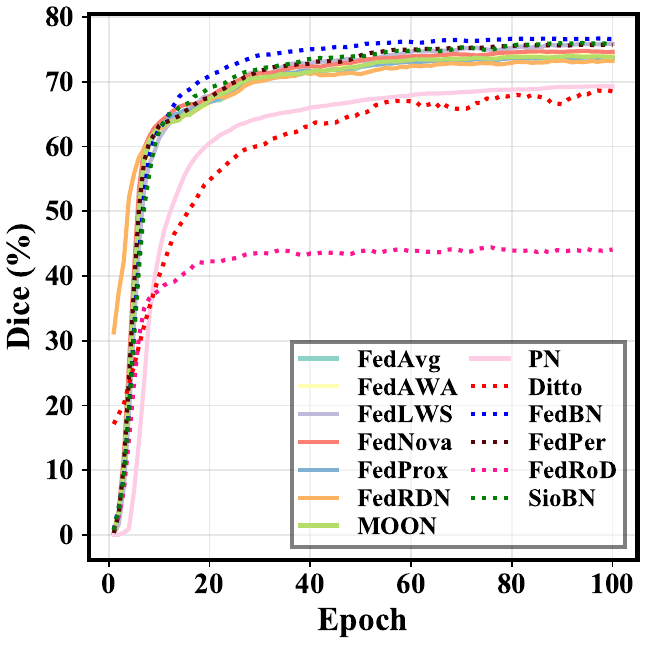}
        \vspace{-3mm}
        \captionof{subfigure}{Fed-Vessel}
    \end{minipage}    
    \hfill
    \begin{minipage}[t]{0.30\textwidth}
        \centering
        \includegraphics[width=\linewidth]{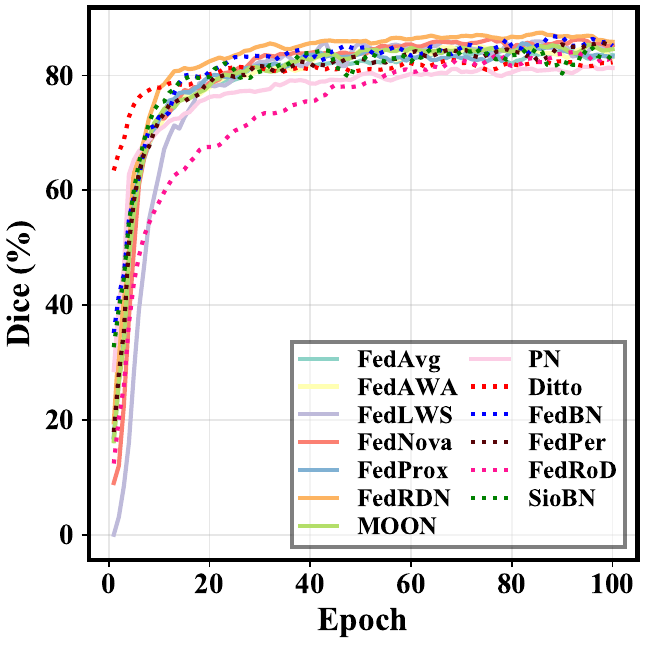}
       \vspace{-3mm}
        \captionof{subfigure}{Fed-Prostate}
    \end{minipage}
    \hfill
    \begin{minipage}[t]{0.30\textwidth}
        \centering
        \includegraphics[width=\linewidth]{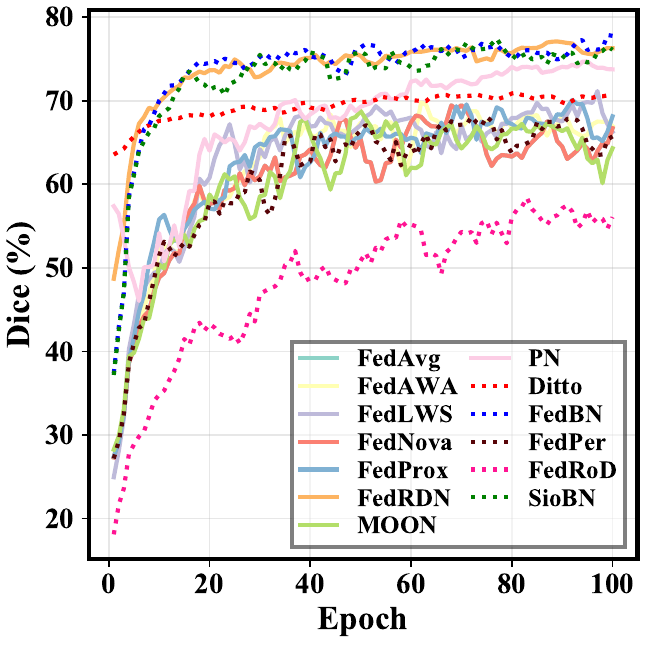}
        \vspace{-3mm}
        \captionof{subfigure}{Fed-COSAS}
    \end{minipage}
    
    
    \begin{minipage}[t]{0.30\textwidth}
        \centering
        \includegraphics[width=\linewidth]{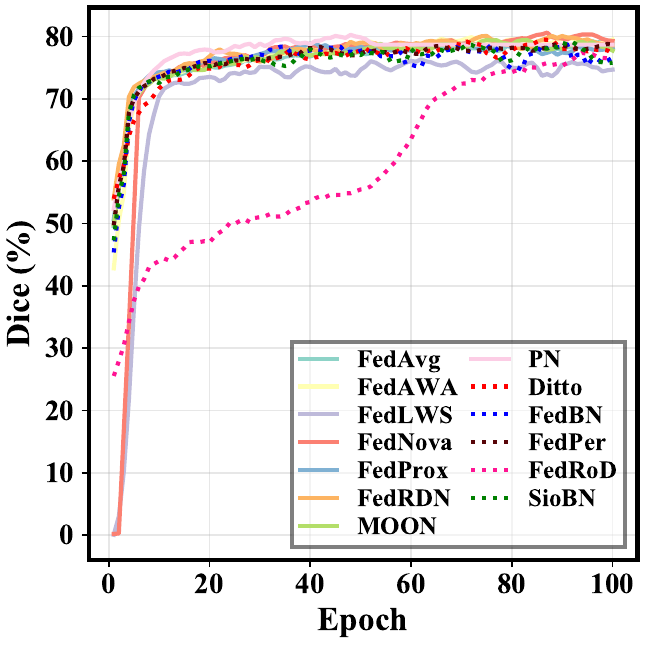}
       \vspace{-3mm}
        \captionof{subfigure}{Fed-BUS}
    \end{minipage}    
    \hfill
    \begin{minipage}[t]{0.30\textwidth}
        \centering
        \includegraphics[width=\linewidth]{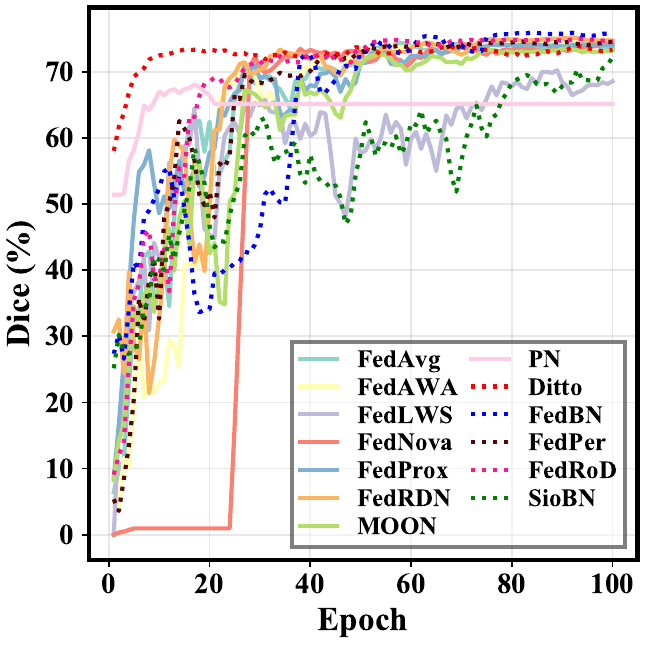}
        \vspace{-3mm}
        \captionof{subfigure}{Fed-MG}
    \end{minipage}
    \hfill
    \begin{minipage}[t]{0.30\textwidth}
        \centering
        \includegraphics[width=\linewidth]{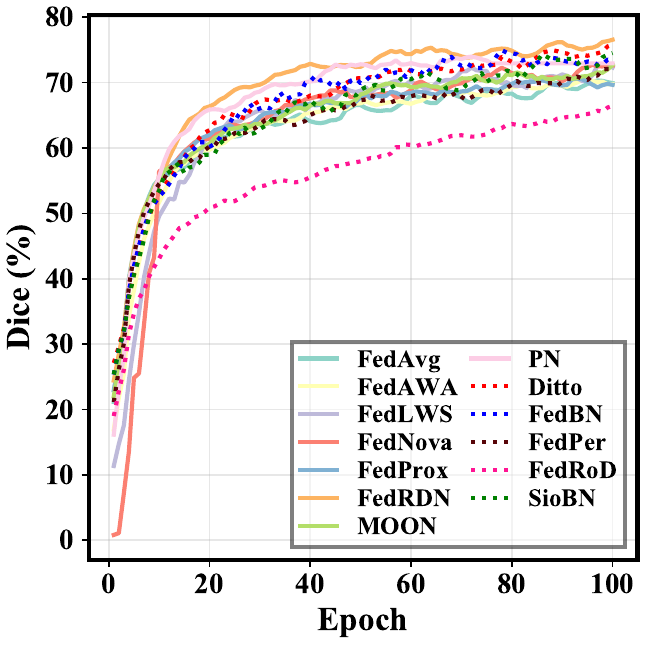}
       \vspace{-3mm}
        \captionof{subfigure}{Fed-Polyp}
    \end{minipage}

    \begin{minipage}[t]{0.30\textwidth}
        \centering
        \includegraphics[width=\linewidth]{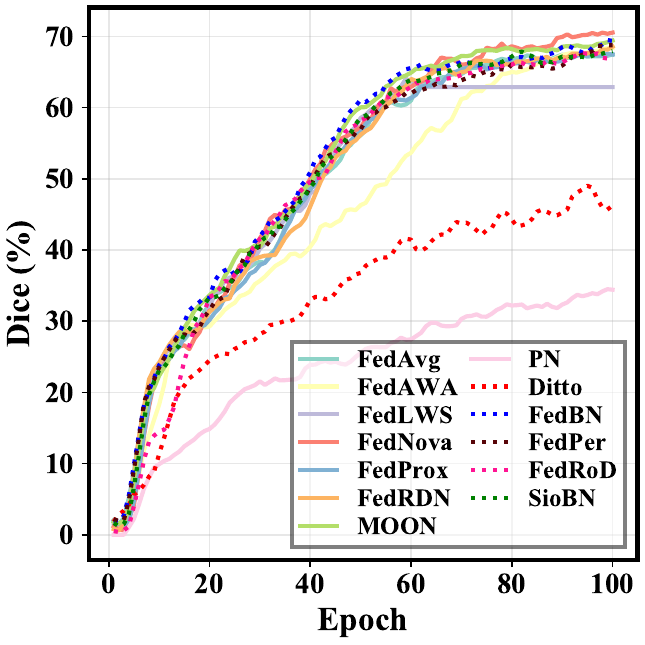}
        \vspace{-3mm}
        \captionof{subfigure}{Fed-Pancreas}
    \end{minipage}
    \hfill
    \begin{minipage}[t]{0.30\textwidth}
        \centering
        \includegraphics[width=\linewidth]{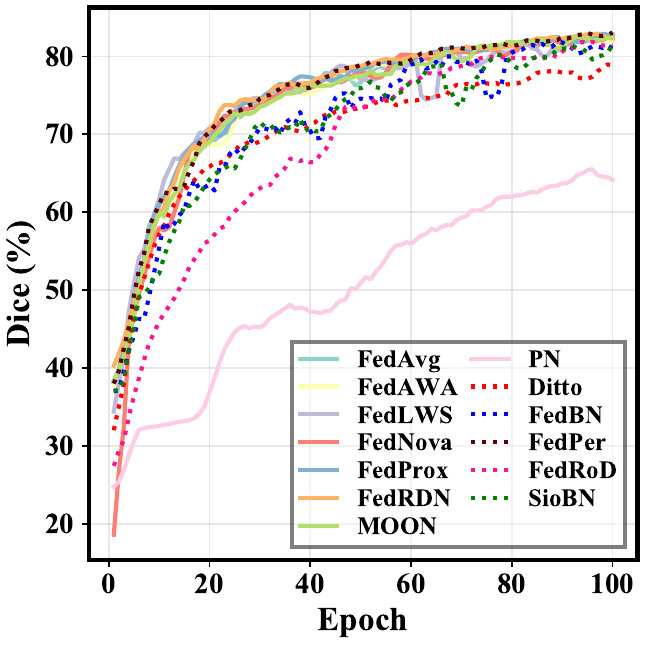}
        \vspace{-3mm}
        \captionof{subfigure}{Fed-M\&Ms}
    \end{minipage}
    \hfill
    \begin{minipage}[t]{0.30\textwidth}
        \centering
        \includegraphics[width=\linewidth]{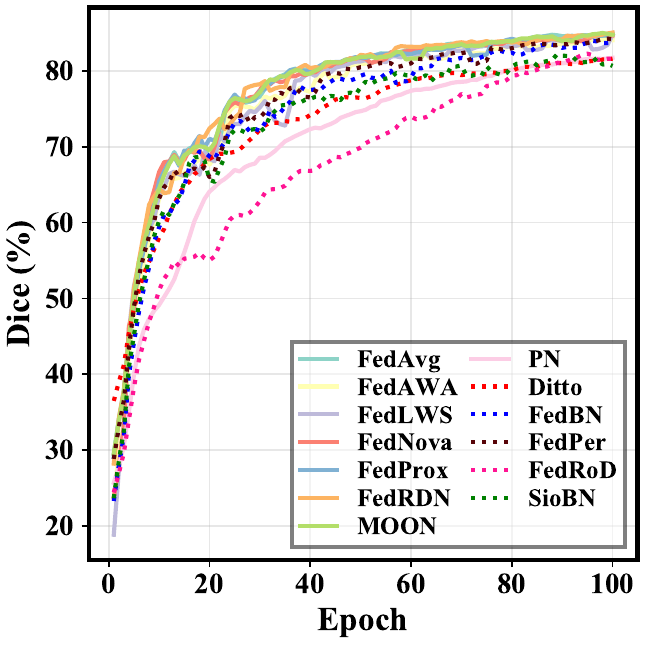}
        \vspace{-3mm}
        \captionof{subfigure}{FeTS2022}
    \end{minipage}
    \caption{The convergence behaviors of different FL methods 2D and 3D datasets.}
    \label{fig:converge}
\end{figure*}

\subsection{Effect of Communication Frequency}
To investigate the impact of communication frequency on federated learning performance, we evaluate various FL algorithms on nine medical image segmentation datasets under varying communication rounds and local update frequencies, as shown in Figs.~\ref{fig:comm_effect_gfl} and \ref{fig:comm_effect_pfl}. For the six datasets (Fed-Vessel, Fed-Prostate, Fed-COSAS, Fed-BUS, Fed-MG, and Fed-Polyp), the total number of training epochs is fixed at 400, meaning that as the number of communication rounds decreases from 400 to 50, the number of local epochs per round correspondingly increases from 1 to 8. For the three datasets (Fed-Pancreas, Fed-M\&MS and FeTS2022), the total epochs are 200, with local epochs per round increasing from 1 to 4 as communication rounds decrease from 200 to 50. This experimental design allows us to investigate the trade-off between communication frequency and local computation, and to assess the robustness of each FL method to different degrees of model staleness and client drift.

\subsubsection{Effect of Communication Frequency for gFL Methods}
As illustrated in Fig.~\ref{fig:comm_effect_gfl}, most methods achieve comparable performance on Fed-Vessel, with FedLWS and FedNova showing a slight edge across all communication settings. FedLWS attains the highest accuracy at 400 rounds and maintains competitive performance as rounds decrease, though it experiences a gradual decline at 50 rounds. In comparison, PN exhibits a marked sensitivity to communication constraints, with its performance collapsing sharply to 64.86\% at 50 rounds. FedAWA and FedProx exhibit stable behavior with minimal degradation when local epochs increase. Fed-Prostate offers a different perspective on algorithm behavior. Although FedLWS achieves the best performance across different communication frequencies, its performance demonstrates a clear downward trend as the communication frequency decreases. Similarly, FedAWA exhibits this same pattern of decline. In contrast, FedNova exhibited highly stable performance across different communication frequencies. The other methods (FedAvg, MOON, FedRDN and FedProx) exhibited a counterintuitive phenomenon: higher communication frequencies led to lower performance. This pattern was also observed on the Fed-BUS dataset, with all methods except PN following this trend. The results on Fed-COSAS reveal a stark contrast between methods. FedRDN and PN both achieve superior performance compared to all other methods; however, they respond differently to communication constraints. The performance of FedRDN improves with fewer communication rounds, while PN's performance declines under the same conditions. The remaining methods, including FedAvg, FedAWA, FedLWS, FedNova, FedProx, and MOON, all fall within the 65$\sim$68\% range, with little variation across different communication rounds.

On the Fed-MG dataset, nearly all methods exhibit a consistent trend: segmentation performance improves as communication frequency increases. Notably, FedRDN suffers a substantial performance degradation when the communication frequency is extremely low. The Fed-Polyp dataset presents a unique pattern. As communication frequency decreases, all methods except PN---which shows a declining trend---exhibits an initial performance improvement followed by a subsequent drop. FedRDN, FedAWA, and FedLWS, in particular, experiences notable performance degradation. This suggests that, in federated learning settings, more frequent communication does not necessarily translate to better performance.On the Fed-Pancreas and FeTS2022 datasets, FedRDN, FedProx, and FedNova all demonstrate low sensitivity to variations in communication frequency. In contrast, PN and FedLWS exhibit opposing performance trends across the two datasets: their performance improves slightly as the communication frequency decreases on Fed-Pancreas, but shows a mild decline on FeTS2022. Meanwhile, FedAWA and FedAvg remain highly stable on Fed-Pancreas, yet achieve substantially lower performance on FeTS2022 when the communication frequency is high.

\subsubsection{Effect of Communication Frequency for pFL Methods}
As shown in Fig.~\ref{fig:comm_effect_pfl}, pFL methods demonstrated minimal performance fluctuation on Fed-Vessel across varying communication frequencies. Notably, FedBN and SioBN consistently achieve the highest Dice scores regardless of the communication setting. This indicates that methods incorporating batch normalization (FedBN and SioBN) exhibit robustness to the trade-off between communication frequency and local update steps. On Fed-Prostate, as communication frequency decreases, FedRoD exhibits an overall downward trend, while the other pFL methods first increase and then decline. On Fed-COSAS and Fed-BUS, FedBN, FedPer, and Ditto exhibit minimal performance fluctuation as communication frequency vary, while FedRoD shows an overall downward trend. On Fed-MG, Ditto stands out as the only method that maintains stable performance across communication frequencies, whereas most other methods show a pronounced decline as frequency decreases. On the Fed-MG, Fed-M\&Ms, and FeTS2022 datasets, with the exception of Ditto, all other methods exhibit a marked downward trend as communication frequency decreases. In contrast, Ditto exhibits a completely different performance trend across these three datasets. In Fed-Polyp, FedBN and SioBN again lead with scores above 80\% at high communication frequency, but both exhibit a clear downward trend as rounds decrease: 
FedBN drops from 80.83\% at 400 rounds to 76.16\% at 50 rounds, and SioBN from 80.51\% at 200 rounds to 75.82\% at 50 rounds. The remaining methods, however, demonstrate remarkable stability across communication rounds. Overall, among these pFL methods, FedBN consistently achieves the best performance across different communication frequencies, while also exhibiting a clear downward trend as communication frequency decreases.

\begin{table*}[!t]
\center
\renewcommand\arraystretch{1}
\setlength{\tabcolsep}{1pt}
\caption{Fairness study of different FL methods on Fed-Vessel, Fed-Prostate, Fed-Polyp and Fed-COSAS (\%).}
\label{General}
\begin{tabular}{p{40pt}|p{34pt}p{34pt}p{34pt}|p{34pt}p{34pt}p{34pt}|p{34pt}p{34pt}p{34pt}|p{34pt}p{34pt}p{34pt}}
\toprule[1pt]
\rowcolor{gray!20}
\multirow{2}{*}{Methods} & \multicolumn{3}{c|}{Fed-Vessel} & \multicolumn{3}{c|}{Fed-Prostate} & \multicolumn{3}{c|}{Fed-Polyp} & \multicolumn{3}{c}{Fed-COSAS} \\\cline{2-13} 
   &  \makecell[c]{aDice$\uparrow$}  & \makecell[c]{Std$\downarrow$}  & \makecell[c]{wDice$\uparrow$}   &\makecell[c]{aDice$\uparrow$}  & \makecell[c]{Std$\downarrow$}  & \makecell[c]{wDice$\uparrow$}& \makecell[c]{aDice$\uparrow$}  & \makecell[c]{Std$\downarrow$}  & \makecell[c]{wDice$\uparrow$}& \makecell[c]{aDice$\uparrow$}  & \makecell[c]{Std$\downarrow$}  & \makecell[c]{wDice$\uparrow$} \\
\hline 
\multicolumn{13}{l}{\textit{Generic FL methods}}  \\
\hline
 FedAvg  & \makecell[c]{73.01}  & \makecell[c]{3.28}  & \makecell[c]{68.50}  &\makecell[c]{84.05}  & \makecell[c]{4.80}  & \makecell[c]{76.35} &\makecell[c]{74.36}  & \makecell[c]{14.23}  & \makecell[c]{49.68}&\makecell[c]{66.76}  & \makecell[c]{4.84}  & \makecell[c]{60.14}\\
\hline                  
 FedAWA  & \makecell[c]{73.79}  & \makecell[c]{3.73}  & \makecell[c]{68.45}  &\makecell[c]{84.01}  & \makecell[c]{5.26}  & \makecell[c]{74.84} &\makecell[c]{74.49}  & \makecell[c]{12.32}  & \makecell[c]{53.55}&\makecell[c]{67.39}  & \makecell[c]{3.96}  & \makecell[c]{61.82}\\
\hline                  
 FedLWS  & \makecell[c]{\textbf{75.28}}  & \makecell[c]{4.01}  & \makecell[c]{\textbf{70.24}}  &\makecell[c]{\textbf{84.92}}  & \makecell[c]{5.56}  & \makecell[c]{74.49} &\makecell[c]{73.58}  & \makecell[c]{13.32}  & \makecell[c]{49.38}&\makecell[c]{65.90}  & \makecell[c]{6.38}  & \makecell[c]{57.11}\\
\hline                  
 FedNova  & \makecell[c]{73.98}  & \makecell[c]{3.68}  & \makecell[c]{68.86}  &\makecell[c]{84.05}  & \makecell[c]{5.46}  & \makecell[c]{74.94} &\makecell[c]{75.92}  & \makecell[c]{10.60}  & \makecell[c]{58.79}&\makecell[c]{66.76}  & \makecell[c]{4.84}  & \makecell[c]{60.14}\\
\hline                  
 FedProx  & \makecell[c]{73.07}  & \makecell[c]{\textbf{3.24}}  & \makecell[c]{68.56}  &\makecell[c]{83.99}  & \makecell[c]{5.37}  & \makecell[c]{76.00} &\makecell[c]{73.77}  & \makecell[c]{14.73}  & \makecell[c]{47.57}&\makecell[c]{66.29}  & \makecell[c]{5.01}  & \makecell[c]{59.25}\\
\hline                  
 FedRDN  & \makecell[c]{73.05}  & \makecell[c]{3.68}  & \makecell[c]{67.81}  &\makecell[c]{83.42}  & \makecell[c]{6.15}  & \makecell[c]{71.92} &\makecell[c]{\textbf{78.85}}  & \makecell[c]{8.95}  & \makecell[c]{65.85}&\makecell[c]{\textbf{75.03}}  & \makecell[c]{5.32}  & \makecell[c]{67.60}\\
\hline                  
 MOON   & \makecell[c]{73.23}  & \makecell[c]{3.40}  & \makecell[c]{68.79}  &\makecell[c]{83.77}  & \makecell[c]{5.25}  & \makecell[c]{75.26} &\makecell[c]{74.58}  & \makecell[c]{13.89}  & \makecell[c]{50.71}&\makecell[c]{66.98}  & \makecell[c]{4.52}  & \makecell[c]{61.29}\\
\hline                  
 PN  & \makecell[c]{67.36}  & \makecell[c]{4.63}  & \makecell[c]{57.94}  &\makecell[c]{83.56}  & \makecell[c]{5.27}  & \makecell[c]{75.24} &\makecell[c]{76.36}  & \makecell[c]{12.52}  & \makecell[c]{52.99}&\makecell[c]{74.05}  & \makecell[c]{5.14}  & \makecell[c]{\textbf{67.79}}\\
\hline 
\multicolumn{13}{l}{\textit{Personalized FL methods}}  \\
\hline 
 Ditto  & \makecell[c]{65.75}  & \makecell[c]{5.33}  & \makecell[c]{55.79}  &\makecell[c]{82.21}  & \makecell[c]{\textbf{2.93}}  & \makecell[c]{\textbf{77.83}} &\makecell[c]{75.42}  & \makecell[c]{\textbf{5.62}}  & \makecell[c]{\textbf{68.07}}&\makecell[c]{69.22}  & \makecell[c]{\textbf{3.26}}  & \makecell[c]{64.64}\\
 \hline 
 FedBN  & \makecell[c]{75.71}  & \makecell[c]{3.84}  & \makecell[c]{69.62}  &\makecell[c]{83.55}  & \makecell[c]{6.31}  & \makecell[c]{73.86} &\makecell[c]{78.30}  & \makecell[c]{9.75}  & \makecell[c]{60.19}&\makecell[c]{74.38}  & \makecell[c]{4.91}  & \makecell[c]{67.64}\\
 \hline 
 FedPer  & \makecell[c]{74.61}  & \makecell[c]{3.64}  & \makecell[c]{68.65}  &\makecell[c]{83.72}  & \makecell[c]{5.35}  & \makecell[c]{75.96} &\makecell[c]{73.91}  & \makecell[c]{14.03}  & \makecell[c]{49.52}&\makecell[c]{67.08}  & \makecell[c]{6.25}  & \makecell[c]{58.24}\\
 \hline 
 FedRoD  & \makecell[c]{51.19}  & \makecell[c]{9.68}  & \makecell[c]{36.99}  &\makecell[c]{79.65}  & \makecell[c]{9.56}  & \makecell[c]{62.38} &\makecell[c]{70.42}  & \makecell[c]{19.36}  & \makecell[c]{33.98}&\makecell[c]{58.15}  & \makecell[c]{7.98}  & \makecell[c]{50.61}\\  
 \hline 
 SioBN  & \makecell[c]{75.08}  & \makecell[c]{3.71}  & \makecell[c]{69.08}  &\makecell[c]{83.49}  & \makecell[c]{4.89}  & \makecell[c]{76.03} &\makecell[c]{78.11}  & \makecell[c]{9.36}  & \makecell[c]{62.76}&\makecell[c]{73.89}  & \makecell[c]{4.74}  & \makecell[c]{67.34}\\
\bottomrule[1pt]               
\end{tabular}
\label{Fairness_1}
\end{table*}

\subsection{Analysis of Convergence Behaviors}
To investigate the convergence behaviors of various FL methods on different medical datasets, we plot their Dice scores of validation sets against the number of training epochs, as shown in Fig.~\ref{fig:converge}. The number of communication rounds $T$ is set to 100 for all datasets.

On the Fed-Vessel dataset, all methods exhibit a rapid performance increase in the first 30 epochs, followed by a gradual plateau. pFL approaches (FedBN, FedPer, SioBN) achieve the highest final Dice scores, marginally outperforming gFL methods (FedAvg, FedProx, FedNova, FedAWA, FedLWS, MOON) which cluster around 73$\sim$75\%. FedRDN shows an anomalously high initial score (around 31\%) but converges similarly to gFL methods. In contrast, PN and Ditto converge to notably lower scores, while FedRoD exhibits severe degradation, plateauing at only around 44\% Dice, indicating poor suitability for this task. For the Fed‑Prostate dataset, FedRDN achieves the highest final Dice, demonstrating superior convergence. FedBN, FedPer, and SioBN also reach competitive levels but exhibit moderate fluctuations. Traditional methods (FedAvg, FedProx, FedNova, MOON, FedAWA) cluster together and converge smoothly. FedLWS starts from near zero but catches up after 40 epochs, eventually matching these traditional methods. Ditto begins with an unusually high initial score yet shows limited growth, plateauing at about 82\%. FedRoD underperforms significantly, converging to only 83\% despite steady progress.

On the Fed‑COSAS dataset as illustrated in Fig.~\ref{fig:converge}(e), FedRDN, FedBN, and SioBN achieve the highest Dice scores with rapid initial improvement and stable convergence after around 60 epochs. Their curves are smooth and exhibit minimal fluctuations, indicating robust optimization. FedAvg, FedNova, FedProx, FedAWA, and FedPer follow a similar trend: steady but slower improvement. They display noticeable oscillations throughout training, particularly in the later stages, suggesting sensitivity to data heterogeneity. FedRoD lags significantly behind all others, peaking at only around 56\%, with a slow and unstable increase. On the Fed-BUS dataset, FedLWS starts from a near-zero Dice but catches up swiftly, reaching over 70\% by epoch 30 and maintaining competitive performance thereafter, indicating its strong adaptability despite poor initialization. FedNova also shows a sharp rise after epoch 10, achieving high scores early and continuing to improve steadily. FedAvg, FedAWA, FedProx, FedRDN, MOON, FedBN, FedPer, and SioBN follow similar smooth trajectories, starting at 45$\sim$55\% and converging to 76$\sim$79\% with minor fluctuations. These methods demonstrate robust and balanced convergence, with FedAWA and MOON slightly outperforming FedAvg in later stages. FedProx and FedRDN exhibit moderate oscillations but remain within the same performance band. FedRoD progresses slowly but consistently, ending near 76.40\%, indicating slower yet steady convergence.

On the Fed‑MG dataset, gradual and stable convergence is observed for FedAvg, FedProx, MOON, FedPer, and FedRoD. Among them, FedPer and FedRoD ascend slightly faster and reach marginally higher plateaus, indicating effective personalization or regularisation. FedAWA, FedRDN, and FedBN also progress steadily but ultimately achieve the top Dice scores. FedBN, in particular, exhibits the highest final value with very low variance after convergence, suggesting robust feature representation via batch normalization. On the Fed-Polyp dataset, FedRDN demonstrates the fastest convergence and attains the highest final Dice score, maintaining stable performance after around 60 epochs. Ditto and SioBN also achieve strong final results but with mild fluctuations in later stages. In contrast, FedAvg and FedProx converge more slowly and plateau at lower accuracies, indicating limited effectiveness in this task. Notably, FedRoD lags behind all other methods, with the slowest convergence and the lowest final Dice score, suggesting it is less suitable for this dataset. On the Fed-Pancreas dataset, most methods demonstrate smooth and steady convergence, with minimal oscillation after 60 epochs. PN and Ditto converge slowly and remain at a relatively low level. On the Fed-M\&MS and FeTS2022 datasets, except for PN, gFL methods demonstrate rapid and steady improvement, converging to high-performance plateaus with minimal fluctuations. PN  exhibits the slowest growth, particularly on the Fed-M\&MS dataset. Among pFL approaches, FedPer achieves the highest final segmentation performance, while maintaining smooth convergence throughout training on the two datasets. The others exhibit pronounced oscillations during training. 

\begin{table*}[!t]
\center
\renewcommand\arraystretch{1}
\setlength{\tabcolsep}{1pt}
\caption{Fairness study of different FL methods on Fed-BUS, Fed-Pancreas, Fed-M\&Ms and FeTS2022 (\%).}
\begin{tabular}{p{40pt}|p{34pt}p{34pt}p{34pt}|p{34pt}p{34pt}p{34pt}|p{34pt}p{34pt}p{34pt}|p{34pt}p{34pt}p{34pt}}
\toprule[1pt]
\rowcolor{gray!20}
\multirow{2}{*}{Methods} & \multicolumn{3}{c|}{Fed-BUS} & \multicolumn{3}{c|}{Fed-Pancreas} & \multicolumn{3}{c|}{Fed-M\&Ms} & \multicolumn{3}{c}{FeTS2022} \\\cline{2-13} 
   &  \makecell[c]{aDice$\uparrow$}  & \makecell[c]{Std$\downarrow$}  & \makecell[c]{wDice$\uparrow$}   &\makecell[c]{aDice$\uparrow$}  & \makecell[c]{Std$\downarrow$}  & \makecell[c]{wDice$\uparrow$}& \makecell[c]{aDice$\uparrow$}  & \makecell[c]{Std$\downarrow$}  & \makecell[c]{wDice$\uparrow$}& \makecell[c]{aDice$\uparrow$}  & \makecell[c]{Std$\downarrow$}  & \makecell[c]{wDice$\uparrow$} \\
\hline 
\multicolumn{13}{l}{\textit{Generic FL methods}}  \\
\hline
 FedAvg  & \makecell[c]{80.62}  & \makecell[c]{7.98}  & \makecell[c]{71.90}  &\makecell[c]{60.98}  & \makecell[c]{15.10}  & \makecell[c]{33.97} &\makecell[c]{82.42}  & \makecell[c]{3.54}  & \makecell[c]{79.00}&\makecell[c]{85.59}  & \makecell[c]{4.07}  & \makecell[c]{78.40}\\
\hline                  
 FedAWA  & \makecell[c]{\textbf{81.81}}  & \makecell[c]{\textbf{6.02}}  & \makecell[c]{\textbf{74.89}}  &\makecell[c]{61.42}  & \makecell[c]{13.81}  & \makecell[c]{36.68} &\makecell[c]{82.25}  & \makecell[c]{3.60}  & \makecell[c]{78.75}&\makecell[c]{85.38}  & \makecell[c]{4.15}  & \makecell[c]{77.60}\\
\hline                  
 FedLWS  & \makecell[c]{77.80}  & \makecell[c]{8.55}  & \makecell[c]{68.59}  &\makecell[c]{56.80}  & \makecell[c]{16.65}  & \makecell[c]{26.44} &\makecell[c]{82.32}  & \makecell[c]{3.41}  & \makecell[c]{79.52}&\makecell[c]{85.40}  & \makecell[c]{4.07}  & \makecell[c]{77.68}\\
\hline                  
 FedNova  & \makecell[c]{81.66}  & \makecell[c]{6.53}  & \makecell[c]{73.44}  &\makecell[c]{62.55}  & \makecell[c]{\textbf{12.73}}  & \makecell[c]{\textbf{41.83}} &\makecell[c]{82.54}  & \makecell[c]{3.37}  & \makecell[c]{79.01}&\makecell[c]{\textbf{85.65}}  & \makecell[c]{4.10}  & \makecell[c]{78.11}\\
\hline                  
 FedProx  & \makecell[c]{80.65}  & \makecell[c]{7.39}  & \makecell[c]{72.52}  &\makecell[c]{\textbf{63.39}}  & \makecell[c]{15.28}  & \makecell[c]{34.43} &\makecell[c]{82.50}  & \makecell[c]{3.49}  & \makecell[c]{79.30}&\makecell[c]{85.56}  & \makecell[c]{4.18}  & \makecell[c]{\textbf{78.63}}\\
\hline                  
 FedRDN  & \makecell[c]{80.43}  & \makecell[c]{8.28}  & \makecell[c]{71.34}  &\makecell[c]{63.23}  & \makecell[c]{16.13}  & \makecell[c]{32.99} &\makecell[c]{82.54}  & \makecell[c]{3.57}  & \makecell[c]{76.92}&\makecell[c]{\textbf{85.65}}  & \makecell[c]{4.27}  & \makecell[c]{77.71}\\
\hline                  
 MOON   & \makecell[c]{81.07}  & \makecell[c]{7.02}  & \makecell[c]{73.88}  &\makecell[c]{61.26}  & \makecell[c]{15.38}  & \makecell[c]{33.37} &\makecell[c]{82.24}  & \makecell[c]{3.53}  & \makecell[c]{78.40}&\makecell[c]{80.83}  & \makecell[c]{4.35}  & \makecell[c]{73.07}\\
\hline                  
 PN  & \makecell[c]{80.80}  & \makecell[c]{7.64}  & \makecell[c]{73.04}  &\makecell[c]{31.52}  & \makecell[c]{14.56}  & \makecell[c]{9.52} &\makecell[c]{60.06}  & \makecell[c]{5.13}  & \makecell[c]{54.37}&\makecell[c]{73.67}  & \makecell[c]{4.81}  & \makecell[c]{66.00}\\
\hline 
\multicolumn{13}{l}{\textit{Personalized FL methods}}  \\
\hline 
 Ditto  & \makecell[c]{79.23}  & \makecell[c]{6.63}  & \makecell[c]{71.64}  &\makecell[c]{45.92}  & \makecell[c]{24.77}  & \makecell[c]{14.60} &\makecell[c]{78.77}  & \makecell[c]{3.01}  & \makecell[c]{74.75}&\makecell[c]{82.16}  & \makecell[c]{5.09}  & \makecell[c]{75.95}\\
 \hline 
 FedBN  & \makecell[c]{80.40}  & \makecell[c]{8.19}  & \makecell[c]{70.62}  &\makecell[c]{60.98}  & \makecell[c]{15.89}  & \makecell[c]{34.36} &\makecell[c]{\textbf{83.11}}  & \makecell[c]{3.09}  & \makecell[c]{78.81}&\makecell[c]{85.63}  & \makecell[c]{\textbf{3.98}}  & \makecell[c]{77.71}\\
 \hline 
 FedPer  & \makecell[c]{81.06}  & \makecell[c]{7.64}  & \makecell[c]{73.36}  &\makecell[c]{62.31}  & \makecell[c]{15.75}  & \makecell[c]{33.60} &\makecell[c]{82.85}  & \makecell[c]{\textbf{2.98}}  & \makecell[c]{\textbf{80.14}}&\makecell[c]{85.01}  & \makecell[c]{4.38}  & \makecell[c]{77.18}\\
 \hline 
 FedRoD  & \makecell[c]{79.51}  & \makecell[c]{6.96}  & \makecell[c]{73.03}  &\makecell[c]{61.94}  & \makecell[c]{16.44}  & \makecell[c]{31.48} &\makecell[c]{81.56}  & \makecell[c]{3.57}  & \makecell[c]{78.00}&\makecell[c]{83.48}  & \makecell[c]{4.20}  & \makecell[c]{76.36}\\  
 \hline 
 SioBN  & \makecell[c]{80.20}  & \makecell[c]{7.88}  & \makecell[c]{71.71}  &\makecell[c]{61.83}  & \makecell[c]{15.74}  & \makecell[c]{33.89} &\makecell[c]{82.49}  & \makecell[c]{3.96}  & \makecell[c]{75.77}&\makecell[c]{84.77}  & \makecell[c]{4.53}  & \makecell[c]{75.79}\\
\bottomrule[1pt]               
\end{tabular}
\label{Fairness_2}
\end{table*}

\subsection{Fairness Study}
In federated learning, fairness is a critical consideration, measuring the variability in model performance across different clients. To comprehensively evaluate the fairness of each method, we use three metrics: the equally weighted average of Dice scores (aDice) across all clients, the standard deviation (Std) of client performance (where lower values indicate greater fairness), and the worst-client Dice score (wDice) representing the performance of the worst-performing client, which serves as a direct indicator of fairness. This section provides a detailed analysis of the fairness characteristics of eight generic FL approaches (FedAvg, FedAWA, FedLWS, FedNova, FedProx, FedRDN, MOON, and PN) and five personalized FL approaches (Ditto, FedBN, FedPer, FedRoD, and SioBN) across eight medical image segmentation datasets. The fairness performance of these methods is demonstrated in Tables \ref{Fairness_1} and \ref{Fairness_2}.

\subsubsection{Fairness Analysis of gFL Methods}
Among the gFL methods, FedRDN demonstrates the most balanced fairness. On the Fed-Polyp dataset, FedRDN achieves the highest aDice (78.85\%), the lowest Std (8.95), and the highest wDice (65.85\%), significantly outperforming other methods. Similarly, on Fed-COSAS, its aDice (75.03\%), Std (5.32), and wDice (67.60\%) are among the best. This suggests that FedRDN effectively mitigates client drift and enhances fairness. FedNova also performs well across multiple datasets; for instance, on Fed-Polyp, it attains an aDice of 75.92\%, a Std of 10.60, and a wDice of 58.79\%, with comprehensive performance second only to FedRDN. In contrast, FedLWS and FedProx show greater volatility in fairness: FedLWS achieves a high aDice (75.28\%) and a relatively good wDice (70.24\%) on Fed-Vessel, but on Fed-Pancreas, its aDice drops to 56.80\%, wDice plummets to 26.44\%, and Std soars to 16.65, indicating sensitivity to heterogeneity. MOON and PN exhibit severe degradation on some datasets: PN's aDice is merely 31.52\% on Fed-Pancreas, with a wDice as low as 9.52\%.

\subsubsection{Fairness Analysis of pFL Methods}
Personalized FL methods aim to tailor models for each client, theoretically reducing Std and improving wDice. Ditto stands out as a prime example: on Fed-Prostate, Ditto achieves the lowest Std (2.93) among all methods and the highest wDice (77.83\%), although its aDice (82.21\%) is slightly lower than some generic methods. On Fed-Polyp, Ditto maintains a remarkably low Std of 5.62 and a high wDice of 68.07\%, while keeping aDice (75.42\%) at a respectable level. This indicates that Ditto's local personalization effectively protects underperforming clients, significantly enhancing fairness. FedBN leverages batch normalization statistics to adapt to client distributions, achieving the highest aDice (75.71\%) and a competitive wDice (69.62\%) on Fed-Vessel, accompanied by a low Std (3.84). On Fed-M\&Ms, FedBN's aDice (83.11\%) and Std (3.09) are among the best. FedPer and SioBN show similar performance but offer limited fairness improvements on highly heterogeneous datasets like Fed-Pancreas. Notably, FedRoD performs the worst across all datasets, with its aDice, Std, and wDice lagging significantly behind other methods, suggesting its dual-classifier design may be unsuitable for medical image segmentation tasks.

\section{Discussion}
\label{sec:discussion}
\subsection{Discussion on Segmentation Performance}
Tables~\ref{Fundus_Table}-\ref{FeTS_Table} present a comprehensive evaluation of 13 FL methods across 9 medical image segmentation tasks, encompassing diverse imaging modalities (such as fundus, ultrasound, colonoscopy, infrared, and MRI) and heterogeneous data distributions. The results reveal several important insights regarding the behavior of generic and personalized FL approaches in medical imaging contexts.

The local training paradigm, where models are trained independently on each client without any federation, establishes a critical reference point. Notably, local training achieves competitive performance in several datasets, particularly on Fed-Vessel (\textit{e.g.}, 78.41\% on C5) and Fed-Prostate (\textit{e.g.}, 86.74\% on C5). This observation underscores the substantial statistical heterogeneity inherent in medical imaging data, where client-specific distributions can diverge significantly, making naive federation potentially detrimental. \textit{The strong local performance motivates the need for personalized approaches that can leverage cross-client knowledge while preserving client-specific features}.
Among generic FL methods, which aim to learn a single global model, we observe considerable variability. FedLWS consistently demonstrates robust performance, achieving the highest results among generic methods on Fed-Vessel and Fed-Prostate, suggesting its adaptive weighting mechanism effectively mitigates client drift. FedRDN also shows strong results, particularly on Fed-COSAS (77.67\% on C1) and Fed-Pancreas (74.37\% on C2), indicating its domain gap elimination strategy benefits generalization across highly heterogeneous domains. However, conventional methods like FedAvg and FedProx frequently underperform relative to Local baselines, especially on severely heterogeneous datasets such as Fed-Vessel and Fed-Pancreas. This confirms that standard aggregation protocols are susceptible to performance degradation under non-IID conditions common in medical imaging. The poor performance of PN across nearly all datasets further highlights the risks of aggressive normalization strategies that may discard valuable client-specific statistical information.

\textit{Personalized FL methods, which retain client-specific parameters or adapt the global model locally, consistently outperform generic approaches}. FedBN emerges as the most robust method overall, achieving top-tier performance on 7 out of 9 datasets, including Fed-Vessel, Fed-BUS, Fed-MG, Fed-Polyp, and FeTS2022. By maintaining local batch normalization statistics while sharing convolutional layers, FedBN effectively captures domain-specific feature distributions without sacrificing collaborative representation learning. This design choice proves particularly advantageous for medical imaging, where intensity distributions and texture patterns vary substantially across institutions due to differing acquisition protocols. SioBN and FedPer also demonstrate competitive performance, frequently ranking among the top two methods on datasets such as Fed-M\&Ms and Fed-Pancreas. SioBN's client-specific affine parameters offer a lightweight personalization mechanism that balances communication efficiency and model fidelity. FedPer's feature extractor sharing with personalized classifiers provides flexibility, though it exhibits higher variance on smaller datasets (\textit{e.g.}, Fed-COSAS).
The collective results underscore that no single FL method dominates universally; however, \textit{FedBN and its variants (SioBN) provide consistently superior and stable performance across a broad spectrum of medical imaging tasks}. The integration of domain-specific normalization layers emerges as a simple yet powerful inductive bias for handling cross-institutional heterogeneity. Future work should explore adaptive personalization strategies that dynamically allocate client-specific capacity based on task difficulty and domain distance.

\subsection{Discussion on Generalization Performance}
The generalization evaluation on five unseen datasets in Table~\ref{generalization}  reveals a critical relationship between a method's ability to generalize across participating clients and its performance on unseen domains. Notably, FedLWS exhibits a strong positive correlation: on Fed-Vessel and Fed-Prostate, where it achieves the highest segmentation accuracy across most clients (see Tables~\ref{Fundus_Table} and \ref{Prostate_Table}), it also attains the best generalization results on unseen fundus datasets (ARIA and IOSTAR) and the unseen prostate dataset (MSD). Conversely, on Fed-Polyp, where FedLWS delivers among the poorest performance on participating clients, as shown in Table~\ref{Polyp_Table}, it catastrophically fails on the unseen endoscopic dataset Kvasir‑SEG, plummeting to only 20.69\% Dice. This consistent pattern suggests that \textit{a method's generalization to unseen domains is strongly tied to its ability to perform well across the participating clients—when it excels on seen clients, it tends to transfer better; when it underperforms on seen clients, it is unlikely to generalize}. This observation challenges the conventional assumption that optimizing for seen clients necessarily leads to overfitting and poor out-of-distribution (OOD) performance. It also provides an important insight for domain generalization in federated learning: rather than solely focusing on explicit domain generalization techniques for unseen sites, \textit{an alternative and potentially more direct pathway is to enhance the model’s generalization capability on the participating clients themselves}. In other words, designing FL algorithms that achieve high and balanced performance across heterogeneous in‑distribution clients may implicitly equip the model with representations that are robust to unseen domains.
Additionally, methods like FedAvg, FedProx, and MOON exhibit moderate but stable performance across all five unseen datasets. While they never achieve top results, they also avoid catastrophic failures, making them safer choices in heterogeneous clinical environments where the target domain cannot be predetermined. 

\subsection{Discussion on Communication Frequency}
From Figs.~\ref{fig:comm_effect_gfl} and \ref{fig:comm_effect_pfl}, we observe that the impact of reducing communication rounds (\textit{i.e.}, increasing local updates per round) is highly dependent on both the dataset and the specific FL method. For generic FL approaches, methods such as FedAvg, FedProx, and FedNova generally maintain stable performance across different round settings on most datasets, indicating that they can tolerate a certain degree of local overfitting when communication is limited. However, their performance occasionally degrades when local epochs become too large (\textit{e.g.}, at 50 rounds), suggesting that excessive local updating may lead to client drift, particularly under severe data heterogeneity. In contrast, FedLWS and FedRDN exhibit more pronounced fluctuations, sometimes achieving superior results at intermediate round numbers but falling behind when rounds are extremely scarce. This behavior implies that these methods are more sensitive to the balance between local and global optimization, and their convergence may require a minimum number of communication steps. From a broader perspective, our findings underscore the importance of considering the communication–performance trade-off when deploying FL in real-world medical applications, where bandwidth constraints often limit the number of permissible rounds. Methods that preserve high accuracy even with infrequent communication—such as FedBN and SioBN—are particularly attractive for such settings. Moreover, the dataset-dependent variability in method rankings emphasizes that no single FL algorithm universally dominates; instead, the choice should be guided by the expected data heterogeneity and the available communication budget. Future research could explore adaptive strategies that dynamically adjust local update steps based on estimated heterogeneity or communication resource availability. Additionally, combining the strengths of normalization-based personalization with robust aggregation techniques may yield further improvements.

\subsection{Discussion on Fairness}
Based on the comprehensive experimental results presented in Tables \ref{Fairness_1} and \ref{Fairness_2}, several important observations can be drawn regarding the fairness characteristics of federated learning methods in medical image segmentation tasks. First, a clear trade-off is observed between gFL methods and pFL methods. pFL methods like Ditto achieve the lowest standard deviation on several datasets (\textit{e.g.}, 2.93 on Fed-Prostate, 5.62 on Fed-Polyp, 3.26 on Fed-COSAS) and the highest worst-client Dice scores (77.83\% on Fed-Prostate, 68.07\% on Fed-Polyp), but this often comes at the cost of average performance. For example, on Fed-Prostate, Ditto's average Dice (82.21\%) is lower than that of FedAvg (84.05\%) and FedLWS (84.92\%). This suggests that enhancing the performance of the worst-performing clients through personalization may involve a trade-off with slightly reduced overall average performance. Second, method performance varies significantly across different datasets. On datasets like Fed-Vessel and Fed-M\&Ms, most methods achieve relatively high average Dice scores and low standard deviations. In contrast, on datasets like Fed-Pancreas and Fed-Polyp, all methods exhibit substantially higher standard deviations and lower worst-client Dice scores. For instance, even the best-performing method on Fed-Pancreas (FedProx, with an average Dice of 63.39\%) suffers from a high standard deviation of 15.28 and a low worst-client Dice of only 34.43\%, indicating that certain datasets pose greater inherent challenges to achieving fairness. Across the eight datasets, FedRDN, FedNova, and FedBN demonstrate the most consistent fairness performance, frequently ranking among the top methods in terms of both standard deviation and worst-client Dice. FedAWA also shows robust performance, particularly on Fed-BUS where it achieves the best average Dice, lowest standard deviation, and highest worst-client Dice. Conversely, FedRoD and PN exhibit the least stable fairness characteristics, with performance deteriorating sharply on specific datasets. These observations have practical implications for method selection: \textit{if safeguarding the worst-performing clients is the primary concern (e.g., requiring all participating institutions to achieve acceptable performance), personalized methods like Ditto are preferable, despite a potential minor reduction in average performance; if pursuing high overall accuracy while maintaining fairness, FedRDN and FedBN emerge as superior choices}. Although existing methods have made progress, on challenging datasets like Fed-Pancreas, all methods exhibit standard deviations exceeding 12 and worst-client Dice scores below 42\%, indicating that the fairness problem is not yet fully resolved. Future research could explore hybrid strategies that integrate the strengths of both gFL and pFL methods---for example, combining FedRDN's heterogeneity reduction with Ditto's local personalization---to simultaneously optimize average performance and fairness. Additionally, investigating adaptive mechanisms that dynamically adjust the degree of personalization based on detected client heterogeneity could yield methods that automatically balance the trade-off between overall accuracy and worst-client performance.

\section{Conclusion}
\label{sec:conclusion}
In this paper, we introduced FL-MedSegBench, the first comprehensive benchmark for federated learning in medical image segmentation. Our benchmark encompasses nine diverse medical segmentation tasks spanning ten imaging modalities across both 2D and 3D formats, with realistic non-IID data partitions that mirror real-world clinical heterogeneity. We systematically evaluated eight generic FL methods and five personalized FL methods, assessing not only segmentation accuracy but also fairness, communication efficiency, convergence behavior, and generalization to unseen clients.

Our extensive experiments yield several key findings. First, personalized FL methods, particularly those incorporating client-specific batch normalization statistics (\textit{e.g.}, FedBN), consistently outperform generic approaches across most datasets, demonstrating the critical importance of adapting to local data distributions while preserving collaborative knowledge sharing. Second, no single FL method dominates universally; performance is highly dataset-dependent, with methods such as FedRDN and FedLWS excelling on specific tasks while underperforming on others. Third, communication frequency analysis reveals that the optimal trade-off between communication frequency and local computation varies substantially across datasets and methods, with normalization-based personalization methods (FedBN, SioBN) exhibiting remarkable robustness to reduced communication. Fourth, generalization assessment uncovers a critical relationship: a method's generalization to unseen domains is strongly tied to its ability to perform well across participating clients. This finding challenges the conventional assumption that optimizing for seen clients necessarily leads to overfitting and poor out-of-distribution performance, suggesting instead that enhancing model performance on participating clients themselves may offer a more direct pathway to robust domain generalization.

Our work provides foundational insights and practical guidelines for deploying FL in real-world medical imaging applications. The open-source benchmark toolkit, including standardized datasets, implementations, and evaluation protocols, aims to foster reproducible research and accelerate the development of more robust, efficient, and clinically applicable federated learning solutions. Future research directions include exploring adaptive personalization strategies that dynamically allocate client-specific capacity based on task difficulty, developing aggregation methods with stronger theoretical guarantees for OOD generalization, and integrating fairness constraints with personalization mechanisms to better protect all participating institutions.

\bibliographystyle{IEEEtran}
\bibliography{bare_jrnl}

@inproceedings{fedavg,
  title={Communication-efficient learning of deep networks from decentralized data},
  author={McMahan, Brendan and Moore, Eider and Ramage, Daniel and Hampson, Seth and y Arcas, Blaise Aguera},
  booktitle={AISTATS},
  pages={1273--1282},
  year={2017},
}

@inproceedings{fedprox,
    title = {Federated Optimization in Heterogeneous Networks},
    author={Li, Tian and Sahu, Anit Kumar and Zaheer, Manzil and Sanjabi, Maziar and Smith, Virginia},
    booktitle = {MLSys},
    pages = {429--450},
    volume = {2},
    year = {2020}
}

@inproceedings{li2021fedbn,
    title={FedBN: Federated Learning on Non-IID Features via Local Batch Normalization},
    author={Xiaoxiao Li and Meirui Jiang and Xiaofei Zhang and Michael Kamp and Qi Dou},
    booktitle={ICLR},
    year={2021}
}

@incollection{andreux2020siloed,
  title={Siloed Federated Learning for Multi-Centric Histopathology Datasets},
  author={Andreux, Mathieu and du Terrail, Jean Ogier and Beguier, Constance and Tramel, Eric W},
  booktitle={MICCAI Workshop on Domain Adaptation and Representation Transfer},
  pages={129--139},
  year={2020},
  publisher={Springer}
}

@article{zhu2021dsi,
  title={DSI-Net: Deep synergistic interaction network for joint classification and segmentation with endoscope images},
  author={Zhu, Meilu and Chen, Zhen and Yuan, Yixuan},
  journal={IEEE Transactions on Medical Imaging},
  volume={40},
  number={12},
  pages={3315--3325},
  year={2021},
}

@ARTICLE{FedOSS,
  author={Zhu, Meilu and Liao, Jing and Liu, Jun and Yuan, Yixuan},
  journal={IEEE Transactions on Medical Imaging},
  title={FedOSS: Federated Open Set Recognition via Inter-Client Discrepancy and Collaboration},
  year={2024},
  volume={43},
  number={1},
  pages={190-202},
  }

@inproceedings{chen2021personalized,
  title={Personalized retrogress-resilient framework for real-world medical federated learning},
  author={Chen, Zhen and Zhu, Meilu and Yang, Chen and Yuan, Yixuan},
  booktitle={MICCAI},
  pages={347--356},
  year={2021},
  organization={Springer}
}

@article{chen2022personalized,
  title={Personalized retrogress-resilient federated learning toward imbalanced medical data},
  author={Chen, Zhen and Yang, Chen and Zhu, Meilu and Peng, Zhe and Yuan, Yixuan},
  journal={IEEE Transactions on Medical Imaging},
  volume={41},
  number={12},
  pages={3663--3674},
  year={2022},
  publisher={IEEE}
}

@inproceedings{li2021model,
  title={Model-contrastive federated learning},
  author={Li, Qinbin and He, Bingsheng and Song, Dawn},
  booktitle={CVPR},
  pages={10713--10722},
  year={2021}
}

@article{chen2021bridging,
  title={On bridging generic and personalized federated learning for image classification},
  author={Chen, Hong-You and Chao, Wei-Lun},
  journal={ICLR},
  year={2022}
}

@inproceedings{kingma2014adam,
  title={Adam: A method for stochastic optimization},
  author={Kingma, Diederik P and Ba, Jimmy},
  booktitle={ICLR},
  year={2015}
}

@ARTICLE{zhufeddm2023,
  author={Zhu, Meilu and Chen, Zhen and Yuan, Yixuan},
  journal={IEEE Transactions on Medical Imaging}, 
  title={FedDM: Federated Weakly Supervised Segmentation via Annotation Calibration and Gradient De-Conflicting}, 
  year={2023},
  volume={42},
  number={6},
  pages={1632-1643},
  }

@article{wang2020tackling,
  title={Tackling the objective inconsistency problem in heterogeneous federated optimization},
  author={Wang, Jianyu and Liu, Qinghua and Liang, Hao and Joshi, Gauri and Poor, H Vincent},
  journal={NeurIPS},
  volume={33},
  pages={7611--7623},
  year={2020}
}

@article{li2020federated,
  title={Federated learning: Challenges, methods, and future directions},
  author={Li, Tian and Sahu, Anit Kumar and Talwalkar, Ameet and Smith, Virginia},
  journal={IEEE Signal Processing Magazine},
  volume={37},
  number={3},
  pages={50--60},
  year={2020},
  publisher={IEEE}
}

@inproceedings{fan2020pranet,
  title={Pranet: Parallel reverse attention network for polyp segmentation},
  author={Fan, Deng-Ping and Ji, Ge-Peng and Zhou, Tao and Chen, Geng and Fu, Huazhu and Shen, Jianbing and Shao, Ling},
  booktitle={MICCAI},
  pages={263--273},
  year={2020},
  organization={Springer}
}

@article{fraz2012ensemble,
  title={An ensemble classification-based approach applied to retinal blood vessel segmentation},
  author={Fraz, Muhammad Moazam and Remagnino, Paolo and Hoppe, Andreas and Uyyanonvara, Bunyarit and Rudnicka, Alicja R and Owen, Christopher G and Barman, Sarah A},
  journal={IEEE Transactions on Biomedical Engineering},
  volume={59},
  number={9},
  pages={2538--2548},
  year={2012},
  publisher={IEEE}
}

@article{holm2017dr,
  title={DR HAGIS—a fundus image database for the automatic extraction of retinal surface vessels from diabetic patients},
  author={Holm, Sven and Russell, Greg and Nourrit, Vincent and McLoughlin, Niall},
  journal={Journal of Medical Imaging},
  volume={4},
  number={1},
  pages={014503--014503},
  year={2017},
  publisher={Society of Photo-Optical Instrumentation Engineers}
}

@article{staal2004ridge,
  title={Ridge-based vessel segmentation in color images of the retina},
  author={Staal, Joes and Abr{\`a}moff, Michael D and Niemeijer, Meindert and Viergever, Max A and Van Ginneken, Bram},
  journal={IEEE Transactions on Medical Imaging},
  volume={23},
  number={4},
  pages={501--509},
  year={2004},
  publisher={IEEE}
}

@article{odstrcilik2013retinal,
  title={Retinal vessel segmentation by improved matched filtering: evaluation on a new high-resolution fundus image database},
  author={Odstrcilik, Jan and Kolar, Radim and Budai, Attila and Hornegger, Joachim and Jan, Jiri and Gazarek, Jiri and Kubena, Tomas and Cernosek, Pavel and Svoboda, Ondrej and Angelopoulou, Elli},
  journal={IET Image Processing},
  volume={7},
  number={4},
  pages={373--383},
  year={2013},
  publisher={Wiley Online Library}
}

@inproceedings{orlando2018towards,
  title={Towards a glaucoma risk index based on simulated hemodynamics from fundus images},
  author={Orlando, Jos{\'e} Ignacio and Barbosa Breda, Jo{\~a}o and Van Keer, Karel and Blaschko, Matthew B and Blanco, Pablo J and Bulant, Carlos A},
  booktitle={MICCAI},
  pages={65--73},
  year={2018},
  organization={Springer}
}

@inproceedings{sarhan2021transfer,
  title={Transfer learning through weighted loss function and group normalization for vessel segmentation from retinal images},
  author={Sarhan, Abdullah and Rokne, Jon and Alhajj, Reda and Crichton, Andrew},
  booktitle={ICPR},
  pages={9211--9218},
  year={2021},
  organization={IEEE}
}

@inproceedings{liu2021feddg,
  title={Feddg: Federated domain generalization on medical image segmentation via episodic learning in continuous frequency space},
  author={Liu, Quande and Chen, Cheng and Qin, Jing and Dou, Qi and Heng, Pheng-Ann},
  booktitle={CVPR},
  pages={1013--1023},
  year={2021}
}

@article{saha2022automated,
  title={Automated quantification of meibomian gland dropout in infrared meibography using deep learning},
  author={Saha, Ripon Kumar and Chowdhury, AM Mahmud and Na, Kyung-Sun and Hwang, Gyu Deok and Eom, Youngsub and Kim, Jaeyoung and Jeon, Hae-Gon and Hwang, Ho Sik and Chung, Euiheon},
  journal={The Ocular Surface},
  volume={26},
  pages={283--294},
  year={2022},
  publisher={Elsevier}
}

@dataset{li2024camg,
  author    = {Li, Li and Xiao, Kunhong},
  title     = {Children and Adolescents Meibomian Glands (CAMG) Dataset},
  year      = {2024},
  publisher = {figshare},
  type      = {Dataset},
  doi       = {10.6084/m9.figshare.27601101.v7}
}

@article{vazquez2017benchmark,
  title={A benchmark for endoluminal scene segmentation of colonoscopy images},
  author={V{\'a}zquez, David and Bernal, Jorge and S{\'a}nchez, F Javier and Fern{\'a}ndez-Esparrach, Gloria and L{\'o}pez, Antonio M and Romero, Adriana and Drozdzal, Michal and Courville, Aaron},
  journal={Journal of Healthcare Engineering},
  volume={2017},
  number={1},
  pages={4037190},
  year={2017},
  publisher={Wiley Online Library}
}

@article{bernal2015wm,
  title={WM-DOVA maps for accurate polyp highlighting in colonoscopy: Validation vs. saliency maps from physicians},
  author={Bernal, Jorge and S{\'a}nchez, F Javier and Fern{\'a}ndez-Esparrach, Gloria and Gil, Debora and Rodr{\'\i}guez, Cristina and Vilari{\~n}o, Fernando},
  journal={Computerized Medical Imaging and Graphics},
  volume={43},
  pages={99--111},
  year={2015},
  publisher={Elsevier}
}

@article{bernal2012towards,
  title={Towards automatic polyp detection with a polyp appearance model},
  author={Bernal, Jorge and S{\'a}nchez, Javier and Vilarino, Fernando},
  journal={Pattern Recognition},
  volume={45},
  number={9},
  pages={3166--3182},
  year={2012},
  publisher={Elsevier}
}

@inproceedings{hicks2021endotect,
  title={The EndoTect 2020 challenge: evaluation and comparison of classification, segmentation and inference time for endoscopy},
  author={Hicks, Steven A and Jha, Debesh and Thambawita, Vajira and Halvorsen, P{\aa}l and Hammer, Hugo L and Riegler, Michael A},
  booktitle={ICPR},
  pages={263--274},
  year={2021},
  organization={Springer}
}

@article{silva2014toward,
  title={Toward embedded detection of polyps in wce images for early diagnosis of colorectal cancer},
  author={Silva, Juan and Histace, Aymeric and Romain, Olivier and Dray, Xavier and Granado, Bertrand},
  journal={International Iournal of Computer Assisted Radiology and Surgery},
  volume={9},
  number={2},
  pages={283--293},
  year={2014},
  publisher={Springer}
}

@article{ardakani2023open,
  title={An open-access breast lesion ultrasound image database: Applicable in artificial intelligence studies},
  author={Ardakani, Ali Abbasian and Mohammadi, Afshin and Mirza-Aghazadeh-Attari, Mohammad and Acharya, U Rajendra},
  journal={Computers in Biology and Medicine},
  volume={152},
  pages={106438},
  year={2023},
  publisher={Elsevier}
}

@dataset{hesaraki2023bus,
  author    = {Hesaraki, Saba},
  title     = {Breast Ultrasound Images Dataset(BUSI)},
  year      = {2023},
  publisher = {Kaggle},
  url       = {https://www.kaggle.com/datasets/sabahesaraki/breast-ultrasound-images-dataset},
}

@dataset{cosas2024,
  author    = {Liu Jingxin and Da Qian and Shen Linlin and Li Yuexiang and Zuo Yanfei},
  title     = {Cross-Organ and Cross-Scanner Adenocarcinoma Segmentation Challenge},
  year      = {2024},
  url       = {https://cosas.grand-challenge.org/},
}

@article{iqbal2024memory,
  title={Memory-efficient transformer network with feature fusion for breast tumor segmentation and classification task},
  author={Iqbal, Ahmed and Sharif, Muhammad},
  journal={Engineering Applications of Artificial Intelligence},
  volume={127},
  pages={107292},
  year={2024},
  publisher={Elsevier}
}

@article{vallez2025bus,
  title={BUS-UCLM: Breast ultrasound lesion segmentation dataset},
  author={Vallez, Noelia and Bueno, Gloria and Deniz, Oscar and Rienda, Miguel Angel and Pastor, Carlos},
  journal={Scientific Data},
  volume={12},
  number={1},
  pages={242},
  year={2025},
  publisher={Nature Publishing Group UK London}
}

@article{zhang2025large,
  title={Large-scale multi-center CT and MRI segmentation of pancreas with deep learning},
  author={Zhang, Zheyuan and Keles, Elif and Durak, Gorkem and Taktak, Yavuz and Susladkar, Onkar and Gorade, Vandan and Jha, Debesh and Ormeci, Asli C and Medetalibeyoglu, Alpay and Yao, Lanhong and others},
  journal={Medical Image Analysis},
  volume={99},
  pages={103382},
  year={2025},
  publisher={Elsevier}
}

@article{martin2023deep,
  title={Deep learning segmentation of the right ventricle in cardiac MRI: the M\&Ms challenge},
  author={Mart{\'\i}n-Isla, Carlos and Campello, V{\'\i}ctor M and Izquierdo, Cristian and Kushibar, Kaisar and Sendra-Balcells, Carla and Gkontra, Polyxeni and Sojoudi, Alireza and Fulton, Mitchell J and Arega, Tewodros Weldebirhan and Punithakumar, Kumaradevan and others},
  journal={IEEE Journal of Biomedical and Health Informatics},
  volume={27},
  number={7},
  pages={3302--3313},
  year={2023},
  publisher={IEEE}
}

@inproceedings{shi2025fedawa,
  title={FedAWA: Adaptive Optimization of Aggregation Weights in Federated Learning Using Client Vectors},
  author={Shi, Changlong and Zhao, He and Zhang, Bingjie and Zhou, Mingyuan and Guo, Dandan and Chang, Yi},
  booktitle={CVPR},
  pages={30651--30660},
  year={2025}
}

@inproceedings{yan2025simple,
  title={A simple data augmentation for feature distribution skewed federated learning},
  author={Yan, Yunlu and Fu, Huazhu and Li, Yuexiang and Xie, Jinheng and Ma, Jun and Yang, Guang and Zhu, Lei},
  booktitle={CVPR},
  pages={25749--25758},
  year={2025}
}

@inproceedings{wang2025population,
  title={Population Normalization for Federated Learning},
  author={Wang, Zhuoyao and Yi, Fan and Gong, Peizhu and He, Caitou and Jin, Cheng and Zhang, Weizhong},
  booktitle={CVPR},
  pages={10214--10223},
  year={2025}
}

@inproceedings{shi2025fedlws,
  title={FedLWS: Federated Learning with Adaptive Layer-wise Weight Shrinking},
  author={Shi, Changlong and Li, Jinmeng and Zhao, He and Guo, Dandan and Chang, Yi},
  booktitle={ICLR},
  year={2025}
}

@article{arivazhagan2019federated,
  title={Federated learning with personalization layers},
  author={Arivazhagan, Manoj Ghuhan and Aggarwal, Vinay and Singh, Aaditya Kumar and Choudhary, Sunav},
  journal={arXiv preprint arXiv:1912.00818},
  year={2019}
}

@inproceedings{li2021ditto,
  title={Ditto: Fair and robust federated learning through personalization},
  author={Li, Tian and Hu, Shengyuan and Beirami, Ahmad and Smith, Virginia},
  booktitle={ICML},
  pages={6357--6368},
  year={2021},
  organization={PMLR}
}

@article{zhong2025cross,
  title={Cross-view discrepancy-dependency network for volumetric medical image segmentation},
  author={Zhong, Shengzhou and Wang, Wenxu and Feng, Qianjin and Zhang, Yu and Ning, Zhenyuan},
  journal={Medical Image Analysis},
  volume={99},
  pages={103329},
  year={2025},
  publisher={Elsevier}
}

@article{huang2025multidimensional,
  title={Multidimensional Directionality-Enhanced Segmentation via large vision model},
  author={Huang, Xingru and Yue, Changpeng and Guo, Yihao and Huang, Jian and Jiang, Zhengyao and Wang, Mingkuan and Xu, Zhaoyang and Zhang, Guangyuan and Liu, Jin and Zhang, Tianyun and others},
  journal={Medical Image Analysis},
  volume={101},
  pages={103395},
  year={2025},
  publisher={Elsevier}
}

@inproceedings{butoi2023universeg,
  title={Universeg: Universal medical image segmentation},
  author={Butoi, Victor Ion and Ortiz, Jose Javier Gonzalez and Ma, Tianyu and Sabuncu, Mert R and Guttag, John and Dalca, Adrian V},
  booktitle={ICCV},
  pages={21438--21451},
  year={2023}
}

@article{azad2024medical,
  title={Medical image segmentation review: The success of u-net},
  author={Azad, Reza and Aghdam, Ehsan Khodapanah and Rauland, Amelie and Jia, Yiwei and Avval, Atlas Haddadi and Bozorgpour, Afshin and Karimijafarbigloo, Sanaz and Cohen, Joseph Paul and Adeli, Ehsan and Merhof, Dorit},
  journal={IEEE Transactions on Pattern Analysis and Machine Intelligence},
  volume={46},
  number={12},
  pages={10076--10095},
  year={2024},
  publisher={IEEE}
}

@article{gostin2009beyond,
  title={Beyond the HIPAA privacy rule: enhancing privacy, improving health through research},
  author={Gostin, Lawrence O and Levit, Laura A and Nass, Sharyl J},
  year={2009},
  publisher={National Academies Press}
}

@article{goddard2017eu,
  title={The EU General Data Protection Regulation (GDPR): European regulation that has a global impact},
  author={Goddard, Michelle},
  journal={International Journal of Market Research},
  volume={59},
  number={6},
  pages={703--705},
  year={2017},
  publisher={SAGE Publications Sage UK: London, England}
}

@article{guan2024federated,
  title={Federated learning for medical image analysis: A survey},
  author={Guan, Hao and Yap, Pew-Thian and Bozoki, Andrea and Liu, Mingxia},
  journal={Pattern Recognition},
  volume={151},
  pages={110424},
  year={2024},
  publisher={Elsevier}
}

@article{tan2022towards,
  title={Towards personalized federated learning},
  author={Tan, Alysa Ziying and Yu, Han and Cui, Lizhen and Yang, Qiang},
  journal={IEEE Transactions on Neural Networks and Learning Systems},
  volume={34},
  number={12},
  pages={9587--9603},
  year={2022},
  publisher={IEEE}
}

@article{ogier2022flamby,
  title={Flamby: Datasets and benchmarks for cross-silo federated learning in realistic healthcare settings},
  author={Ogier du Terrail, Jean and Ayed, Samy-Safwan and Cyffers, Edwige and Grimberg, Felix and He, Chaoyang and Loeb, Regis and Mangold, Paul and Marchand, Tanguy and Marfoq, Othmane and Mushtaq, Erum and others},
  journal={NeurIPS},
  volume={35},
  pages={5315--5334},
  year={2022}
}

@article{manthe2024federated,
  title={Federated brain tumor segmentation: An extensive benchmark},
  author={Manthe, Matthis and Duffner, Stefan and Lartizien, Carole},
  journal={Medical Image Analysis},
  volume={97},
  pages={103270},
  year={2024},
  publisher={Elsevier}
}

@article{liu2025federated,
  title={Federated modality-specific encoders and partially personalized fusion decoder for multimodal brain tumor segmentation},
  author={Liu, Hong and Wei, Dong and Dai, Qian and Wu, Xian and Zheng, Yefeng and Wang, Liansheng},
  journal={Medical Image Analysis},
  volume = {106},
  pages = {103759},
  year={2025},
  publisher={Elsevier}
}

@article{zhang2025pathfl,
  title={PathFL: Multi-alignment Federated Learning for pathology image segmentation},
  author={Zhang, Yuan and Chen, Feng and Qi, Yaolei and Yang, Guanyu and Fu, Huazhu},
  journal={Medical Image Analysis},
  volume = {105},
  pages = {103670},
  year={2025},
  publisher={Elsevier}
}

@article{zhou2023fedftn,
  title={FedFTN: Personalized federated learning with deep feature transformation network for multi-institutional low-count PET denoising},
  author={Zhou, Bo and Xie, Huidong and Liu, Qiong and Chen, Xiongchao and Guo, Xueqi and Feng, Zhicheng and Hou, Jun and Zhou, S Kevin and Li, Biao and Rominger, Axel and others},
  journal={Medical Image Analysis},
  volume={90},
  pages={102993},
  year={2023},
  publisher={Elsevier}
}

@article{jin2023backdoor,
  title={Backdoor attack and defense in federated generative adversarial network-based medical image synthesis},
  author={Jin, Ruinan and Li, Xiaoxiao},
  journal={Medical Image Analysis},
  volume={90},
  pages={102965},
  year={2023},
  publisher={Elsevier}
}

@article{kim2024federated,
  title={Federated learning with knowledge distillation for multi-organ segmentation with partially labeled datasets},
  author={Kim, Soopil and Park, Heejung and Kang, Myeongkyun and Jin, Kyong Hwan and Adeli, Ehsan and Pohl, Kilian M and Park, Sang Hyun},
  journal={Medical Image Analysis},
  volume={95},
  pages={103156},
  year={2024},
  publisher={Elsevier}
}

@article{yang2021federated,
  title={Federated semi-supervised learning for COVID region segmentation in chest CT using multi-national data from China, Italy, Japan},
  author={Yang, Dong and Xu, Ziyue and Li, Wenqi and Myronenko, Andriy and Roth, Holger R and Harmon, Stephanie and Xu, Sheng and Turkbey, Baris and Turkbey, Evrim and Wang, Xiaosong and others},
  journal={Medical Image Analysis},
  volume={70},
  pages={101992},
  year={2021},
  publisher={Elsevier}
}

@article{t2020personalized,
  title={Personalized federated learning with moreau envelopes},
  author={T Dinh, Canh and Tran, Nguyen and Nguyen, Josh},
  journal={NeurIPS},
  volume={33},
  pages={21394--21405},
  year={2020}
}

@misc{spyridon_bakas_2022_6362409,
  author       = {Spyridon Bakas and
                  Sarthak Pati and
                  Micah Sheller and
                  Alexandros Karargyris and
                  Peter Mattson and
                  Brandon Edwards and
                  Ujjwal Baid and
                  Yong Chen and
                  Russell (Taki) Shinohara and
                  Jason Martin and
                  Bjoern Menze and
                  Maximilian Zenk and
                  Klaus Maier-Hein and
                  Ralf Floca and
                  Annika Reinke and
                  Lena Maier-Hein and
                  Fabian Isensee and
                  David Zimmerer and
                  Yong Chen},
  title        = {The Federated Tumor Segmentation (FeTS) Challenge
                   2022
                  },
  month        = mar,
  year         = 2022,
  publisher    = {Zenodo},
  doi          = {10.5281/zenodo.6362409},
  url          = {https://doi.org/10.5281/zenodo.6362409},
}

@ARTICLE{chen2025decentralized,
  author={Chen, Jingyun and Yuan, Yading},
  journal={IEEE Transactions on Medical Imaging}, 
  title={Decentralized Personalization for Federated Medical Image Segmentation via Gossip Contrastive Mutual Learning}, 
  year={2025},
  volume={44},
  number={7},
  pages={2768-2783},
  }

@inproceedings{ronneberger2015u,
  title={U-net: Convolutional networks for biomedical image segmentation},
  author={Ronneberger, Olaf and Fischer, Philipp and Brox, Thomas},
  booktitle={MICCAI},
  pages={234--241},
  year={2015},
  organization={Springer}
}

@article{farnell2008enhancement,
  title={Enhancement of blood vessels in digital fundus photographs via the application of multiscale line operators},
  author={Farnell, Damian JJ and Hatfield, Fraser N and Knox, Paul and Reakes, Michael and Spencer, Stan and Parry, David and Harding, Simon P},
  journal={Journal of the Franklin institute},
  volume={345},
  number={7},
  pages={748--765},
  year={2008},
  publisher={Elsevier}
}

@ARTICLE{7530915,
    author={J. {Zhang} and B. {Dashtbozorg} and E. {Bekkers} and J. P. W. {Pluim} and R. {Duits} and B. M. {ter Haar Romeny}},
    journal={IEEE Transactions on Medical Imaging},
    title={Robust Retinal Vessel Segmentation via Locally Adaptive Derivative Frames in Orientation Scores},
    year={2016},
    volume={35},
    number={12},
    pages={2631-2644},
    ISSN={0278-0062},
    month={Dec},
}

@article{antonelli2022medical,
  title={The medical segmentation decathlon},
  author={Antonelli, Michela and Reinke, Annika and Bakas, Spyridon and Farahani, Keyvan and Kopp-Schneider, Annette and Landman, Bennett A and Litjens, Geert and Menze, Bjoern and Ronneberger, Olaf and Summers, Ronald M and others},
  journal={Nature Communications},
  volume={13},
  number={1},
  pages={4128},
  year={2022},
  publisher={Nature Publishing Group UK London}
}

@inproceedings{jha2020kvasir, 
title={Kvasir-seg: A segmented polyp dataset}, 
author={Jha, Debesh and Smedsrud, Pia H and Riegler, Michael A and Halvorsen, P{\aa}l and de Lange, Thomas and Johansen, Dag and Johansen, H{\aa}vard D}, 
booktitle={MMM}, 
pages={451--462}, 
year={2020}, 
organization={Springer} 
}

@article{yap2017automated,
  title={Automated breast ultrasound lesions detection using convolutional neural networks},
  author={Yap, Moi Hoon and Pons, Gerard and Marti, Joan and Ganau, Sergi and Sentis, Melcior and Zwiggelaar, Reyer and Davison, Adrian K and Marti, Robert},
  journal={IEEE Journal of Biomedical and Health Informatics},
  volume={22},
  number={4},
  pages={1218--1226},
  year={2017},
  publisher={IEEE}
}

@article{zhu2025fedbm,
  title={FedBM: Stealing knowledge from pre-trained language models for heterogeneous federated learning},
  author={Zhu, Meilu and Yang, Qiushi and Gao, Zhifan and Yuan, Yixuan and Liu, Jun},
  journal={Medical Image Analysis},
  volume={102},
  pages={103524},
  year={2025},
  publisher={Elsevier}
}

@inproceedings{zhu2024stealing,
  title={Stealing knowledge from pre-trained language models for federated classifier debiasing},
  author={Zhu, Meilu and Yang, Qiushi and Gao, Zhifan and Liu, Jun and Yuan, Yixuan},
  booktitle={MICCAI},
  pages={685--695},
  year={2024},
  organization={Springer}
}

@article{li2025challenges,
  title={From challenges and pitfalls to recommendations and opportunities: Implementing federated learning in healthcare},
  author={Li, Ming and Xu, Pengcheng and Hu, Junjie and Tang, Zeyu and Yang, Guang},
  journal={Medical Image Analysis},
  volume={101},
  pages={103497},
  year={2025},
  publisher={Elsevier}
}



%

\end{document}